\documentclass{elsarticle}

\usepackage[margin=1in]{geometry}

\usepackage{bbm}
\usepackage{graphicx}
\usepackage{pstool}
\usepackage{wrapfig}
\usepackage{caption}
\usepackage{subcaption}
\usepackage[section]{placeins} 

\usepackage{fancyhdr} 
\usepackage{lastpage} 
\usepackage{extramarks} 

\usepackage{xcolor} 
\usepackage{enumerate} 
\usepackage{paralist} 
\usepackage{amsmath, amsthm, amssymb, mathtools}
\usepackage{mathabx, pifont, stmaryrd} 
\usepackage[explicit]{titlesec} 
\usepackage{etoolbox} 
\usepackage{bibentry} 
\makeatletter\let\saved@bibitem\@bibitem\makeatother 
\usepackage[colorlinks, bookmarksopen, bookmarksnumbered,
            citecolor=red,urlcolor=red]{hyperref} 
\makeatletter\let\@bibitem\saved@bibitem\makeatother 

\usepackage{algorithm}
\usepackage{algorithmic}

\theoremstyle{definition}

\usepackage{bm} 
\usepackage{amsfonts} 

\newcommand{\argoptunc}[2]{\underset{#1}{\arg\min} ~~ #2}

\newcommand{\pder}[2]{\ensuremath{\frac{\partial #1}{\partial #2}}}

\newcommand{\Acal}{\ensuremath{\mathcal{A}}}

\newcommand{\Dcal}{\ensuremath{\mathcal{D}}}
\newcommand{\Ecal}{\ensuremath{\mathcal{E}}}
\newcommand{\Fcal}{\ensuremath{\mathcal{F}}}

\newcommand{\Hcal}{\ensuremath{\mathcal{H}}}

\newcommand{\Kcal}{\ensuremath{\mathcal{K}}}

\newcommand{\Mcal}{\ensuremath{\mathcal{M}}}
\newcommand{\Ncal}{\ensuremath{\mathcal{N}}}

\newcommand{\Qcal}{\ensuremath{\mathcal{Q}}}

\newcommand{\Scal}{\ensuremath{\mathcal{S}}}

\newcommand{\Ucal}{\ensuremath{\mathcal{U}}}
\newcommand{\Vcal}{\ensuremath{\mathcal{V}}}

\newcommand{\Ebb}{\ensuremath{\mathbb{E}}}

\newcommand{\Nbb}{\ensuremath{\mathbb{N} }}

\newcommand{\Rbb}{\ensuremath{\mathbb{R} }}

\newcommand\Ibm{{\ensuremath{\bm{I}}}}

\newcommand\nbm{{\ensuremath{\bm{n}}}}

\newcommand\sbm{{\ensuremath{\bm{s}}}}

\newcommand\ubm{{\ensuremath{\bm{u}}}}

\newcommand\zbm{{\ensuremath{\bm{z}}}}

\newcommand\thetabold{{\ensuremath{\boldsymbol{\theta}}}}

\newcommand\iotabold{{\ensuremath{\boldsymbol{\iota}}}}

\newcommand\psibold{{\ensuremath{\boldsymbol{\psi}}}}

\usepackage{tikz}
\usepackage{pgfplots}
\usepackage{pgfplotstable, filecontents, booktabs}
\pgfplotsset{compat=1.9}

\usetikzlibrary{pgfplots.groupplots}
\usepgfplotslibrary{fillbetween}
\usetikzlibrary{calc,fit,matrix,arrows,automata,positioning,shapes}
\usetikzlibrary{arrows.meta}

\pgfplotsset{select coords between index/.style 2 args={
    x filter/.code={
        \ifnum\coordindex<#1\fi
        \ifnum\coordindex>#2\fi
    }
}}

\tikzset{
 invisible/.style={opacity=0},
 visible on/.style={alt={#1{}{invisible}}},
 alt/.code args={<#1>#2#3}{%
   \alt<#1>{\pgfkeysalso{#2}}{\pgfkeysalso{#3}}
 },
}

\newcommand{\colorbarMatlabJet}[5]{
\begin{tikzpicture}
\begin{axis}[
   hide axis, scale only axis,
   height=0pt, width=0pt,
   colormap={jet}{rgb255=(0,0,131) rgb255=(0,0,135) rgb255=(0,0,139) rgb255=(0,0,143) rgb255=(0,0,147) rgb255=(0,0,151) rgb255=(0,0,155) rgb255=(0,0,159) rgb255=(0,0,163) rgb255=(0,0,167) rgb255=(0,0,171) rgb255=(0,0,175) rgb255=(0,0,179) rgb255=(0,0,183) rgb255=(0,0,187) rgb255=(0,0,191) rgb255=(0,0,195) rgb255=(0,0,199) rgb255=(0,0,203) rgb255=(0,0,207) rgb255=(0,0,211) rgb255=(0,0,215) rgb255=(0,0,219) rgb255=(0,0,223) rgb255=(0,0,227) rgb255=(0,0,231) rgb255=(0,0,235) rgb255=(0,0,239) rgb255=(0,0,243) rgb255=(0,0,247) rgb255=(0,0,251) rgb255=(0,0,255) rgb255=(0,4,255) rgb255=(0,8,255) rgb255=(0,12,255) rgb255=(0,16,255) rgb255=(0,20,255) rgb255=(0,24,255) rgb255=(0,28,255) rgb255=(0,32,255) rgb255=(0,36,255) rgb255=(0,40,255) rgb255=(0,44,255) rgb255=(0,48,255) rgb255=(0,52,255) rgb255=(0,56,255) rgb255=(0,60,255) rgb255=(0,64,255) rgb255=(0,68,255) rgb255=(0,72,255) rgb255=(0,76,255) rgb255=(0,80,255) rgb255=(0,84,255) rgb255=(0,88,255) rgb255=(0,92,255) rgb255=(0,96,255) rgb255=(0,100,255) rgb255=(0,104,255) rgb255=(0,108,255) rgb255=(0,112,255) rgb255=(0,116,255) rgb255=(0,120,255) rgb255=(0,124,255) rgb255=(0,128,255) rgb255=(0,131,255) rgb255=(0,135,255) rgb255=(0,139,255) rgb255=(0,143,255) rgb255=(0,147,255) rgb255=(0,151,255) rgb255=(0,155,255) rgb255=(0,159,255) rgb255=(0,163,255) rgb255=(0,167,255) rgb255=(0,171,255) rgb255=(0,175,255) rgb255=(0,179,255) rgb255=(0,183,255) rgb255=(0,187,255) rgb255=(0,191,255) rgb255=(0,195,255) rgb255=(0,199,255) rgb255=(0,203,255) rgb255=(0,207,255) rgb255=(0,211,255) rgb255=(0,215,255) rgb255=(0,219,255) rgb255=(0,223,255) rgb255=(0,227,255) rgb255=(0,231,255) rgb255=(0,235,255) rgb255=(0,239,255) rgb255=(0,243,255) rgb255=(0,247,255) rgb255=(0,251,255) rgb255=(0,255,255) rgb255=(4,255,251) rgb255=(8,255,247) rgb255=(12,255,243) rgb255=(16,255,239) rgb255=(20,255,235) rgb255=(24,255,231) rgb255=(28,255,227) rgb255=(32,255,223) rgb255=(36,255,219) rgb255=(40,255,215) rgb255=(44,255,211) rgb255=(48,255,207) rgb255=(52,255,203) rgb255=(56,255,199) rgb255=(60,255,195) rgb255=(64,255,191) rgb255=(68,255,187) rgb255=(72,255,183) rgb255=(76,255,179) rgb255=(80,255,175) rgb255=(84,255,171) rgb255=(88,255,167) rgb255=(92,255,163) rgb255=(96,255,159) rgb255=(100,255,155) rgb255=(104,255,151) rgb255=(108,255,147) rgb255=(112,255,143) rgb255=(116,255,139) rgb255=(120,255,135) rgb255=(124,255,131) rgb255=(128,255,128) rgb255=(131,255,124) rgb255=(135,255,120) rgb255=(139,255,116) rgb255=(143,255,112) rgb255=(147,255,108) rgb255=(151,255,104) rgb255=(155,255,100) rgb255=(159,255,96) rgb255=(163,255,92) rgb255=(167,255,88) rgb255=(171,255,84) rgb255=(175,255,80) rgb255=(179,255,76) rgb255=(183,255,72) rgb255=(187,255,68) rgb255=(191,255,64) rgb255=(195,255,60) rgb255=(199,255,56) rgb255=(203,255,52) rgb255=(207,255,48) rgb255=(211,255,44) rgb255=(215,255,40) rgb255=(219,255,36) rgb255=(223,255,32) rgb255=(227,255,28) rgb255=(231,255,24) rgb255=(235,255,20) rgb255=(239,255,16) rgb255=(243,255,12) rgb255=(247,255,8) rgb255=(251,255,4) rgb255=(255,255,0) rgb255=(255,251,0) rgb255=(255,247,0) rgb255=(255,243,0) rgb255=(255,239,0) rgb255=(255,235,0) rgb255=(255,231,0) rgb255=(255,227,0) rgb255=(255,223,0) rgb255=(255,219,0) rgb255=(255,215,0) rgb255=(255,211,0) rgb255=(255,207,0) rgb255=(255,203,0) rgb255=(255,199,0) rgb255=(255,195,0) rgb255=(255,191,0) rgb255=(255,187,0) rgb255=(255,183,0) rgb255=(255,179,0) rgb255=(255,175,0) rgb255=(255,171,0) rgb255=(255,167,0) rgb255=(255,163,0) rgb255=(255,159,0) rgb255=(255,155,0) rgb255=(255,151,0) rgb255=(255,147,0) rgb255=(255,143,0) rgb255=(255,139,0) rgb255=(255,135,0) rgb255=(255,131,0) rgb255=(255,128,0) rgb255=(255,124,0) rgb255=(255,120,0) rgb255=(255,116,0) rgb255=(255,112,0) rgb255=(255,108,0) rgb255=(255,104,0) rgb255=(255,100,0) rgb255=(255,96,0) rgb255=(255,92,0) rgb255=(255,88,0) rgb255=(255,84,0) rgb255=(255,80,0) rgb255=(255,76,0) rgb255=(255,72,0) rgb255=(255,68,0) rgb255=(255,64,0) rgb255=(255,60,0) rgb255=(255,56,0) rgb255=(255,52,0) rgb255=(255,48,0) rgb255=(255,44,0) rgb255=(255,40,0) rgb255=(255,36,0) rgb255=(255,32,0) rgb255=(255,28,0) rgb255=(255,24,0) rgb255=(255,20,0) rgb255=(255,16,0) rgb255=(255,12,0) rgb255=(255,8,0) rgb255=(255,4,0) rgb255=(255,0,0) rgb255=(251,0,0) rgb255=(247,0,0) rgb255=(243,0,0) rgb255=(239,0,0) rgb255=(235,0,0) rgb255=(231,0,0) rgb255=(227,0,0) rgb255=(223,0,0) rgb255=(219,0,0) rgb255=(215,0,0) rgb255=(211,0,0) rgb255=(207,0,0) rgb255=(203,0,0) rgb255=(199,0,0) rgb255=(195,0,0) rgb255=(191,0,0) rgb255=(187,0,0) rgb255=(183,0,0) rgb255=(179,0,0) rgb255=(175,0,0) rgb255=(171,0,0) rgb255=(167,0,0) rgb255=(163,0,0) rgb255=(159,0,0) rgb255=(155,0,0) rgb255=(151,0,0) rgb255=(147,0,0) rgb255=(143,0,0) rgb255=(139,0,0) rgb255=(135,0,0) rgb255=(131,0,0) rgb255=(128,0,0) },
   colorbar horizontal,
   point meta min=#1, point meta max=#5,
   colorbar style={width=10cm, xtick={#1, #2, #3, #4, #5}}
]
\addplot [draw=none] coordinates {(0,0)};
\end{axis}
\end{tikzpicture}
}

\usepackage{setspace}
\usepackage{enumitem}
\usepackage{ulem}

\newbool{fastcompile}
\setbool{fastcompile}{true}

\begin{document}
\title{Generative Learning for Forecasting the Dynamics of Complex Systems}

\author[rvt1]{Han Gao}
\ead{hgao1@seas.harvard.edu}
\author[rvt1]{Sebastian Kaltenbach}
\ead{skaltenbach@seas.harvard.edu}
\author[rvt1]{Petros Koumoutsakos\corref{cor1}}
\ead{petros@seas.harvard.edu}

\address[rvt1]{School of Engineering and Applied Sciences, Harvard University, 29 Oxford Street, Cambridge, MA 02138, US}
\cortext[cor1]{Corresponding author}

\begin{keyword} 
{\color{black} diffusion models, learning effective dynamics, turbulent flows}
\end{keyword}

\begin{abstract}
We introduce generative models for accelerating simulations of complex systems through learning and evolving their effective dynamics. In the proposed Generative Learning of Effective Dynamics (G-LED), instances of high dimensional data are down sampled to a lower dimensional manifold that is evolved through an auto-regressive attention mechanism. In turn, Bayesian diffusion models, that map this low-dimensional manifold onto its corresponding  high-dimensional space, capture the statistics of the system dynamics.  We demonstrate the capabilities and drawbacks of G-LED in simulations of several benchmark systems, including the Kuramoto–Sivashinsky (KS) equation,  two-dimensional high Reynolds number flow over a backward-facing step, and simulations of three-dimensional turbulent channel flow. The results demonstrate that generative learning offers new frontiers for the accurate forecasting of the statistical properties of complex systems at a reduced computational cost.
\end{abstract}
    
\maketitle

\section{Introduction}
Reliable forecasting of critical phenomena, such as weather and epidemics, depends on the efficiency and veracity of numerical simulations. A vast number of these simulations are founded on models described by Partial Differential Equations (PDEs) expressing multiphysics and multiscale dynamics of turbulence~\cite{wilcox1988multiscale}, neuroscience~\cite{dura2019netpyne}, climate~\cite{climatenas} and ocean dynamics~\cite{mahadevan2016impact}.
Large-scale simulations that resolve all spatiotemporal scales often address only idealized systems while their computational cost can prevent experimentation, design optimization, and uncertainty quantification \cite{rossinelli2013a,Palmer2015,moser2023numerical,sharma2024exawind}. On the other hand, reduced order and coarse-grained models are fast but limited by the simplification/linearization of the complex system dynamics and long-term inaccurate predictions due to unresolved scales \cite{bui2008model,peng2021multiscale}. 
Hybrid approaches rely on judicious approximations of the interactions between processes occurring over different scales, and a number of notable frameworks have been proposed, including the Equation Free Framework (EFF)~\cite{kevrekidis2004equation, kevrekidis2003equation,bar2019learning}, 
Heterogeneous Multiscale Method (HMM)~\cite{weinan2007heterogeneous}, and Flow Averaged Integrators (FLAVOR)~\cite{tao2010nonintrusive}. 
Their success depends on the separation of scales in the system dynamics and the capability to capture the transfer of information between scales. While it is undisputed that  EFF, HMM, and FLAVOR have revolutionized the field of multiscale modeling and simulation,  two critical issues have limited their potential: (i) the accuracy of propagating the coarse-grained dynamics hinges on the employed time integrators, and (ii) ineffective information transfer, in particular from coarse to fine scale dynamics, greatly limits their potential. 

Several notable algorithms have been proposed in the last few years to overcome these limitations by learning a lower-dimensional representation of the current state of the system and evolving the temporal dynamics in the low-dimensional manifold \cite{han2021predicting,kaltenbach2021physics, vlachas2022multiscale,ren2022phycrnet,sun2023unifying,cachay2023dyffusion,agathos2024accelerating}. These approaches, in general, deploy an encoder from the high-dimensional state of the targeted system to its lower-dimensional representation as well as a decoder, the reverse mappping, with DNN \cite{kingma2013auto}. The temporal dynamics can be represented via a recurrent neural network such as an LSTM \cite{vlachas2022multiscale},  Echo-State networks \cite{racca2022modelling} or other autoregressive dynamics \cite{fukami2019super,geneva2020multi,kaltenbach2021physics,kaltenbach2023interpretable}. Moreover, interpretable latent space dynamics can  be constructed \cite{menier2023interpretable,gupta2023mori} based on the Koopman Operator theory and the Mori-Zwanzig formalism \cite{mezic2013analysis,mezic2021koopman}. 
The algorithm of Learned Effective Dynamics (LED) ~\cite{vlachas2022multiscale} addresses the aforementioned issues of hybrid approaches while maintaining the feature of systematically alternating between high-and low-resolution simulations. In LED, Variational Autoencoders (AEs)  transfer the information between coarse and fine-scale descriptions and  Recurrent Neural Networks (RNNs)  with Long-Short Term Memory  (LSTMs) gating ~\cite{hochreiter1997long,vlachas2018data} evolve the latent dynamics. However, to the best of our knowledge, LED as well as the above-mentioned methodologies have not been able to capture complex multiscale dynamics such as those involved in turbulent flows.  

In the present work we introduce the Generative Learning of Effective Dynamics (G-LED) framework by deploying recent advances in generative learning \cite{song2019generative,ho2020denoising,song2020score,song2020denoising,saharia2022photorealistic, prabhudesai2023diffusion} to extend our previous work \cite{vlachas2022multiscale, menier2023interpretable, kivcic2023adaptive} for capturing the dynamics of complex systems. Generative learning has found great success in computer vision and extensions to flow reconstruction \cite{shu2023physics}, multi-fidelity design \cite{ogoke2023inexpensive}, inverse inference \cite{jacobsen2023cocogen}, and forward prediction \cite{kohl2020learning,jadhav2023stressd,lienen2023generative,price2023gencast,li2023seeds}. 
Within G-LED, a Bayesian diffusion model is employed as a decoder. This diffusion-based decoder is able to incorporate physical information due to a novel formulation based on conditional diffusion \cite{gao2023bayesian} and virtual observables \cite{kaltenbach2020incorporating}. The  reverse process of the diffusion model serves as a flexible method for expressing the statistics of the fields described by the governing PDEs. To identify the latent space, we deploy a simple subsampling of the high-dimensional snapshots in G-LED, thus allowing the decoder to be trained independently from the latent dynamics.
In the temporal evolution, the critical challenge is to capture the coarse-grained dynamics without sacrificing the expressive power of the diffusion model. The proposed framework applies a multi-head auto-regressive attention model \cite{vaswani2017attention,child2019generating,katharopoulos2020transformers}, taking advantage of its low memory footprint and improved expressivity as shown in the latest advances in natural language processing \cite{jiang2023mistral}.
Due to the mentioned changes, G-LED significantly extends the LED framework and enables the identification of effective dynamics for complex systems such as complex three-dimensional (3D) turbulent flows. In case the system of interest is parameterized, we note that by adding explicit dependence on the parameters for the latent-space dynamics similar to parameter embeddings \cite{geneva2022transformers,han2021predicting} and neural operators \cite{kovachki2023neural,pan2023neural,li2022fourier,lu2021learning,kaltenbach2023semi}, G-LED would be applicable to these systems as well.

The remainder of the paper is structured as follows. Section~\ref{sec:method} introduces the proposed methodology including  the Bayesian Diffusion model and the incorporation of physical information. Section~\ref{sec:result} presents three applications of the framework to benchmark problems, including the 1D Kuramoto–Sivashinsky (KS) equation, a 2D high Reynolds number flow over a backward-facing step, and a 3D turbulent channel flow. We show that G-LED outperforms previous state-of-the-art neural surrogates and is able to accurately represent global flow quantities, such as the energy spectrum in a turbulent flow. Finally, Section~\ref{sec:conclusion} summarizes this work's contribution and discusses possible extensions.

\section{Methodology}
\label{sec:method}
\subsection{Problem formulation}
We investigate the dynamics of complex spatiotemporal systems governed by a set of nonlinear ordinary or partial differential equations.
The equations are discretized through  a high dimensional representation and numerically  advanced in time according to 
\begin{equation}
	\sbm_{t+\delta t} = \Fcal(\sbm_t),
	\label{eqn:micro_dynamics}
\end{equation}
where $\sbm_t\in\Rbb^{d_s}$ is the high-dimensional ($d_s \gg 1$)  \textit{micro} state vector at time $t$ and $\Fcal:\Rbb^{d_s}\rightarrow \Rbb^{d_s}$ denotes a \textit{micro propagator} (for example a numerical simulator) that advances the state in time by $\delta t$. 

The system dynamics can be produced by a legacy code or any other simulation software, but the main premise is that
\eqref{eqn:micro_dynamics} is computationally demanding to simulate for an extended time horizon. We assume that the state of the system in \eqref{eqn:micro_dynamics} can be well-approximated for a short period of time by lower-dimensional macro states $\zbm_t\in \Mcal_z\subset\Rbb^{d_z}$ such that $d_z\ll d_s$. The macro space can be identified by employing various types of autoencoders \cite{baldi2021deep}. Here we deploy a \textit{non-trainable encoder} $\Ecal:\Rbb^{d_s}\rightarrow\Rbb^{d_z}$ which maps micro states $\sbm_t$ to macro states $\zbm_t$
\begin{equation}
	\zbm_t = \Ecal(\sbm_t).
\end{equation}
To map the macro state $\zbm_t$ to the micro state $\sbm_t$, a decoder based upon \textit{the probabilistic diffusion model} is introduced,
\begin{equation}
	\sbm_t = \Dcal^{\thetabold_D}(\zbm_t,\boldsymbol{\epsilon}), 
\end{equation}
where $\Dcal^{\thetabold_D}:\Rbb^{d_z}\times \Rbb^{d_s} \rightarrow \Rbb^{d_s}$ represents the diffusion model which implicitly learns how to transform a vector of white noise $\boldsymbol{\epsilon} \in \Rbb^{d_s}$ into a micro state given the macro state $\zbm_t$. The diffusion model is parameterized by trainable parameters $\thetabold_D$. The encoder and decoder are detailed in Section~\ref{sec:encoder_decoder}.

\begin{figure}[H]
\includegraphics[width = 1\textwidth]{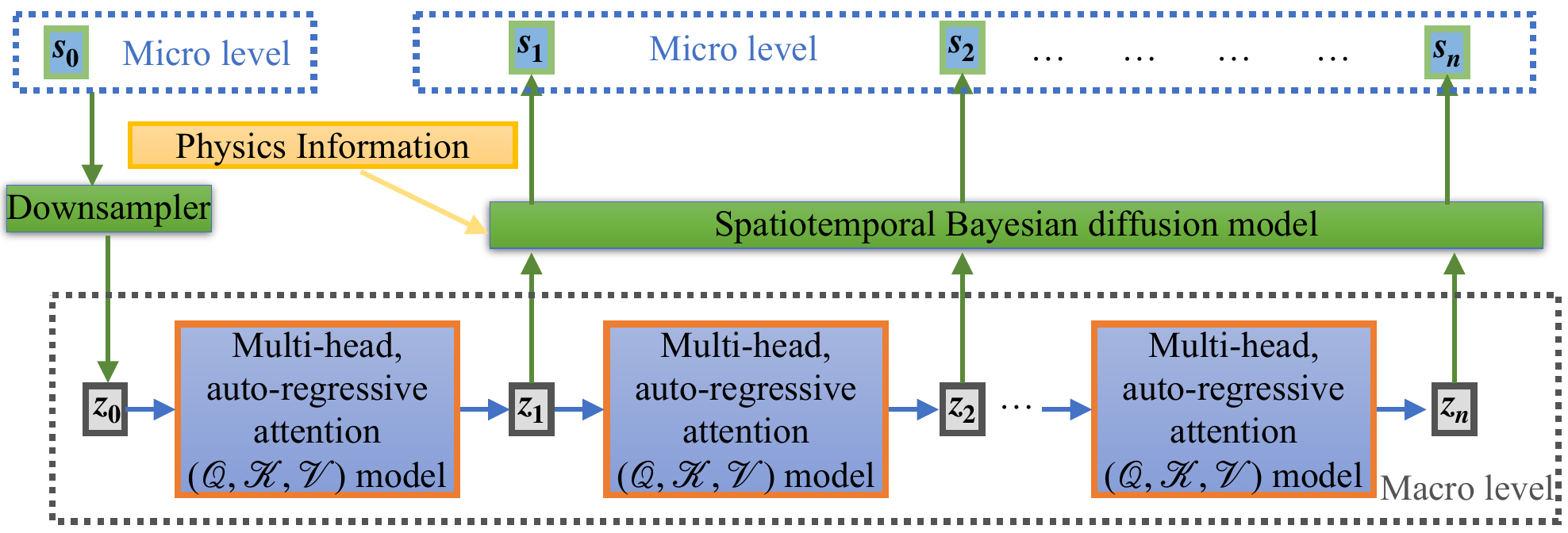}
\caption{Summary of G-LED (Generative Learning for the effective dynamics): The initial condition, a high-dimensional micro state, is encoded with a non-trainable downsampler to its corresponding macro state. Then, the dynamics are evolved in the low-dimensional space by a multi-head, auto-regressive attention model. All obtained macro states are decoded to their micro states by the spatiotemporal Bayesian diffusion model using the reverse process of the diffusion model and potentially additional physical information.}
\label{fig:diagram}
\end{figure}

Correspondingly, a non-linear macro propagator $\Acal^{\thetabold_A}$ built upon the attention model with trainable parameter $\thetabold_A$ which captures non-Markovian effect is applied to evolve the system dynamics in the macro scale
\begin{equation}
	\Acal^{\thetabold_A}:
	\begin{cases}
	 (\zbm_{0},\zbm_{1\Delta t},...,\zbm_{(n_t - 1)\Delta t})\mapsto\zbm_{n_t\Delta t} \quad\text{if }n_t  \leq N_t,\\
	(\zbm_{(n_t - N_t)\Delta t},\zbm_{(n_t - N_t+1)\Delta t}...,\zbm_{(n_t-1)\Delta t})\mapsto\zbm_{n_t\Delta t} \quad\text{if }n_t > N_t,
	\end{cases}
\end{equation}
where $\Delta t$ ($\Delta t \gg \delta t$) is the time step of the macro propagator, and $N_t$ is the maximum number of the time steps the macro propagator can look back at. Section~\ref{sec:temporal_model} provides a detailed explanation of the macro propagator. In Figure~\ref{fig:diagram}, we present a summary of the G-LED approach. 

We remark that the present approach differs in several key elements with the LED \cite{vlachas2022multiscale} and Ada-LED \cite{kivcic2023adaptive} frameworks. Here, we employ a non-trainable encoder, generative diffusion models for downsampling/lifting, and transformers with attention mechanisms for advancing the latent space dynamics. Moreover the present algorithm does not iterate between high and low dimensional spaces (as in LED and AdaLED) but the generative models are trained to forecast the statistics of the system dynamics. The advantages and drawbacks of G-LED over previous methodologies are  discussed along with its applications.

\subsection{Restricting and Lifting Operations}
\label{sec:encoder_decoder}
The transfer of information between the high-dimensional data and their latent representation is a key component of the method.
For the restriction operation, we deploy a linear interpolation to map a microstate $\sbm_t$ to a lower dimensional macro state $\zbm_t$. This enables decoupling the training of the decoder and the attention model. We note that this operation can also be performed by subsampling or averaging.

Lifting from the latent space dynamics to the high dimensional space is performed by constructing a generative decoder starting with a micro state $\sbm_t$. A diffusion process introduces a series of noised variables $\boldsymbol{\epsilon}_{i}\in\Rbb^{d_s}$:
\begin{equation}
\sbm_t \mapsto \boldsymbol{\epsilon}_1\mapsto\dots  \mapsto\boldsymbol{\epsilon}_{N_\epsilon-1}\mapsto \boldsymbol{\epsilon}_{N_\epsilon} \quad\text{for }0<i\leq N_{\epsilon}.
\end{equation}
These variables are sampled from
\begin{equation}
\boldsymbol{\epsilon}_{i}\sim q_i(\boldsymbol{\epsilon}_i|\sbm_t)\coloneqq\Ncal(\sbm_t,\sigma^2_i\Ibm) \quad\text{for }0<i\leq N_{\epsilon},
\label{eqn:forward_process}
\end{equation}
where $\sigma_i$ are strictly monotonically increasing such that $q_{N_\epsilon}\approx \Ncal(\mathbf{0},\sigma_{N_\epsilon}^2\Ibm)$ with $\sigma_{N_\epsilon} \gg ||\sbm_t||_2$ \cite{kingma2021variational}. This  diffusion process is also known as variance-exploding diffusion process \cite{song2020improved} and has the remarkable property that the reverse process, conditioned on the micro state $\sbm_t$, is  tractable \cite{ho2020denoising, song2020denoising, kingma2021variational}
\begin{equation}
	p(\boldsymbol{\epsilon}_i         |\sbm_t,  \boldsymbol{\epsilon}_{i+1})=\Ncal\Biggl(
\frac{\sigma^2_{i+1}  -  \sigma^2_i}{\sigma^2_{i+1} }	\sbm_t +  
\frac{\sigma^2_i}{\sigma^2_{i+1}}\boldsymbol{\epsilon}_{i+1}, 
\frac{ (\sigma^2_{i+1}   -   \sigma^2_i)\sigma^2_i     }{\sigma^2_{i+1}}
	\Biggr).
	\label{eqn:reverse_process}
\end{equation}
We now use the macro state $\zbm_t$ to forecast the microstates $\sbm_t$. This reverse process is approximated \eqref{eqn:reverse_process} with a deep neural network (DNN):
\begin{equation}
	p(\boldsymbol{\epsilon}_i         |\sbm_t,  \boldsymbol{\epsilon}_{i+1})
	\approx
	p_{\thetabold_D}(\boldsymbol{\epsilon}_i|\zbm_t,  \boldsymbol{\epsilon}_{i+1},i) =   
	\Ncal\Biggl(
	\frac{\sigma^2_{i+1}  -  \sigma^2_i}{\sigma^2_{i+1} }	\hat{\sbm}_{\thetabold_D}(\zbm_t,\boldsymbol{\epsilon}_{i+1},i)+  
	\frac{\sigma^2_i}{\sigma^2_{i+1}}\boldsymbol{\epsilon}_{i+1}, 
	\frac{ (\sigma^2_{i+1}   -   \sigma^2_i)\sigma^2_i     }{\sigma^2_{i+1}}
	\Biggr),
	\label{eqn:approx_rev}   
\end{equation}
Here, $\hat{\sbm}_{\thetabold_D}:\Rbb^{d_z}\times\Rbb^{d_s}\times \Nbb \rightarrow \Rbb^{d_s} $ is the DNN which is parameterized by the trainable weights $\thetabold_D$. In more detail, we employ an U-Net architecture as proposed by \cite{saharia2022photorealistic}.  By minimizing the Kullback-Leibler (KL) divergence between \eqref{eqn:reverse_process} and \eqref{eqn:approx_rev}, the training of the DNN is expressed as:
\begin{equation}
	\thetabold_D^* = \argoptunc{\thetabold_D}{
		\sum_{\sbm_t\in\Scal_\mathrm{train}}
		\Ebb_{i\sim\Ucal(1,N_\epsilon) , \boldsymbol{\epsilon}_i\sim q_i(\boldsymbol{\epsilon}_i|\sbm_t)}
		[|| \hat{\sbm}_{\thetabold_D}(\zbm_t,\boldsymbol{\epsilon}_{i},i) - \sbm_t ||_2],
	}
\end{equation}
where a set of micro states $\Scal_\mathrm{train}$ has been used as the training set. Once the model is trained, approximations to $\sbm_t$ can be obtained by using
\begin{equation}
	\sbm_t \approx \hat{\sbm}_{\thetabold^*_D}(\zbm_t,\boldsymbol{\epsilon}_{N_{\epsilon}},N_{\epsilon}) \quad\text{with } \boldsymbol{\epsilon}_{N_{\epsilon}} \sim q_{N_\epsilon},
	\label{eqn:approximation}
\end{equation} 
and iteratively applying the reverse process in \eqref{eqn:approx_rev} for decoding.

\subsubsection{Incorporation of physical information}
The lifting process (from low to high dimensions) is cast within the context of the score-based generative modeling \cite{song2020score}. The gradient of the log probability of $\boldsymbol{\epsilon_i}$ can be approximated using the learned function $\hat{\sbm}_{\thetabold^*_D}$
\begin{equation}
	\nabla_{\boldsymbol{\epsilon}}\log p(\boldsymbol{\epsilon}_i)  \approx  \frac{   \hat{\sbm}_{\thetabold^*_D}(\zbm_t,\boldsymbol{\epsilon}_i,i) -  \boldsymbol{\epsilon}_i     }{   \sigma_i^2      }.
	\label{eqn:score_based}
\end{equation}
Moreover, we can inject physical information or prior knowledge (e.g., statistical properties of the turbulent flow) or other constraints associated with the microstates, which are expressed as a vector of residuals $\boldsymbol{R}(\Hcal(\sbm_t))$.  Here, $\Hcal:\Rbb^{d_s}\rightarrow\Rbb^{d_R}$   maps the micro state to such physical information. The residuals are assumed to be virtually observed \cite{kaltenbach2020incorporating} with $\hat{\boldsymbol{R}}=0$ and virtual likelihood:
\begin{equation}
		p(\hat{\boldsymbol{R}}=0 | \sbm_t,\sigma_{R})= \Ncal(0 | \boldsymbol{R}(\Hcal(\sbm_t)),\sigma^2_R\Ibm).
	\label{eqn:guidence_assumption}
\end{equation}
The parameter $\sigma_R$ corresponds to the intensity of the enforcement of the virtual observations and the residuals. The incorporation of physical information in the diffusion model is subsequently formulated using Bayes' law:
\begin{equation}
p(\boldsymbol{\epsilon}_i|\hat{\boldsymbol{R}}=0)=\frac{p(\hat{\boldsymbol{R}}=0|\boldsymbol{\epsilon}_i) p(\boldsymbol{\epsilon}_i)}{C},
\end{equation}
where $C$ is the normalizing constant. By taking the gradient of log-likelihood on both sides, we obtain
\begin{equation}
\nabla_{\boldsymbol{\epsilon}}\log p(\boldsymbol{\epsilon}_i|\hat{\boldsymbol{R}}=0) = \nabla_{\boldsymbol{\epsilon}} \log p(\hat{\boldsymbol{R}}=0|\boldsymbol{\epsilon}_i) + \nabla_{\boldsymbol{\epsilon}}\log p(\boldsymbol{\epsilon}_i),
\label{eqn:log_bayes}
\end{equation}
where the second term on the right-hand side can be approximated using equation \eqref{eqn:score_based}. Substituting equation \eqref{eqn:guidence_assumption} and $\sbm_t=\hat{\sbm}_{\thetabold_D}(\zbm_t,\boldsymbol{\epsilon}_{i+1},i)$ into $\nabla_{\boldsymbol{\epsilon}} \log(\psibold|\boldsymbol{\epsilon}_i)$, it is trivial to show the first term in \eqref{eqn:log_bayes} is 
\begin{equation}
\nabla_{\boldsymbol{\epsilon}} \log p(\hat{\boldsymbol{R}}=0|\boldsymbol{\epsilon}_i) = -\frac{1}{\sigma^2_R}\nabla_{\boldsymbol{\epsilon}}|| \boldsymbol{R}(\Hcal(\sbm_t)) ||_2 \approx -\frac{1}{\sigma^2_R}\nabla_{\boldsymbol{\epsilon}}|| \boldsymbol{R}(\Hcal(\hat{\sbm}_{\thetabold^*_D}(\zbm_t,\boldsymbol{\epsilon}_{i},i) )) ||_2,
\end{equation}
where $\nabla_{\boldsymbol{\epsilon}}|| \boldsymbol{R}(\Hcal(\hat{\sbm}_{\thetabold^*_D}(\zbm_t,\boldsymbol{\epsilon}_{i},i) )) ||_2$ can be computed via auto differentiation.

To compute the effect of the changed gradient on the reverse diffusion process, the unit vector $\nbm_{\thetabold^*_D}(\zbm_t,\boldsymbol{\epsilon}_i)  \coloneqq \frac{\nabla_{\boldsymbol{\epsilon}}||  \boldsymbol{R}(\Hcal(\hat{\sbm}_{\thetabold^*_D}(\zbm_t,\boldsymbol{\epsilon}_{i},i) )) ||_2}{||  \nabla_{\boldsymbol{\epsilon}}||  \boldsymbol{R}(\Hcal(\hat{\sbm}_{\thetabold^*_D}(\zbm_t,\boldsymbol{\epsilon}_{i},i) )) ||_2 ||_2} $ that lies in the direction of $\nabla_{\boldsymbol{\epsilon}} \log p(\boldsymbol{\epsilon}_i|\hat{\boldsymbol{R}}=0|\boldsymbol{\epsilon}_i)$ is defined. With this definition the reverse update can be formulated as the sequential sampling procedure \cite{gao2023bayesian}  
\begin{equation}
\boldsymbol{\epsilon}_{i} \sim\Ncal\Biggl(
\frac{\sigma^2_{i+1}  -  \sigma^2_i}{\sigma^2_{i+1} }	\hat{\sbm}_{\thetabold^*_D}(\zbm_t,\boldsymbol{\epsilon}_{i+1},i)+  
\frac{\sigma^2_i}{\sigma^2_{i+1}}\boldsymbol{\epsilon}_{i+1} + \iotabold_{\thetabold^*_D}(\zbm_t,\boldsymbol{\epsilon}_{i+1}) , 
\frac{ (\sigma^2_{i+1}   -   \sigma^2_i)\sigma^2_i     }{\sigma^2_{i+1}}
\Biggr),
\label{eqn:guide_reverse}
\end{equation} 
where $\iotabold_{\thetabold^*_D}(\zbm_t,\boldsymbol{\epsilon}_{i+1})$ is defined as 
\begin{equation}
\iotabold_{\thetabold^*_D}(\psibold,\zbm_t,\boldsymbol{\epsilon}_{i+1})\coloneqq  \beta_{\mathrm{guide}} \max\Biggl(\frac{ (\sigma^2_{i+1}   -   \sigma^2_i)\sigma^2_i     }{\sigma^2_{i+1}}, \sigma^2_\mathrm{guide}         \Biggr)\nbm_{\thetabold^*_D}(\zbm_t,\boldsymbol{\epsilon}_{i+1}),
\end{equation}
where $\beta_\mathrm{guide}$ and $\sigma_\mathrm{guide}$ are hyper parameters \cite{gao2023bayesian}. By iteratively calling \eqref{eqn:guide_reverse}, the final micro state can be decoded from the sequential sampling $\boldsymbol{\epsilon}_{N_\epsilon}\mapsto\boldsymbol{\epsilon}_{N_\epsilon-1}\mapsto\dots\mapsto \boldsymbol{\epsilon}_1 \mapsto \sbm_t$.

\subsection{Learning macro dynamics via the attention model}
\label{sec:temporal_model}
The attention model evolves the macro-dynamics in an auto-regressive way \cite{geneva2022transformers}. The three core functions in each employed transformer layer map a macro state $\zbm_t$ to three different hidden states
\begin{equation}
\Qcal:\zbm_t \mapsto \Qcal(\zbm_t), \quad \Kcal: \zbm_t\mapsto \Kcal(\zbm_t), \quad \Vcal:\zbm_t\mapsto\Vcal(\zbm_t),
\end{equation}
where $\Qcal:\Rbb^{d_z}\rightarrow \Rbb^{d_{QK}}$ is the query function, $\Kcal:\Rbb^{d_z}\rightarrow \Rbb^{d_{QK}}$ is the key function and $\Vcal:\Rbb^{d_z}\rightarrow\Rbb^{d_z}$ is the value function. Given a sequence of macro states $\{\zbm_{0 }, \zbm_{1\Delta t },\dots,\zbm_{(N-1)\Delta t}|N\leq N_t  \}$, the attention model predicts the next macro state
\begin{equation}
	\zbm_{N\Delta t} = \sum_{i=0}^{N-1}\frac{\exp(\Qcal(\zbm_{N\Delta t}) \cdot   \Kcal(\zbm_{i\Delta t})    ) \Vcal(\zbm_{i\Delta t})}{\sum_{i=0}^{N-1}\exp(\Qcal(\zbm_{N\Delta t}) \cdot   \Kcal(\zbm_{i\Delta t})    )   }  .
\end{equation}
During the computation, the evaluations of the key and value functions are cached within the length of $N_t$ \cite{radford2019language}. The multi-head implementation of the attention model is detailed in \cite{vaswani2017attention}. The training in this paper is performed via the standard auto-regressive method used in \cite{geneva2022transformers, sun2023unifying} 
\begin{equation}
\thetabold_A^* = \argoptunc{\thetabold_A}{
	\sum_{i=1}^{N_t}	\sum_{ \Scal \subset\Scal_\mathrm{train}}
		||\zbm_{n_t\Delta t} - \Acal^{\thetabold_A}(\zbm_{(n_t-i)\Delta t},  \dots, \zbm_{(n_t-1)\Delta t})||_2,
	}
\end{equation}
where $S=\{ \sbm_{(n_t-i)\Delta t},  \dots, \sbm_{(n_t-1)\Delta t} \}$ represents all the sequences that are shorter than $N_t$ in the training sequence $\Scal_\mathrm{train}$ and the macro states are obtained by evaluating the non-trainable encoder $\zbm = \Ecal(\sbm)$.

\section{Results}
\label{sec:result}
We examine the capabilities of  G-LED in simulations of three benchmark problems: (i) the  1D KS equation, (ii) the 2D flow past a backward-facing step, and (iii) the 3D turbulent channel flow. We compare its performance to numerical solutions of these equations and other reduced-order modeling methods.
\subsection{Kuramoto-Sivashinsky equation}
\begin{figure}[htp]
	\centering
	\includegraphics[width = 1\textwidth]{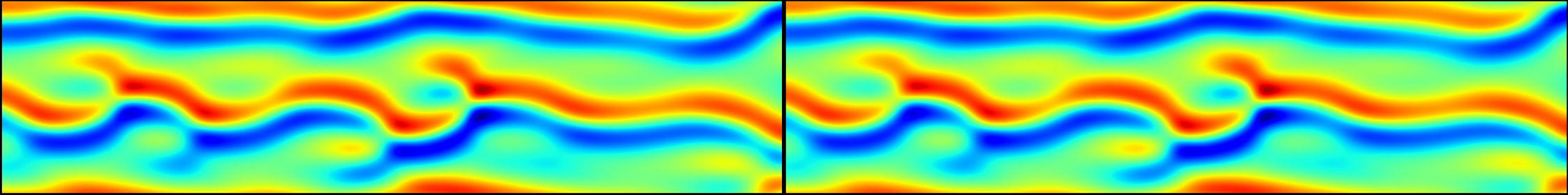}
	\vfil
	\includegraphics[width = 1\textwidth]{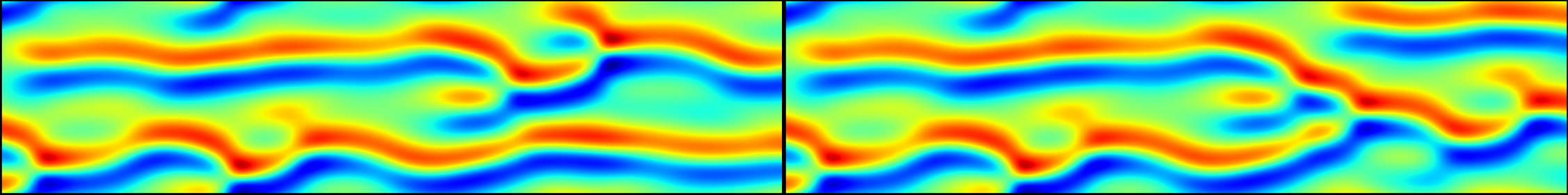}
	\vfil
	\includegraphics[width = 1\textwidth]{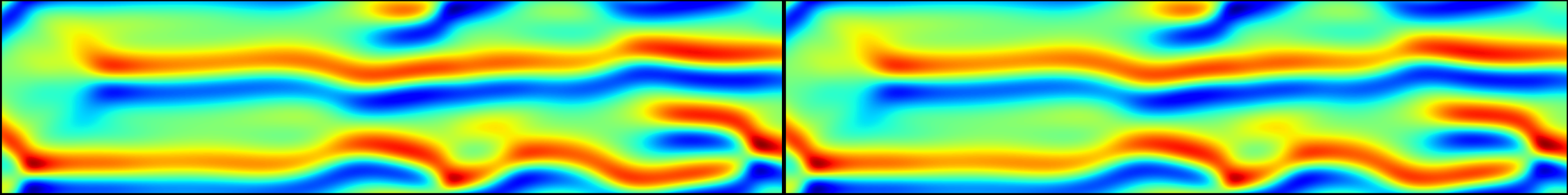}
	\colorbarMatlabJet{-3.1}{-1.55}{0}{1.55}{3.1}
\caption{Results obtained with G-LED (\textit{left}) and numerical simulation (\textit{right}) on different test trajectories (\textit{from top to bottom}). The horizontal direction depicts the time $t \in[0,100]$ and the vertical direction corresponds to the spatial coordinate $x \in[0,22]$.}
\label{fig:KS_equation_sample_contours}
\end{figure}

\begin{figure}[htp]
\centering
\includegraphics*[width=0.32\textwidth]{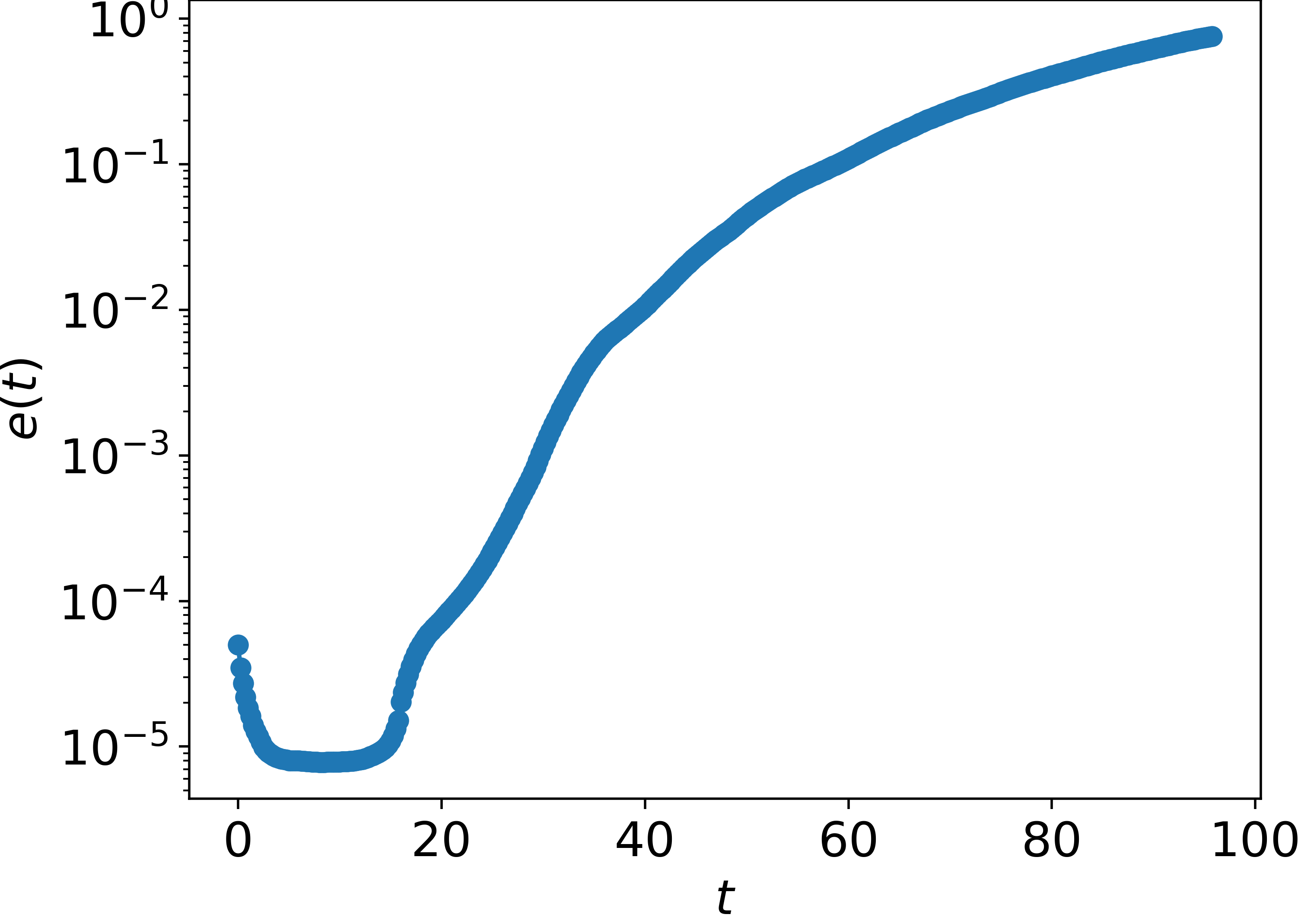}
\includegraphics*[width=0.32\textwidth]{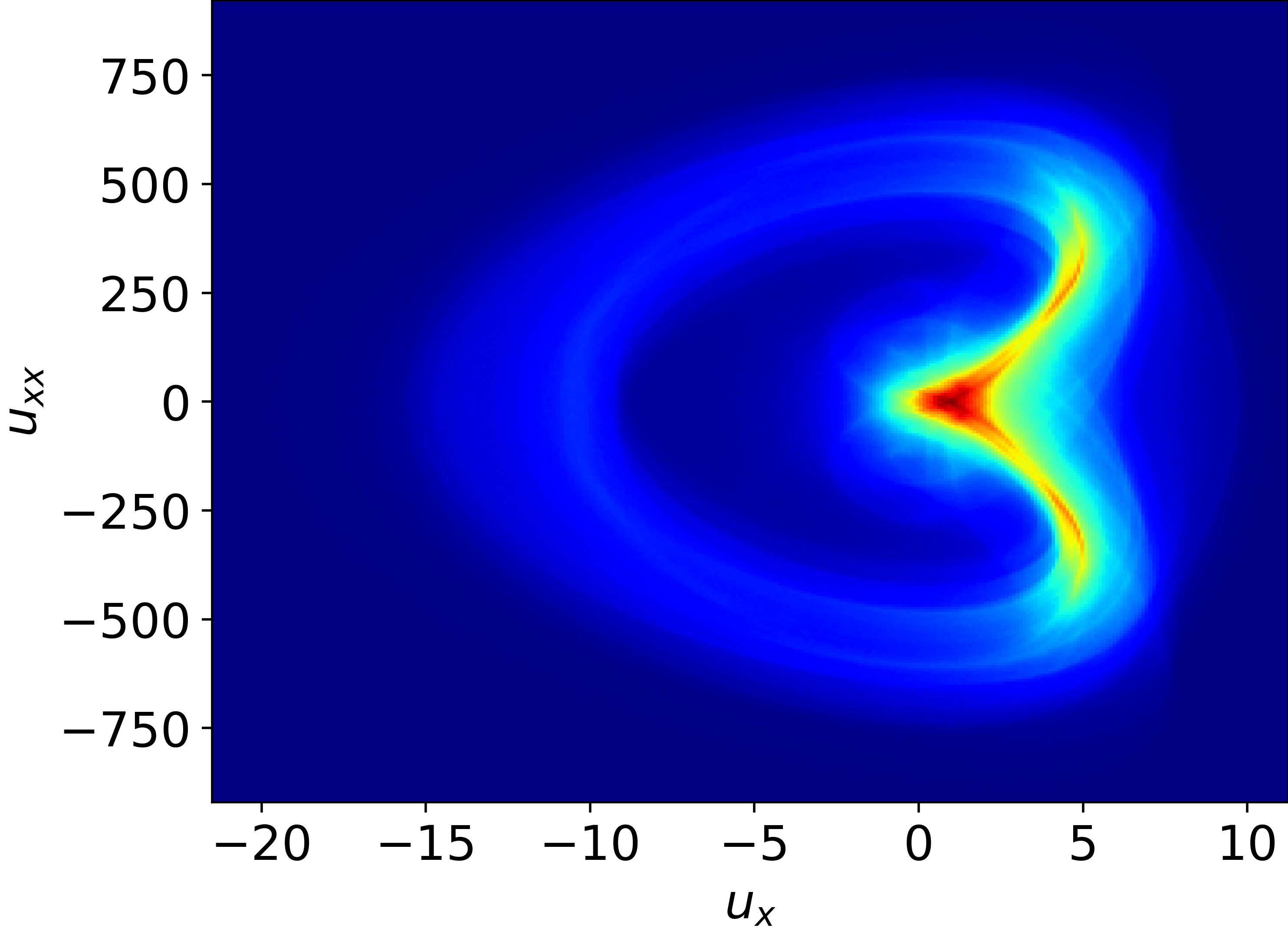}
\includegraphics*[width=0.32\textwidth]{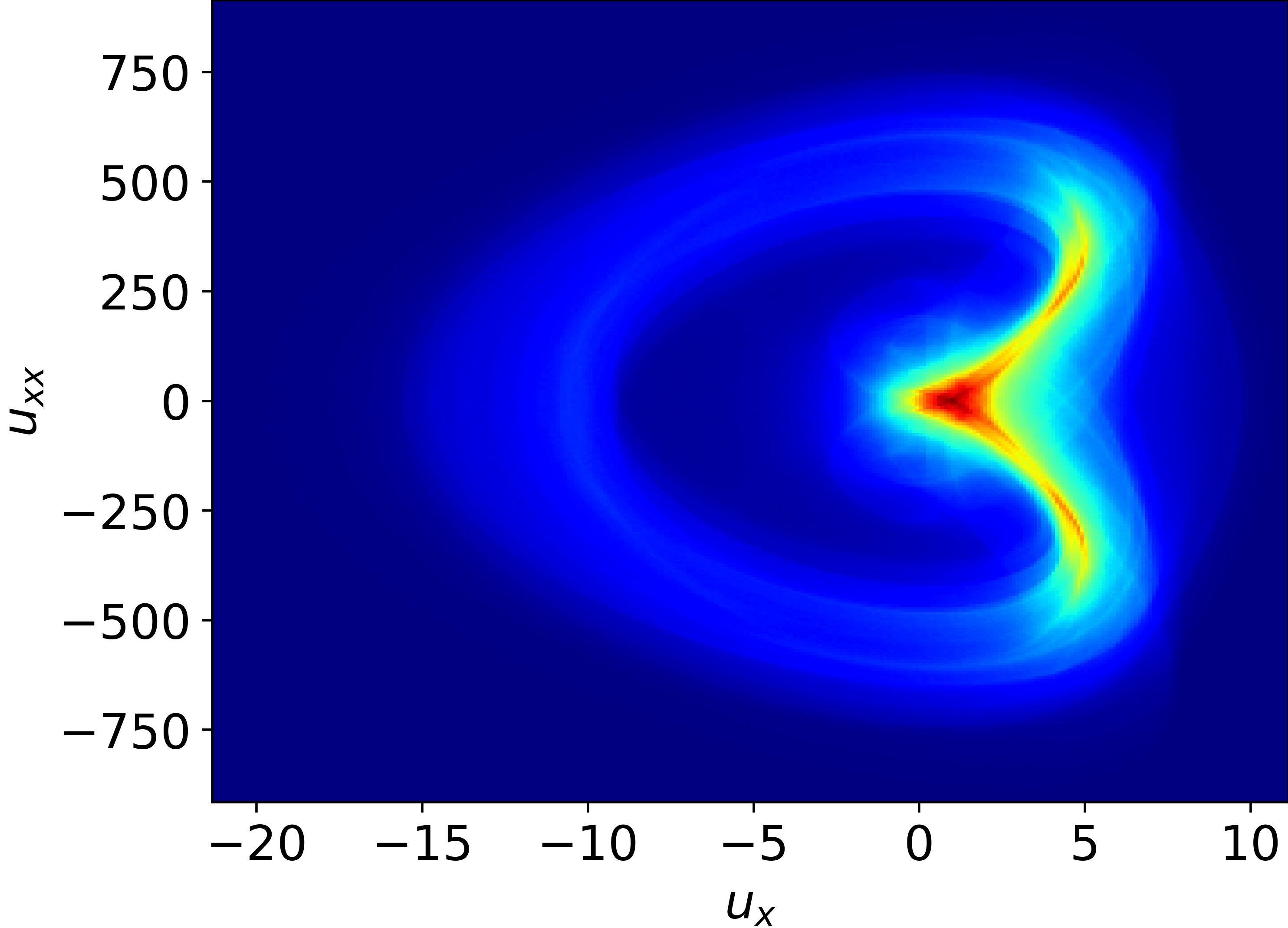}
\colorbarMatlabJet{0}{17000}{34000}{51000}{68000}
\caption{The error ($e(t)=\frac{||\sbm_t^{\mathrm{numerical}} - \sbm_t^{\mathrm{LED}}||^2_2}{||\sbm_t^{\mathrm{numerical}}||^2_2}$) over all testing sequences (\textit{left}); The density of values in
the $u_x - u_{xx}$ space computed from all testing sequences from G-LED (\textit{middle}) and numerical simulations (\textit{right}).}
\label{fig:error_ux_uxx}
\end{figure}

The G-LED  is applied to the time-dependent Kuramoto-Sivashinsky (KS) equation, 
\begin{equation}
    \pder{u}{t} = -\nu\pder{^4 u}{^4 t} - \pder{^2 u}{^2 x} - u\pder{u}{x}, 
\end{equation}
on the line $\Omega = [0, 22]$ with periodic boundary conditions $u(0, t) = u(L, t)$ and $\nu=1$. This equation is a model for a range of  non-linear phenomena and has a stable, chaotic attractor. The KS-equation has been extensively studied using classical numerical methods and machine learning algorithms \cite{cvitanovic2010state,vlachas2018,linot2020deep,linot2023stabilized}. The equation is solved with a high order finite difference scheme for stiff PDEs \cite{kassam2005fourth} and the simulation details can be found in~\ref{sec:ks_appendix}. Figure~\ref{fig:KS_equation_sample_contours} shows the predictions of several testing sequences with new initial conditions by both G-LED and the numerical simulation. G-LED predicts well the  system dynamics in accordance  with the results of the numerical method. Due to the chaotic nature of the system, an accumulation of relative error (Figure~\ref{fig:error_ux_uxx} \textit{left}) is observed over time, and the zero-mean snapshots emphasize this trend. However, the well-matched density distributions (Figure~\ref{fig:error_ux_uxx} \textit{middle and right}) demonstrate that the proposed G-LED framework can replicate the statistics of the dynamical system and capture the system's stable attractor.

\subsection{Two-dimensional flow over backward-facing step at Re=5000}
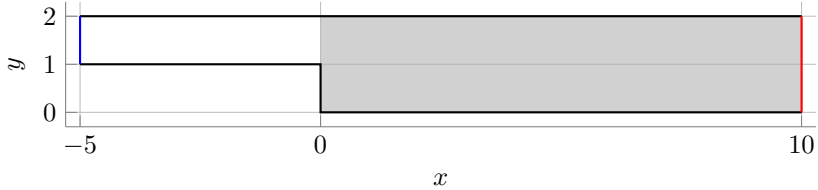
\begin{figure}[htp]
    \centering
    \begin{tikzpicture}
\begin{axis}[
axis equal image,
axis line style={gray},
axis x line*=bottom,
axis y line*=left,
width=0.7\textwidth,
xtick={-5, 0, 10},
ytick={0, 1, 2},
grid=major,
ymax=2.3,
xmax=10.3,
ylabel=$y$,
xmin=-5.3,
xlabel=$x$,
ymin=-0.3]
\addplot [opacity=0.6, fill=black!30!white, opacity=0.6, forget plot]
coordinates {
( 0.00000000e+00,  0.00000000e+00)
( 1.00000000e+01,  0.00000000e+00)
( 1.00000000e+01,  2.00000000e+00)
( 0.00000000e+00,  2.00000000e+00)};

\addplot [thick, color=black]
coordinates {
(-5.00000000e+00,  1.00000000e+00)
( 0.00000000e+00,  1.00000000e+00)
( 0.00000000e+00,  0.00000000e+00)
( 1.00000000e+01,  0.00000000e+00)};\label{line:cyl0:wall}

\addplot [thick, color=black, forget plot]
coordinates {
(-5.00000000e+00,  2.00000000e+00)
( 1.00000000e+01,  2.00000000e+00)};

\addplot [thick, color=red]
coordinates {
( 1.00000000e+01,  0.00000000e+00)
( 1.00000000e+01,  2.00000000e+00)};\label{line:cyl0:out}

\addplot [thick, color=blue]
coordinates {
(-5.00000000e+00,  1.00000000e+00)
(-5.00000000e+00,  2.00000000e+00)};\label{line:cyl0:inlet}

\end{axis}
\end{tikzpicture}
    \caption{Geometry of flow domain (solid lines), area of interest (shadow zone) for prediction, and boundary conditions of inlet (\ref{line:cyl0:inlet}), outlet (\ref{line:cyl0:out}), and no-slip wall (\ref{line:cyl0:wall}) for the case of flow over backward-facing step.}
    \label{fig:bfs-geo}
\end{figure}

We examine the performance of G-LED on the 2D viscous, incompressible flow  over a backward-facing step described by the Navier-Stokes (NS) equations:
\begin{equation}
    \nabla\cdot\ubm = 0,\quad\pder{\ubm}{t}+\nabla\cdot(\ubm\otimes\ubm) = -\frac{1}{\rho}\nabla p + \nu\Delta \ubm\quad\text{in }\Omega\subset\Rbb^{d_\Omega},
    \label{eqn:ns}
\end{equation}
where $\ubm=[u,v]:\Omega\rightarrow \Rbb^{d_\Omega}$ is the velocity of the fluid ($u$ is the streamwise velocity and $v$ is wall normal velocity), $\rho\in\Rbb_{+}$ is the density, $p:\Omega\rightarrow \Rbb$ is the pressure and $\nu\in\Rbb_{+}$ is the kinematic viscosity. The flow entails a separation region after the step (Figure~\ref{fig:bfs-geo}). The NS equations in \eqref{eqn:ns} are solved in the context of Large Eddy Simulations (LES) \cite{it:2015-014} with a finite-volume method  and subgrid-scale stress modeling based on the velocity gradient tensor \cite{nicoud1999subgrid}.  Here, we aim to learn the dynamics of this  flow at $Re=5000$ based on the inlet velocity and height. The setting of the simulations  are detailed in \ref{sec:bfs_appnedix}.

\begin{figure}[htp]
\centering
\includegraphics[width=1\textwidth]{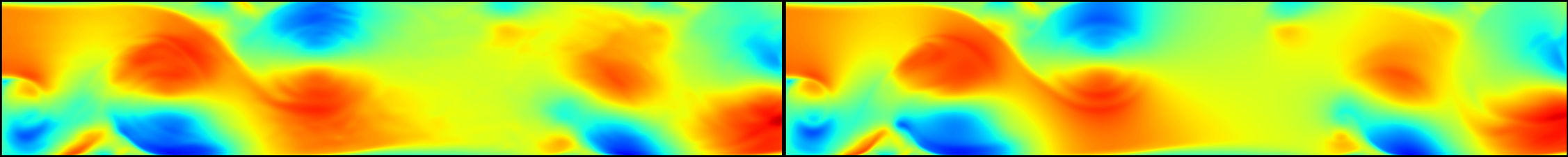}
 \vfil
 \includegraphics[width=1\textwidth]{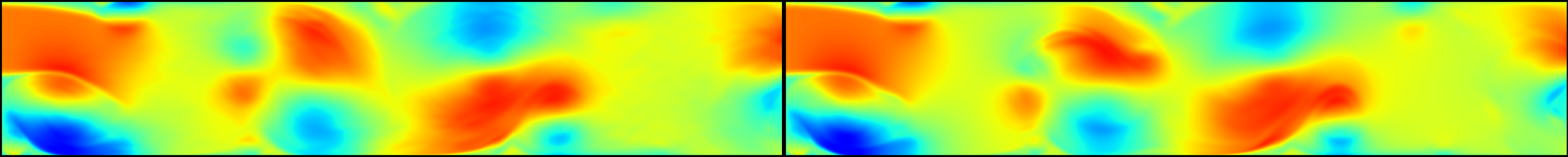}
\includegraphics[width=1\textwidth]{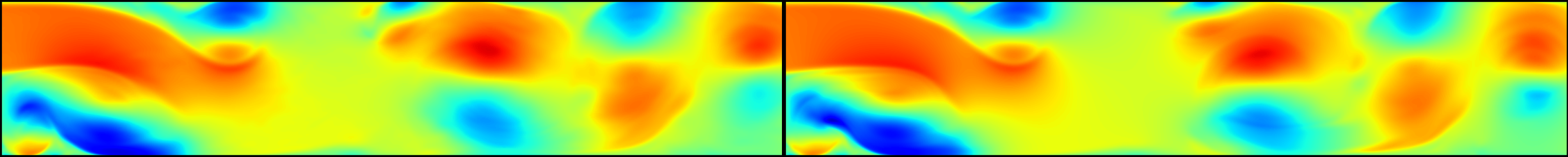}
\vfil
\includegraphics[width=1\textwidth]{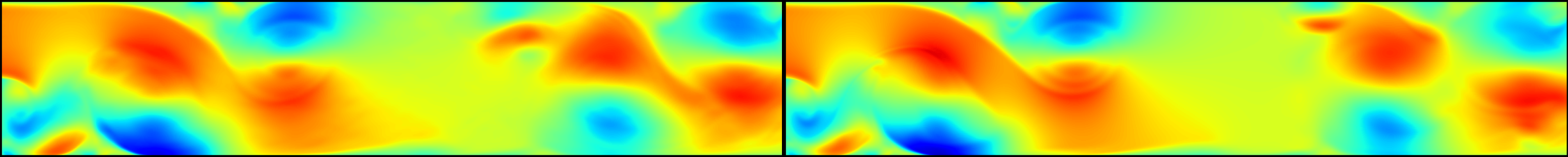}
\vfil
\colorbarMatlabJet{-13}{-5.5}{2}{9.5}{17}
\caption{The streamwise velocity rollout of G-LED (\textit{left}) and LES (\textit{right}) given the same initial state at $t=0.05, 0.50,  0.95,  1.25$ (\textit{from top to bottom})} 
\label{fig:bfs_u_contour}
\end{figure}

\begin{figure}[htp]
\centering
\includegraphics[width=0.32\textwidth]{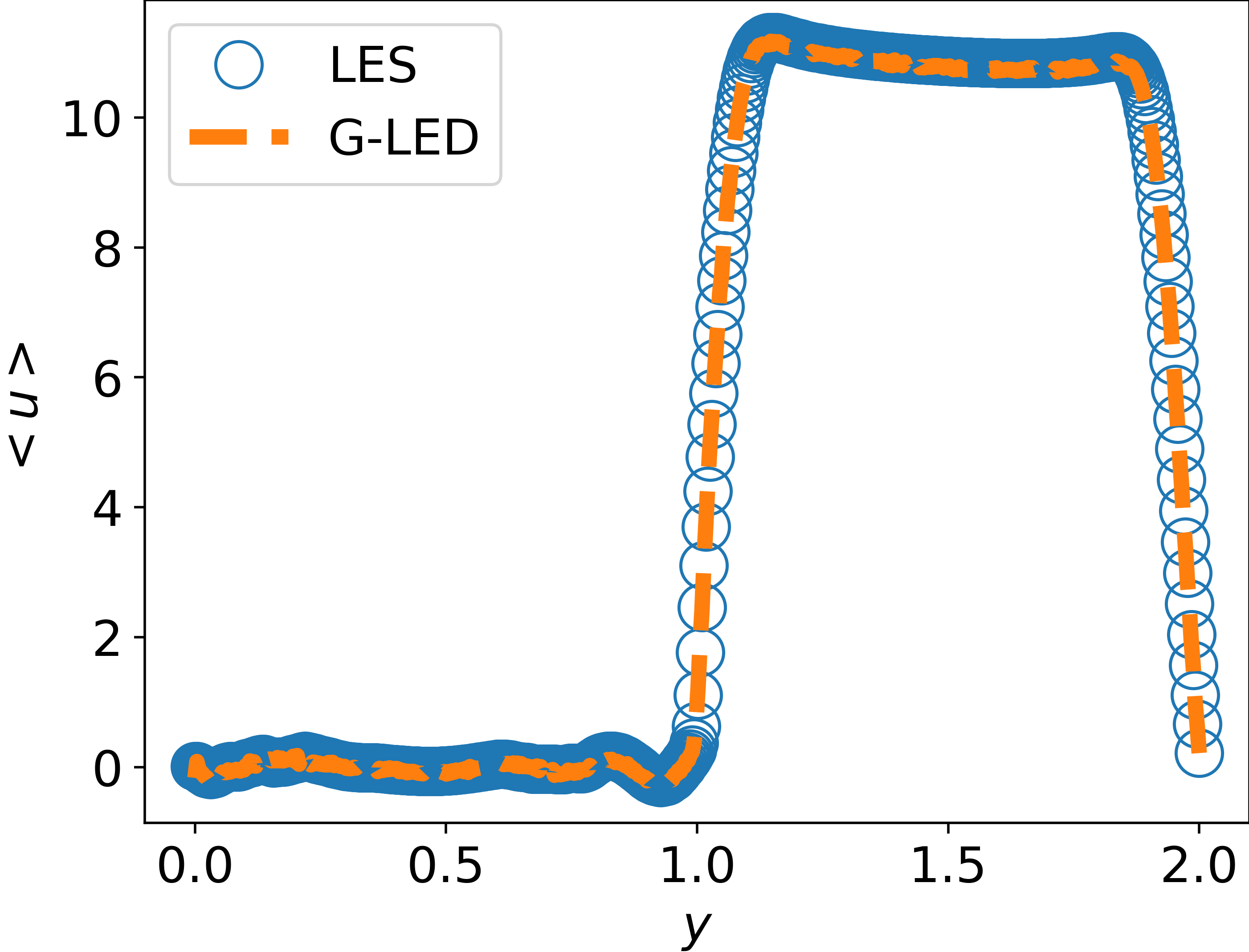}
\includegraphics[width=0.32\textwidth]{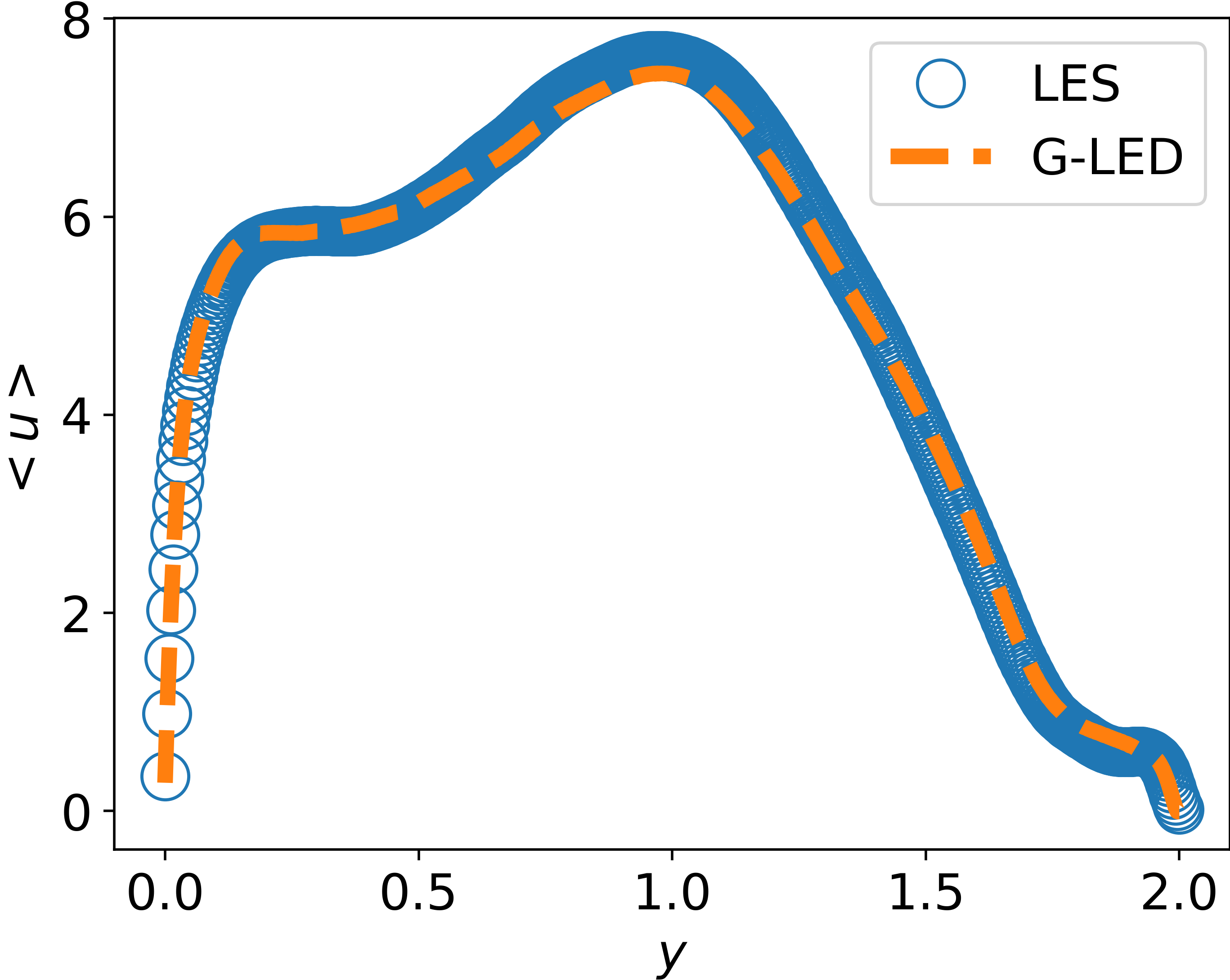}
\includegraphics[width=0.32\textwidth]{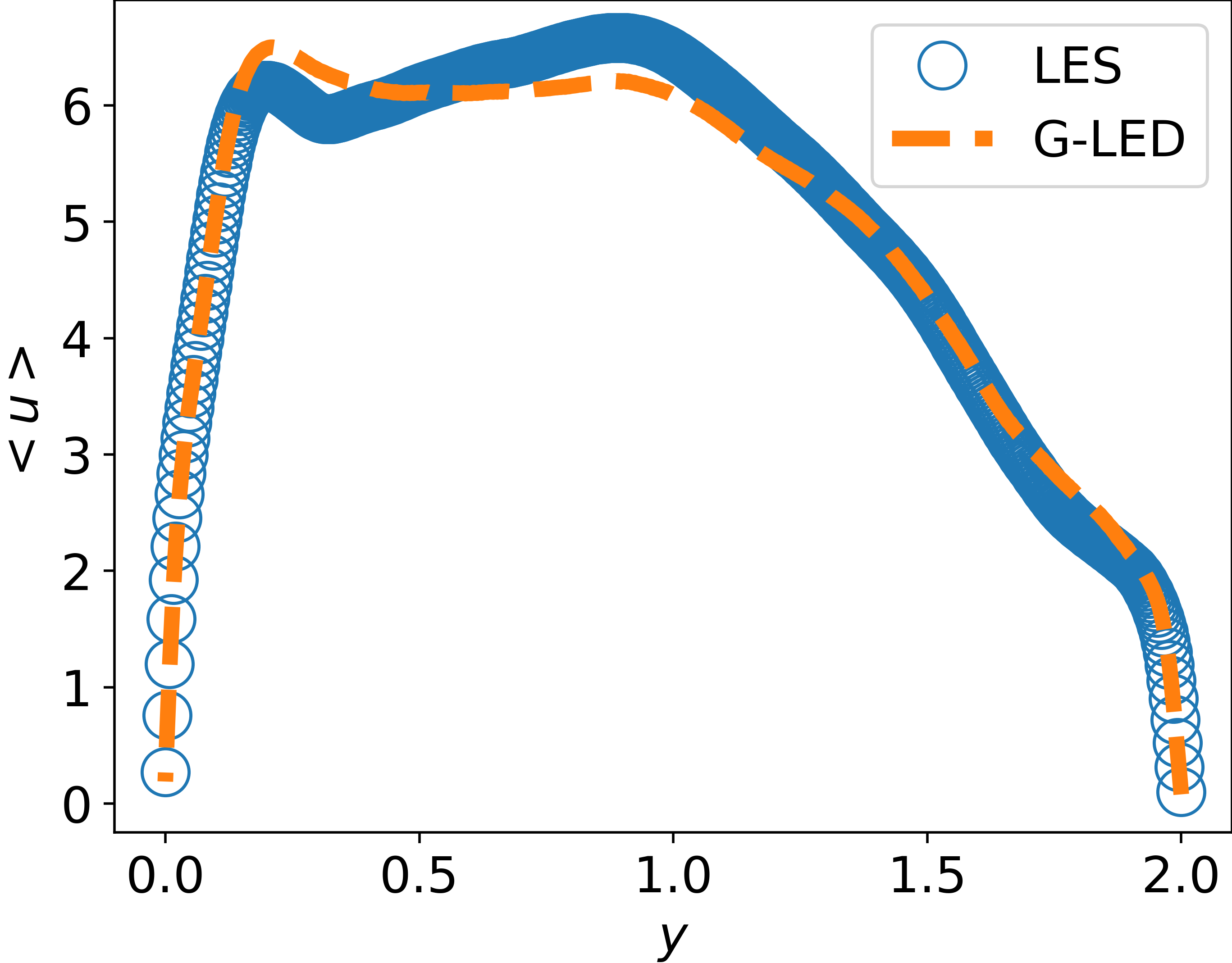}
\caption{Mean of streamwise velocity at $x=0, 4, 8$ (\textit{from left to right, top to bottom}). LED refers to the proposed G-LED framework. } 
\label{fig:stream_mean_bfs}
\end{figure}

The G-LED forecasts the dynamics in the lower-dimensional manifold (Figure~\ref{fig:bfs_u_contour} and \ref{fig:bfs_v_contour}) for one flow through time ($T_{\mathrm{flowthrough}} \approx 1$) providing a online speedup of about 5000X \footnote{LES is performed in parallel on AMD Ryzen 9 7950x 16-core processor and LES is on a single NVIDIA A100.}, given the initial state from the testing snapshots.  The G-LED evolves the dynamic of the turbulent flow in good agreement with the LES results in terms of the streamwise and wall normal velocities. Furthermore, G-LED produces accurate mean statistics. 
Figure~\ref{fig:stream_mean_bfs} shows the comparison between G-LED and LES for the mean streamwise velocity profiles. The comparison is made at nine representative vertical lines in the recirculation, reattachment, and recovery regions. We remark the excellent agreement between G-LED and LES results at all lines. 

\begin{figure}[htp]
	\centering
	\includegraphics[width=0.32\textwidth]{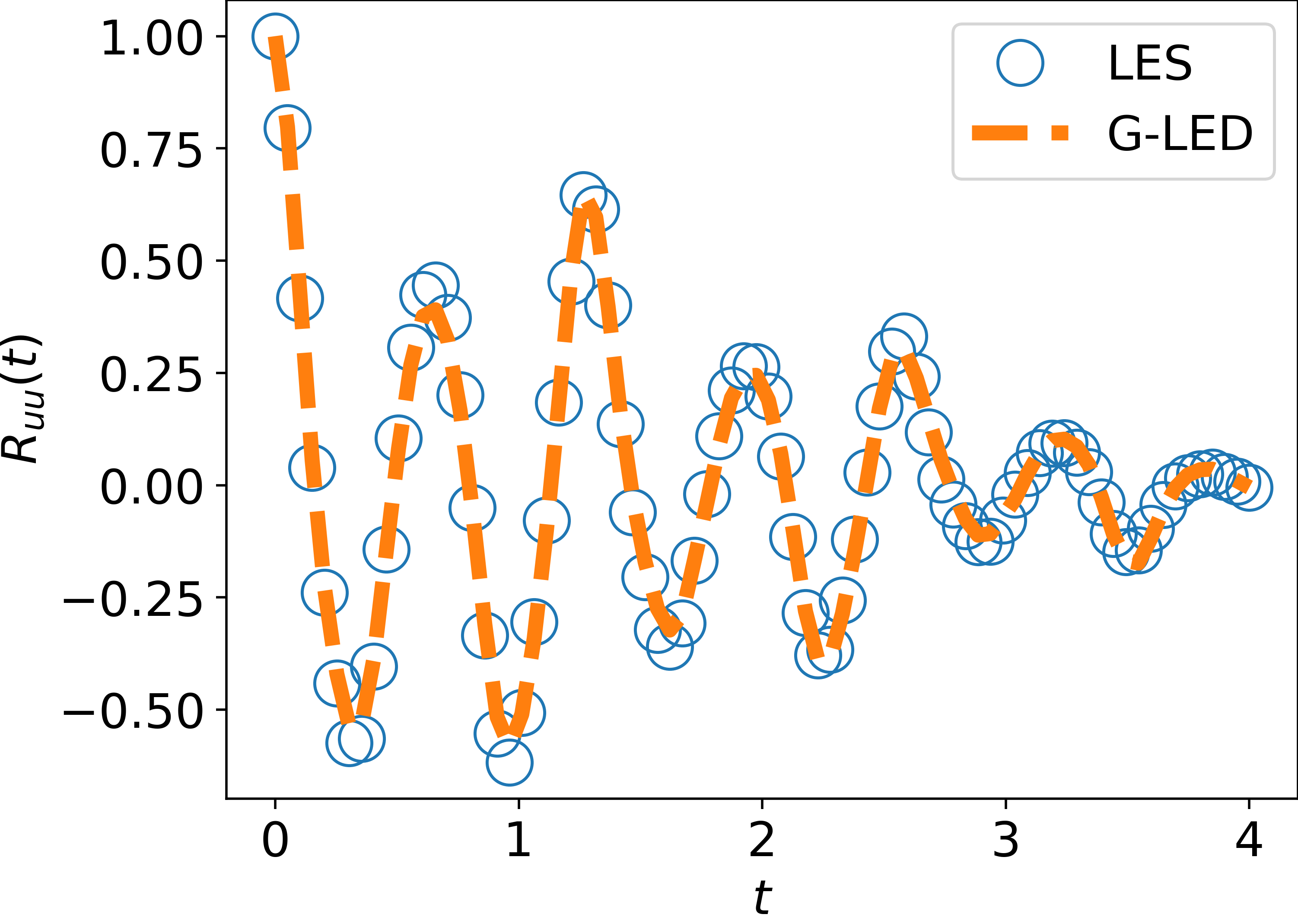}
	\includegraphics[width=0.32\textwidth]{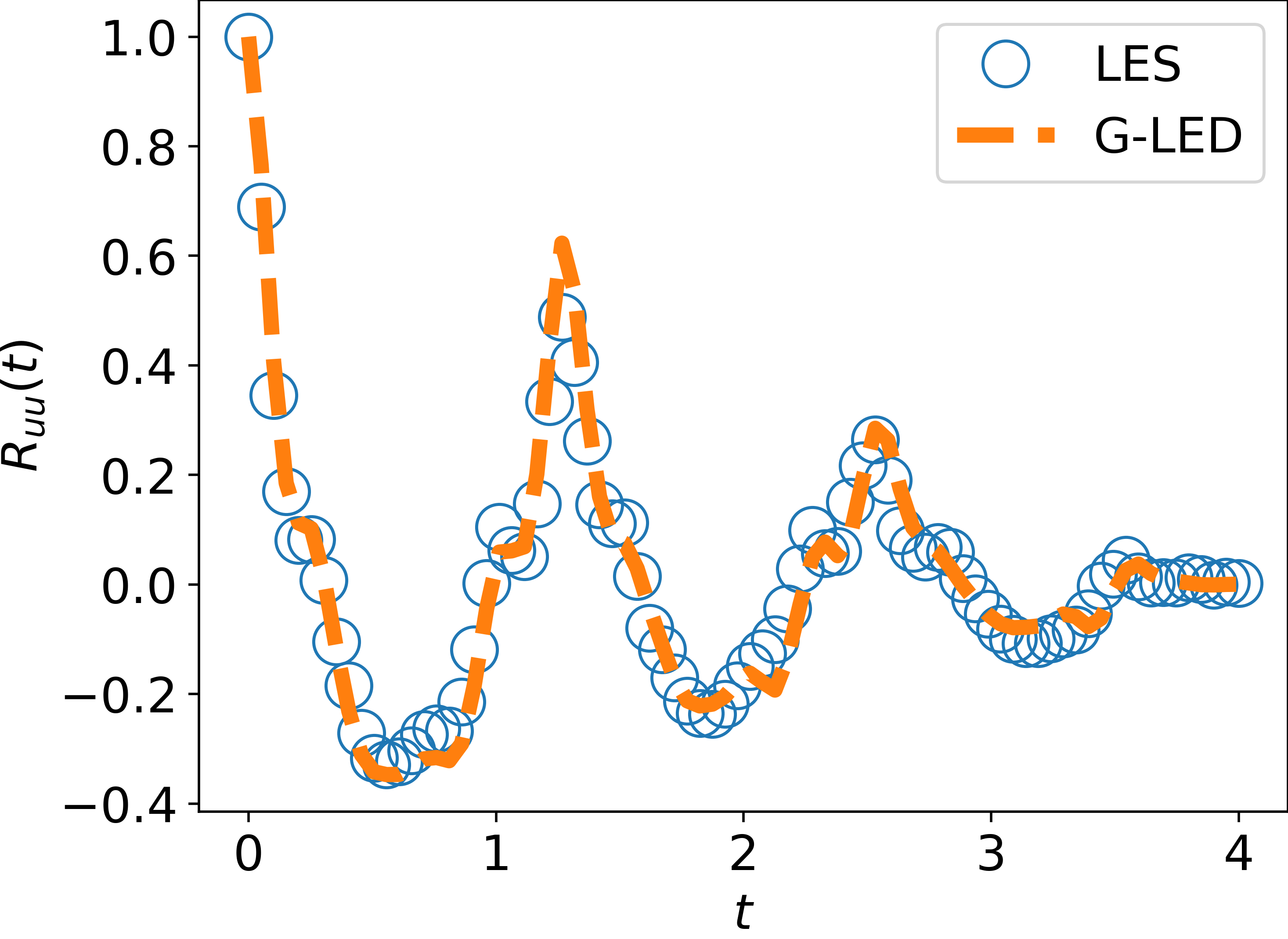}
	\includegraphics[width=0.32\textwidth]{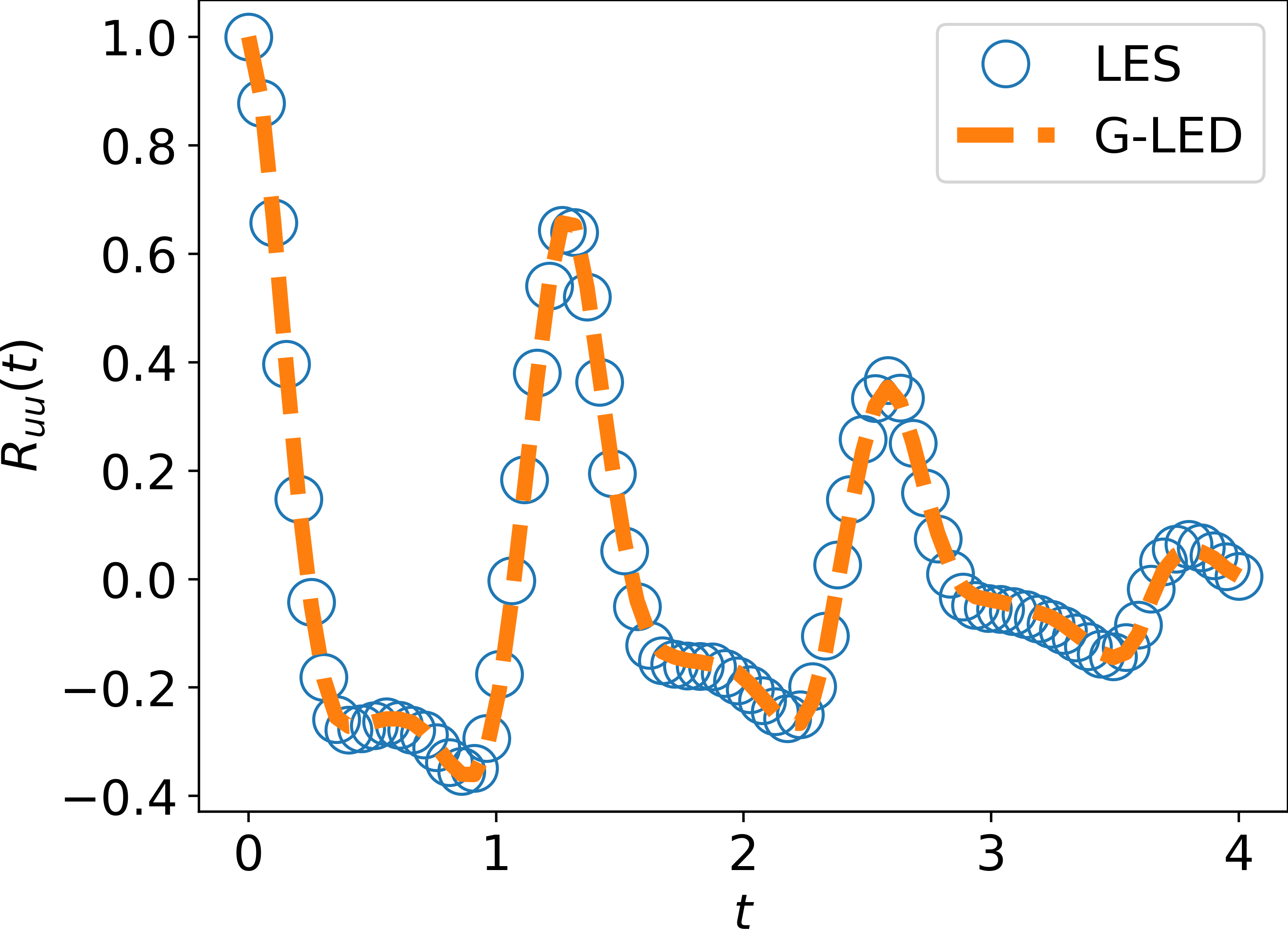}
	\caption{Temporal correlation of streamwise velocity at $(x,y)=(0.6,0.6), (0.6,1.2), (0.6,1.8)$ (\textit{from left to right, top to bottom}). } 
 \label{fig:tmpcor_u_bfs}
\end{figure}

The root mean square (rms) profiles of the streamwise and wall normal velocity fluctuations and the Reynolds shear stress component are compared for the results of  G-LED and LES at the same nine streamwise locations in Figure~\ref{fig:u_rms_bfs}, \ref{fig:v_rms_bfs} and \ref{fig:uv_rms_bfs}, respectively. The good agreement between G-LED and LES simulations  shows that G-LED can reproduce statistics for the instantaneous velocity fluctuations. Following in Figure~\ref{fig:u_energy_bfs}, \ref{fig:v_energy_bfs}, the energy spectrum at different heights is illustrated, and the overall agreement of energy cascade from the bottom to the top indicates that G-LED captures the vertices at different scales well. The spatial correlations in Figure~\ref{fig:spacor_u_bfs} and \ref{fig:spacor_v_bfs} further show that that G-LED reflects the spacing of vortex shedding well compared with LES. In the proposed G-LED framework, spatial and temporal motions at scales of micro level (LES) are learned. Therefore, the G-LED framework provides opportunities for the accurate capture of temporal statistics, demonstrated in Figure~\ref{fig:tmpcor_u_bfs} and \ref{fig:tmpcor_v_bfs}. 

\subsection{Three-dimensional turbulent channel flow}
\begin{figure}[htp]
    \centering
    \includegraphics[width=0.32\textwidth]{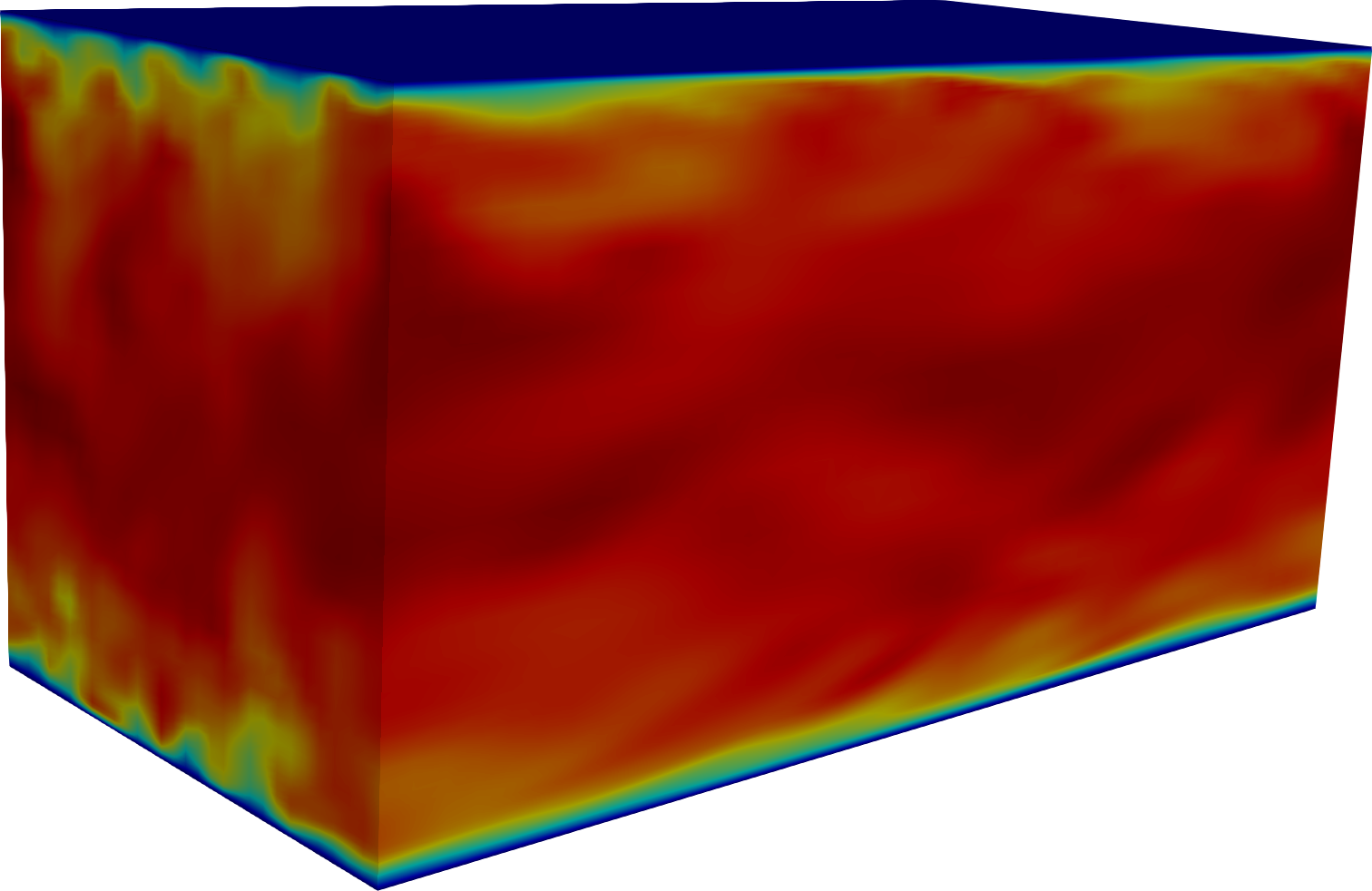}
    \includegraphics[width=0.32\textwidth]{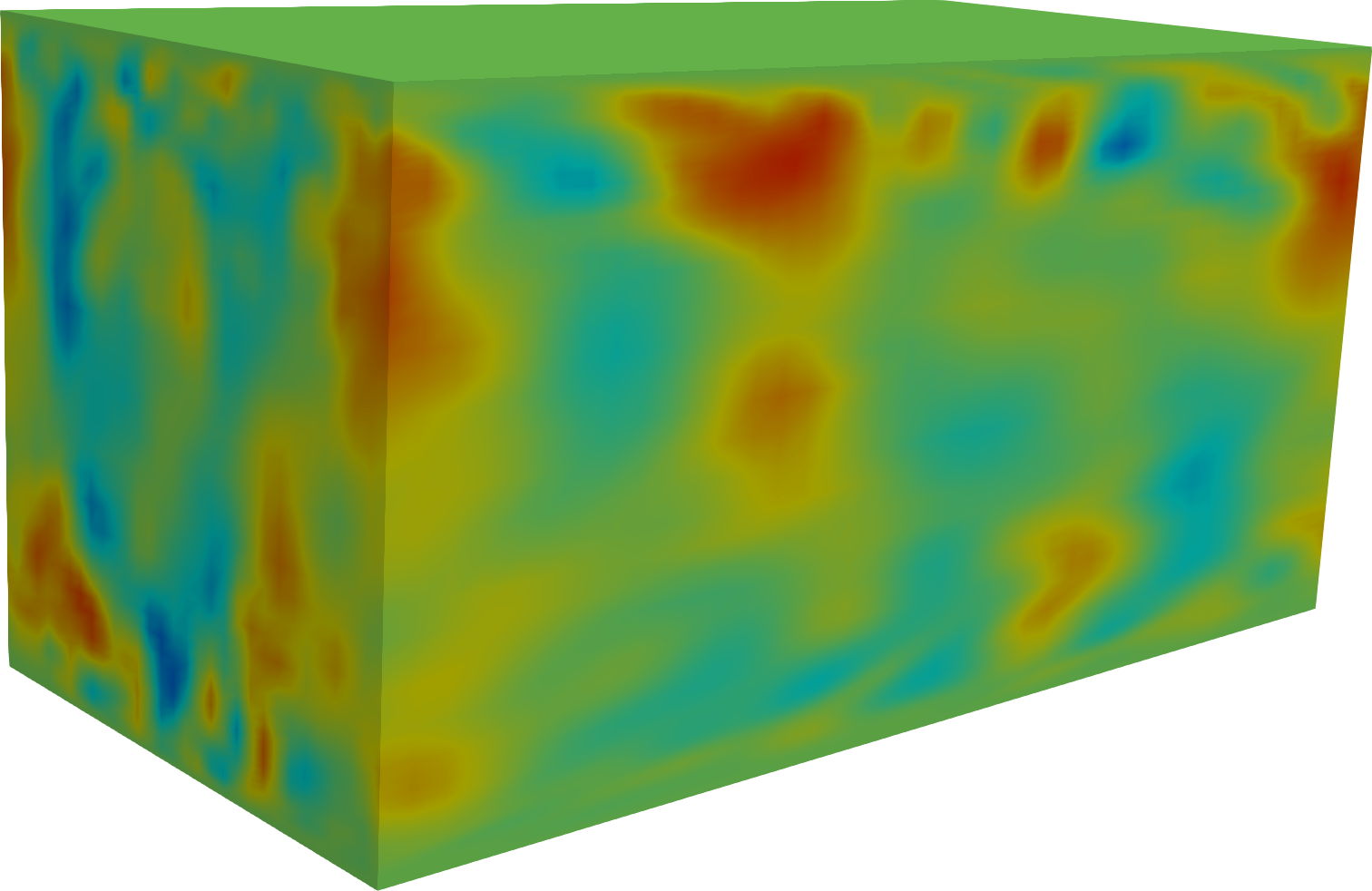}
    \includegraphics[width=0.32\textwidth]{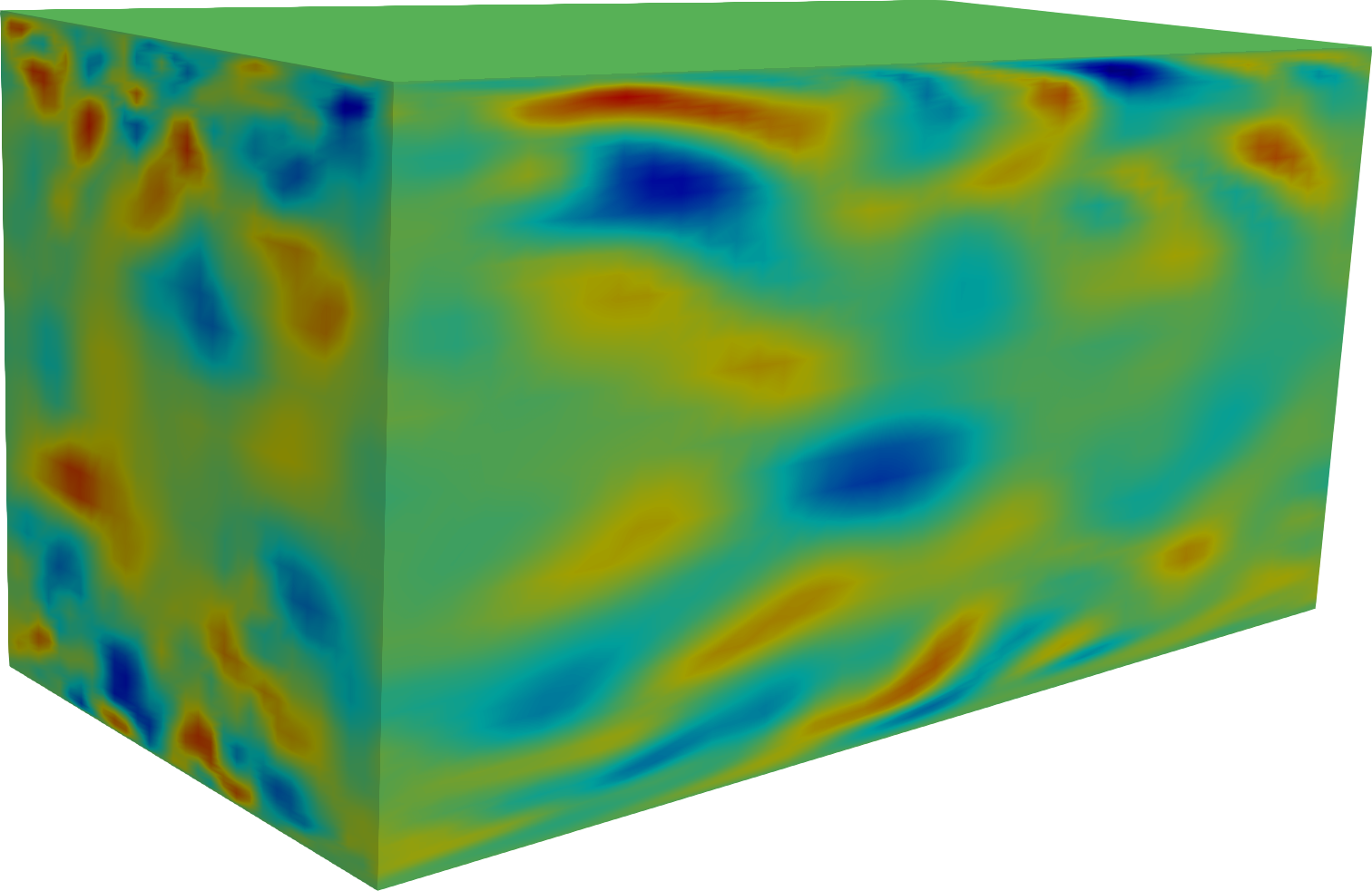}
    \caption{The overview streawise (\textit{left}), wallnormal (\textit{middle}), spanwise (\textit{right}) velocity field of channel flow at $Re_\tau = 395$. }
    \label{fig:3d_channel_overview}
\end{figure}

\begin{figure}[htp]
\centering
\includegraphics[width = 0.15\textwidth]{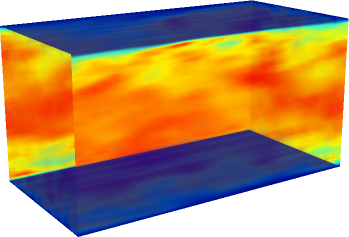}
\includegraphics[width = 0.15\textwidth]{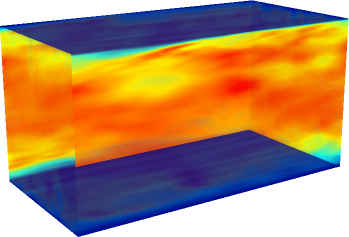}
\includegraphics[width = 0.15\textwidth]{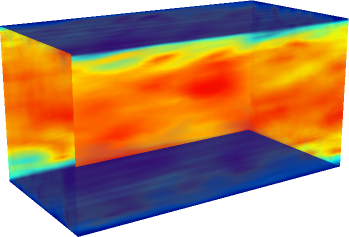}
\includegraphics[width = 0.15\textwidth]{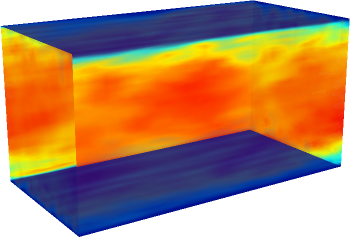}
\includegraphics[width = 0.15\textwidth]{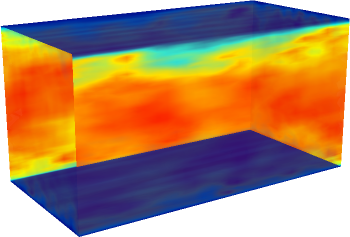}
\includegraphics[width = 0.15\textwidth]{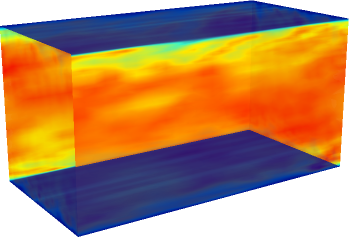}
\vfil
\includegraphics[width = 0.15\textwidth]{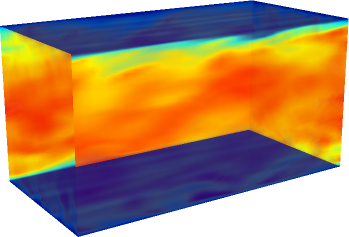}
\includegraphics[width = 0.15\textwidth]{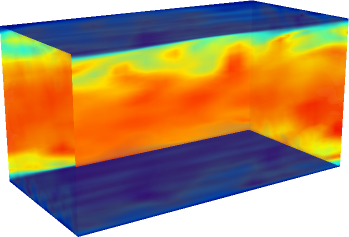}
\includegraphics[width = 0.15\textwidth]{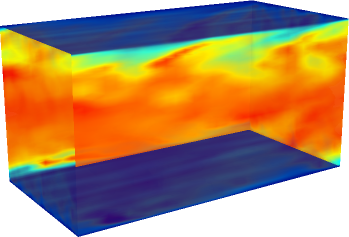}
\includegraphics[width = 0.15\textwidth]{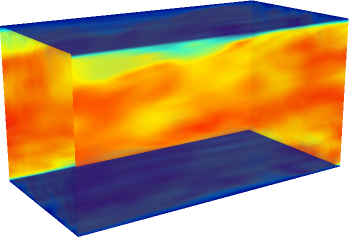}
\includegraphics[width = 0.15\textwidth]{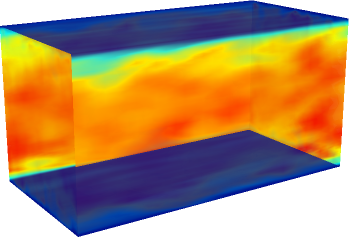}
\includegraphics[width = 0.15\textwidth]{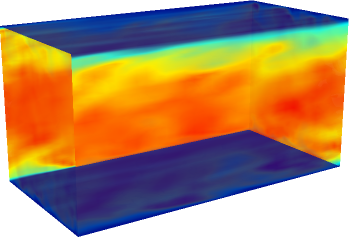}
\vfil
\colorbarMatlabJet{0}{0.05}{0.1}{0.15}{0.2}
\caption{Volume plots of streamwise velocity for the turbulent channel flow from G-LED (\textit{top row}) and LES (\textit{bottom row}) at $t=60,120,180,240,300,360$ which is equivalent to flow-through time at $t_{\mathrm{ft}}=2.27, 4.55, 6.82, 9.10, 11.3, 13.65$ (\textit{from left to right}). }
\label{fig:channel_flow_u_contour}
\end{figure}

We examine the application of G-LED in simulations of the 3D channel flow at $Re_\tau = 395$ (Figure~\ref{fig:3d_channel_overview}). The setting is detailed in \ref{sec:channel_appendix}. Figure~\ref{fig:channel_flow_u_contour}, \ref{fig:channel_flow_v_contour}, \ref{fig:channel_flow_w_contour} show the evolution of the three-dimensional velocity using the proposed G-LED and LES \cite{nicoud1999subgrid}. G-LED provides an evolution of the system dynamics  with an online speedup of approximately 73X  \footnote{LES is performed in parallel on AMD Ryzen 9 7950x 16-core processor and G-LED on a single NVIDIA A100.}. 

\begin{figure}[htp]
	\centering
	\includegraphics[width=0.32\textwidth]{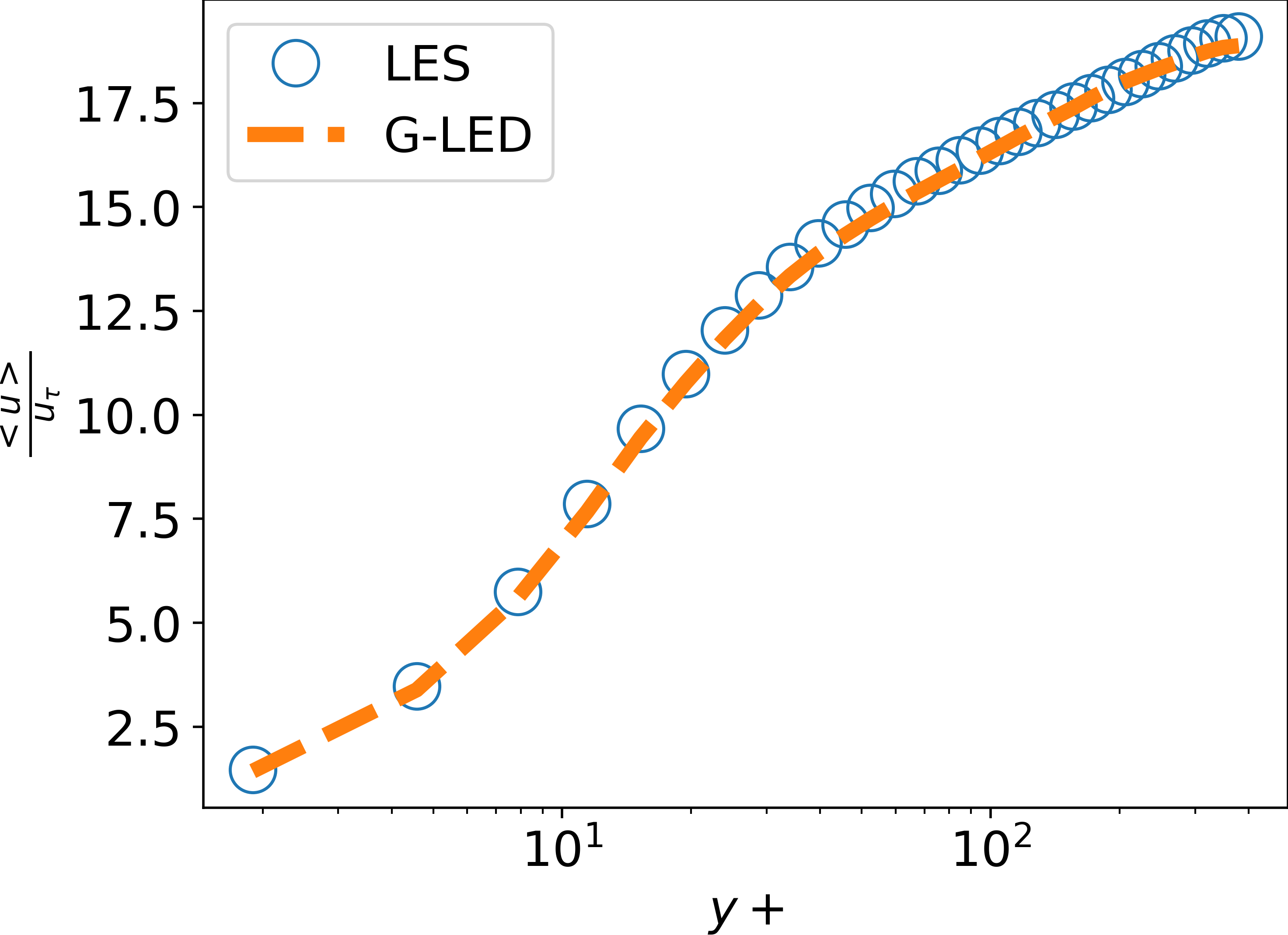}
	\includegraphics[width=0.32\textwidth]{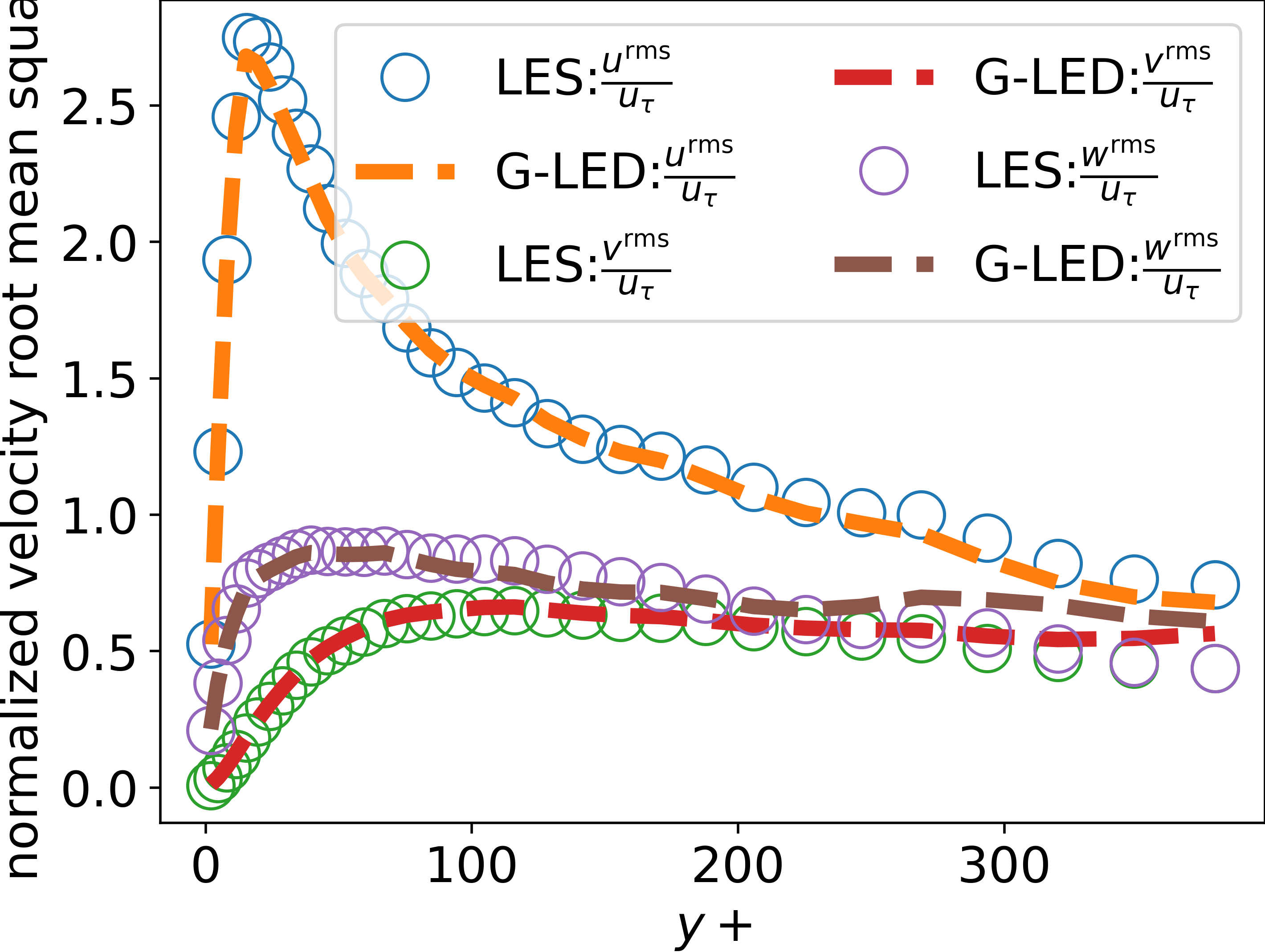}
	\includegraphics[width=0.32\textwidth]{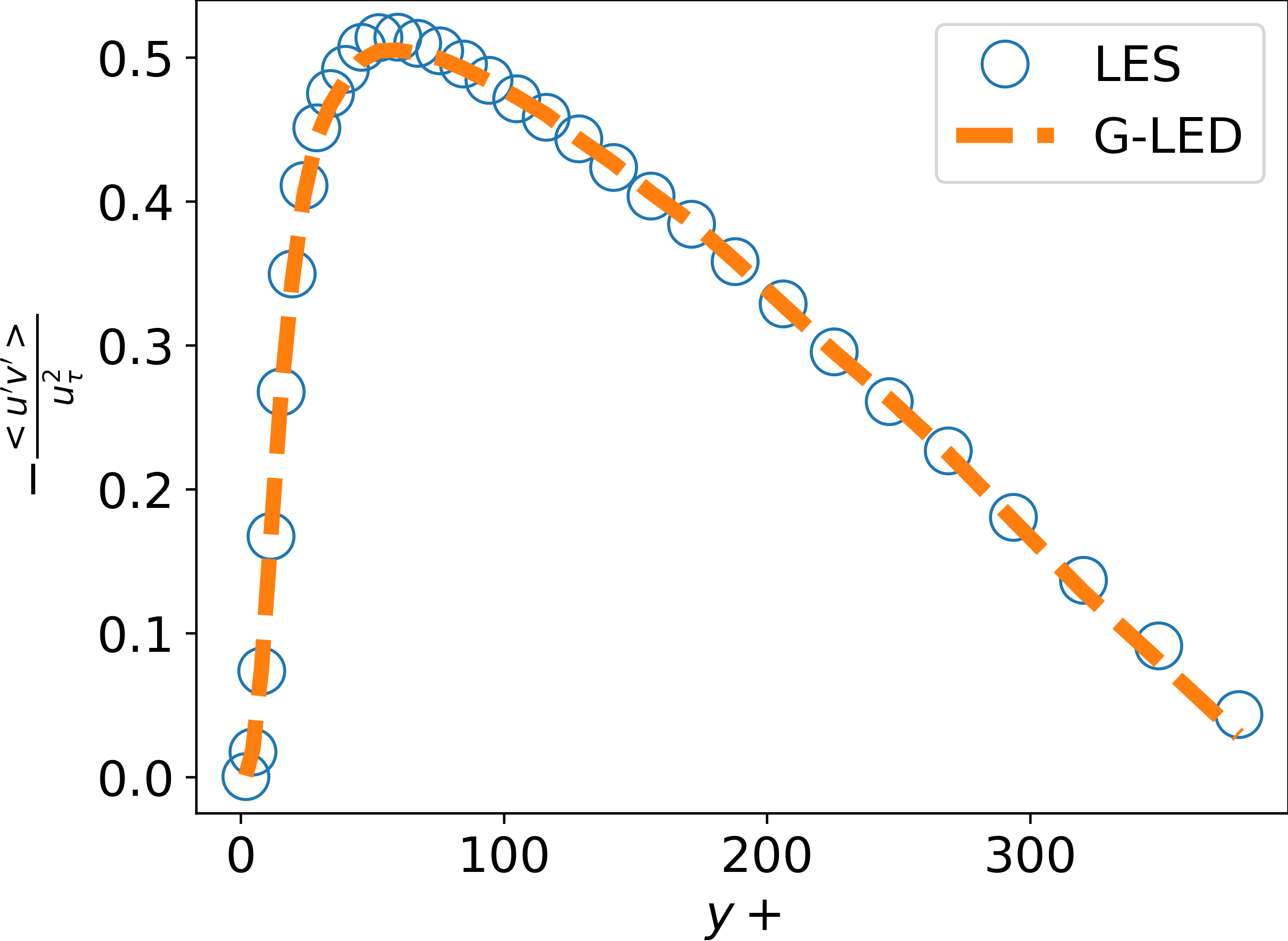}
	\caption{Mean of streamwise velocity (\textit{left}),  root mean square velocity profiles (\textit{middle}) and non-zero shear stress (\textit{right}). } 
 \label{fig:mean_rms_stress}
\end{figure}

We compare the average streamwise velocity profiles from  G-LED and LES in Figure~\ref{fig:mean_rms_stress}. On the \textit{left} side of Figure~\ref{fig:mean_rms_stress}, the profiles are presented in wall coordinates and scaled with the friction velocity \cite{pope2001turbulent}. The results from G-LED and LES are almost indistinguishable. In turbulent flows, the fundamental parameters characterizing the fluctuating motion are encapsulated in the Reynolds stress tensor components. The diagonal components of this tensor are presented in the form of the standard (root-mean-square) deviations (Figure~\ref{fig:mean_rms_stress} \textit{middle}), and the only one non-zero off-diagonal component is usually referred to as the turbulent shear stress (Figure~\ref{fig:mean_rms_stress} \textit{right}). The stress tensor components for G-LED are computed based on the predicted velocities for the unseen test data. Overall, the components of the stress tensor from G-LED are in reasonable agreement with the LES.

The performance of G-LED is further analysed by considering the energy spectrum (Figure~\ref{fig:channel_energy_spectrum}), spatial (Figure~\ref{fig:channel_spatial_cor}) and temporal (Figure~\ref{fig:channel_temporal_cor}) correlations of the velocity components. The good agreement of these quantities demonstrates that G-LED can evolve dynamics with internal length scales of the turbulent structures present in the LES.\\

\begin{figure}[htp]
    \centering
    \includegraphics[width=0.45\textwidth]{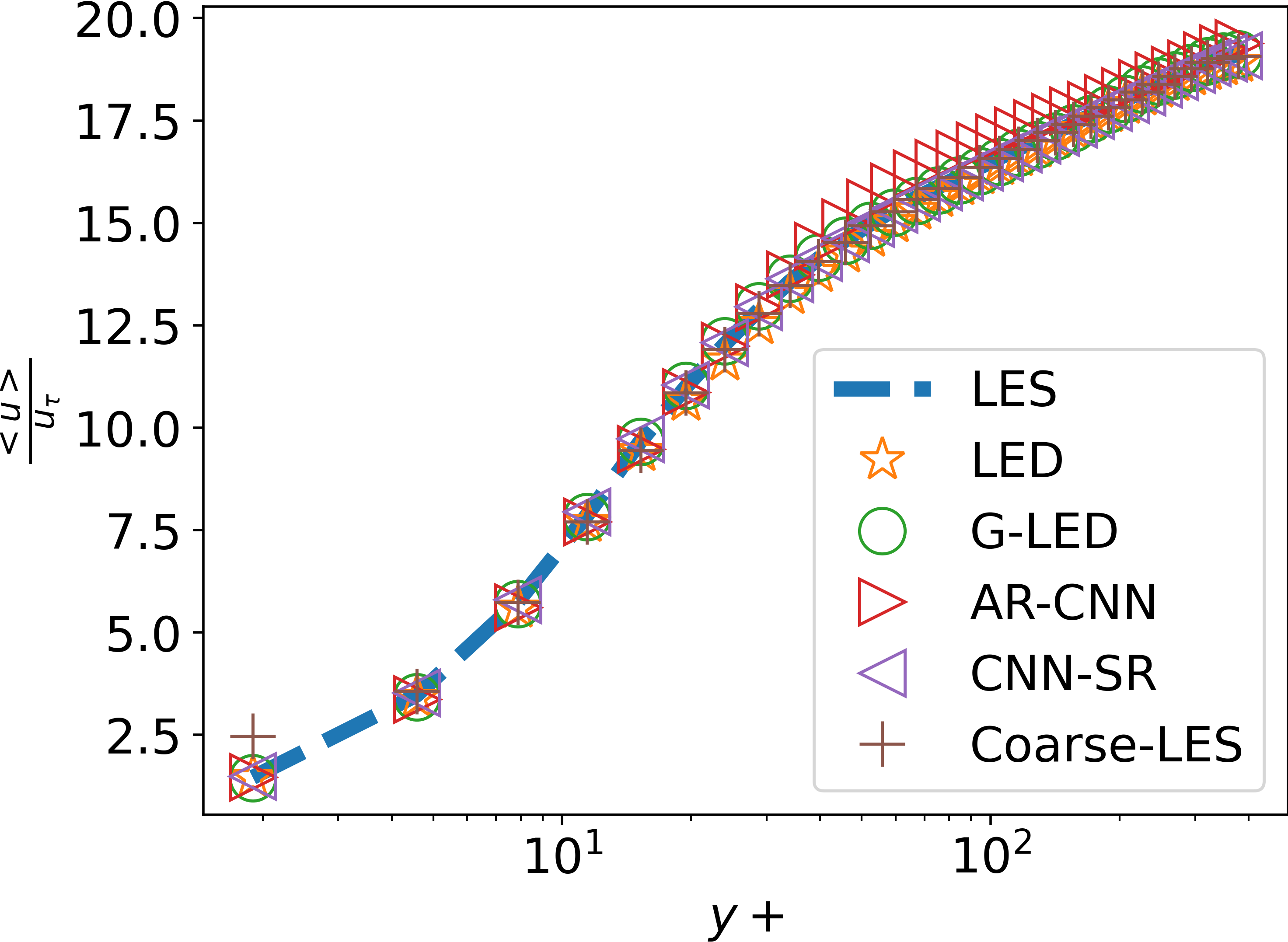}
    \includegraphics[width=0.45\textwidth]{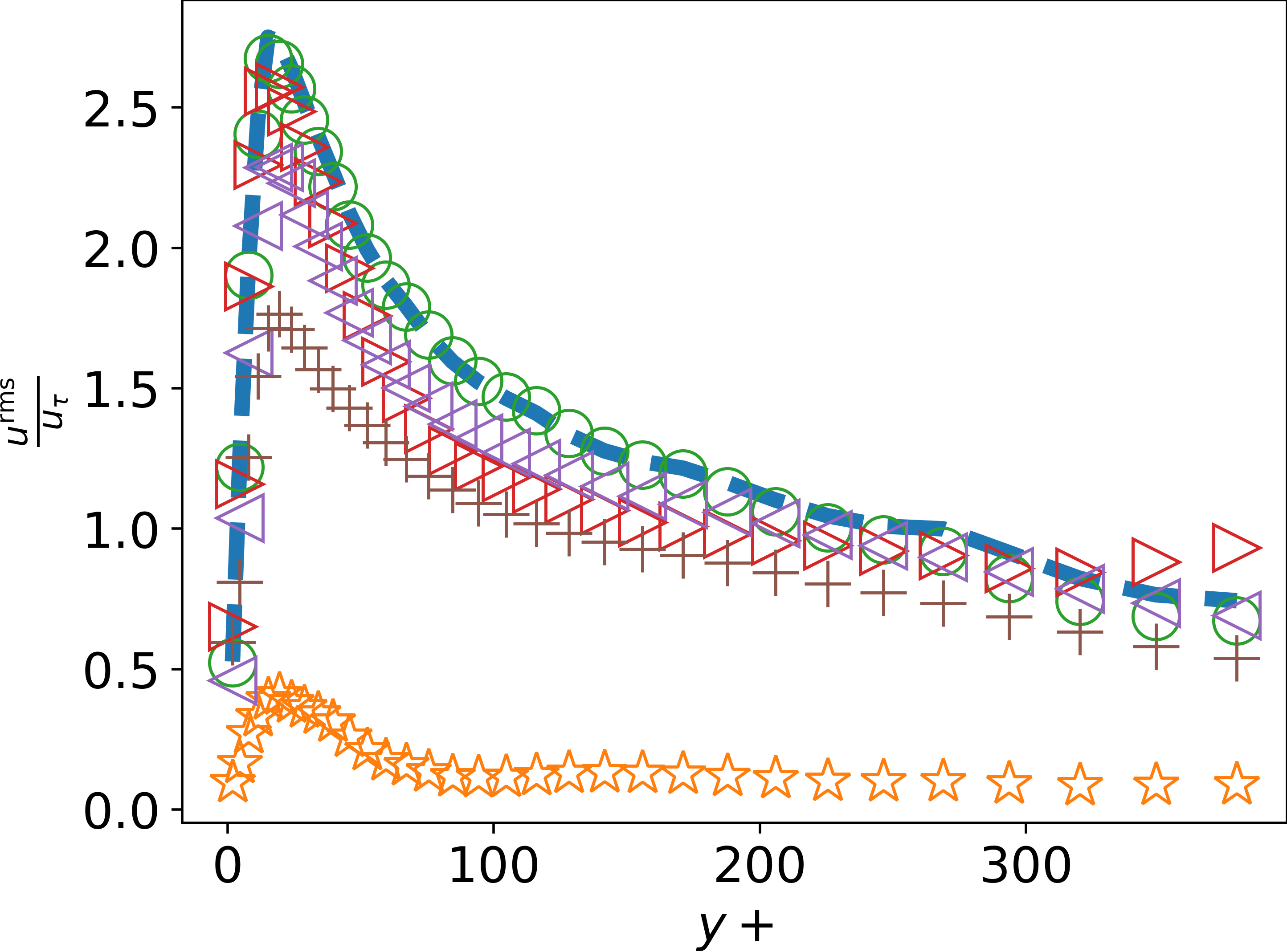}
    \vfil
    \includegraphics[width=0.45\textwidth]{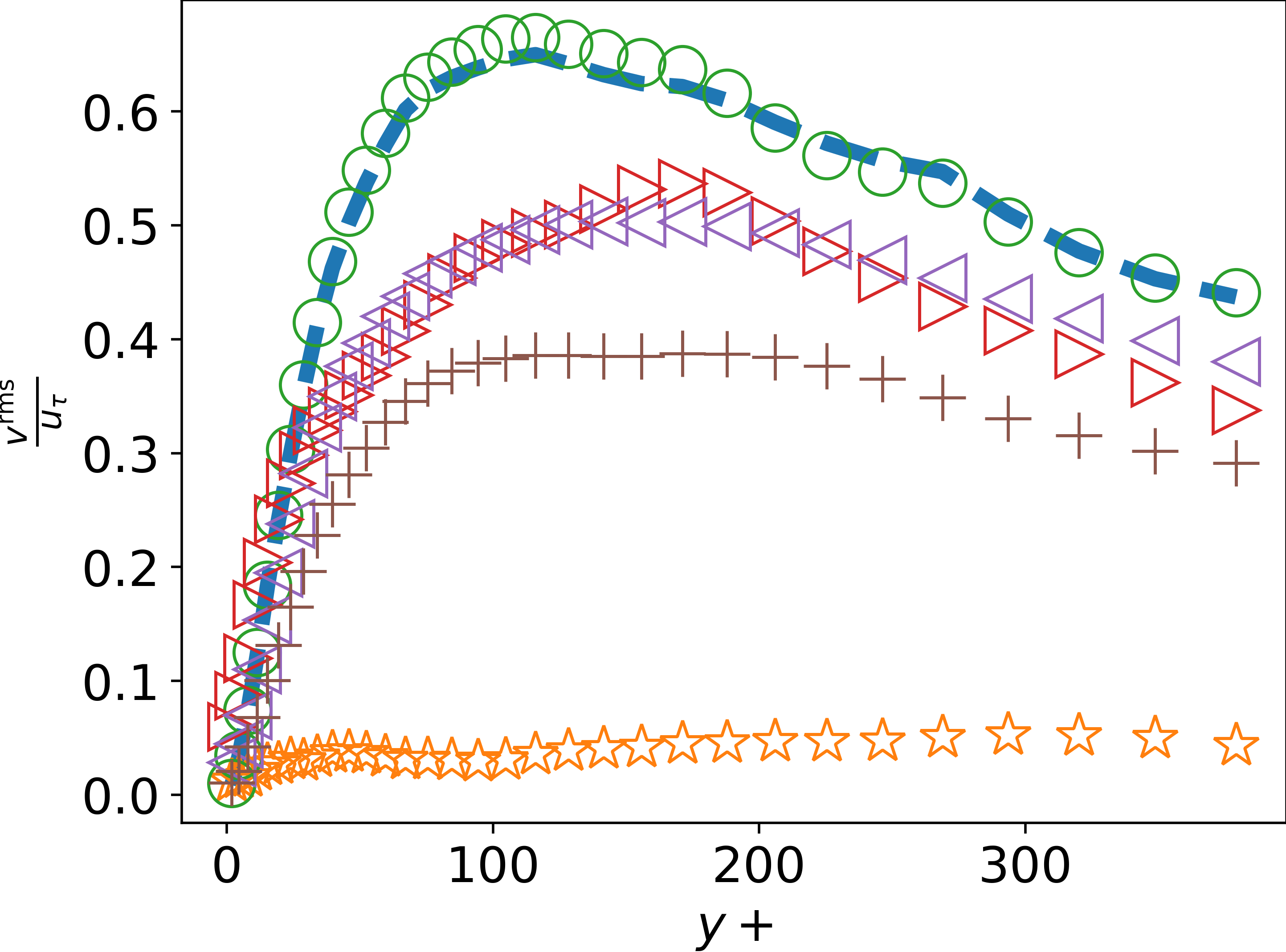}
    \includegraphics[width=0.45\textwidth]{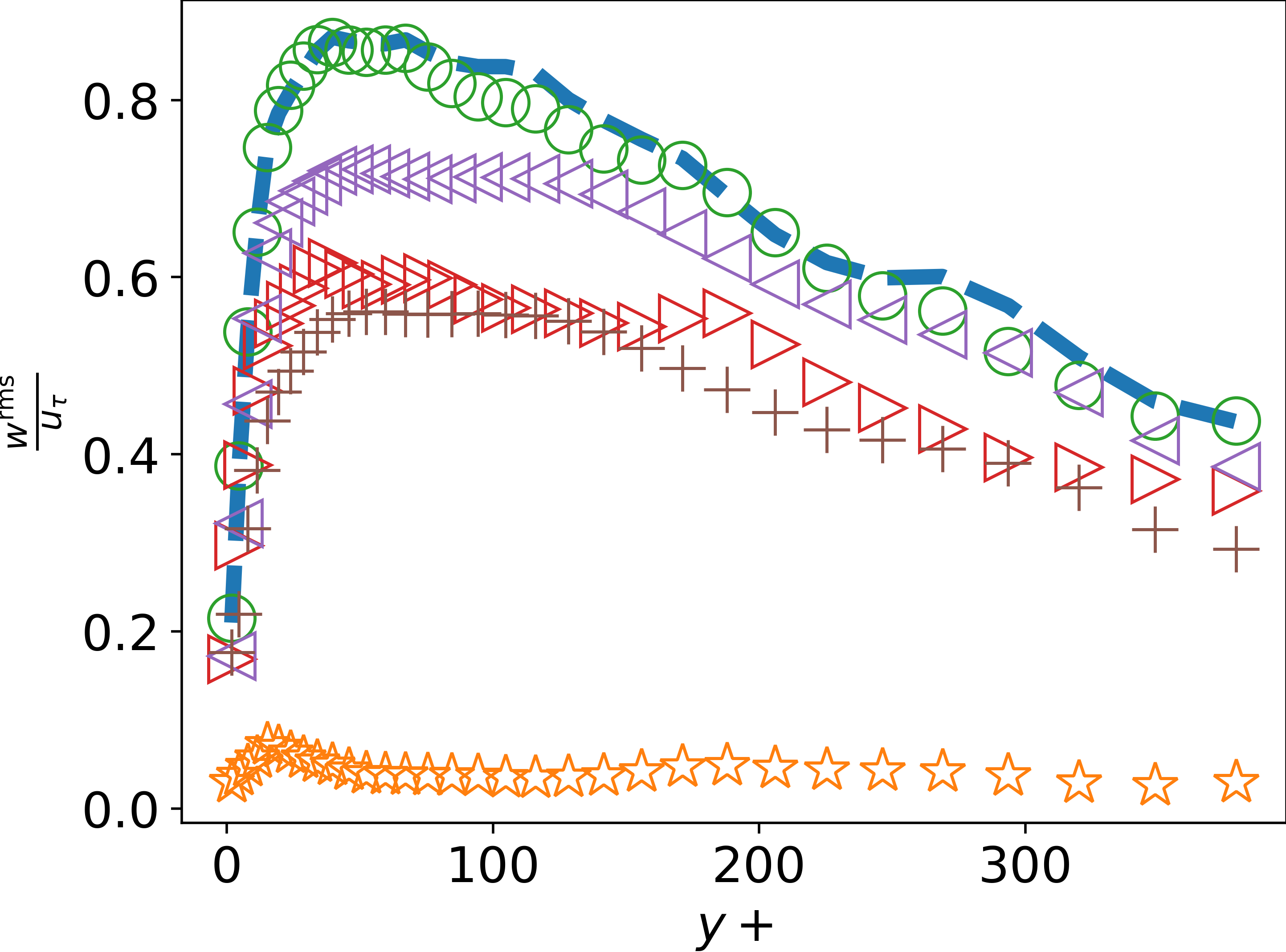}
    \caption{Several baselines are compared for the 3-D channel flow. LES represents the Large Eddy Simulation, which serves as the referred ground truth in this paper. LED represents Convolutional Neural Network and Long Short-Term Memory network. AR-CNN denotes the use of CNN for auto-regressive method, and CNN-SR represents the utilization of CNN to super-resolve the coarse LES.}
    \label{fig:compare}
\end{figure}

Several baselines illustrate the merits of the proposed method: Predictions using  LSTM \cite{vlachas2022multiscale,zhang2019deep, zhang2020physics, ren2022phycrnet} to evolve the dynamics of the macro states fall  short in accurately capturing turbulent fluctuations. A second baseline are Convolutional Neural networks (CNN) that evolve the dynamics in the micro states in an auto-regressive manner, a technique widely employed in \cite{geneva2020modeling, pfaff2020learning, sanchez2020learning}. Subsequently, the ground truth (LES) is downsampled to a coarser grid to represent the optimal performance of coarse LES simulation due to effectively minimizing drift errors. As a final baseline, another CNN is employed to super-resolve the coarse LES, representing the optimal performance using a series of hybrid methods proposed in \cite{ren2023physr, fan2024differentiable,dresdner2022learning,akhare2023physics,geneva2020multi}. Overall, while most methods reasonably capture turbulent fluctuations, none have achieved the level of accuracy exhibited by the proposed framework as shown in Figure \ref{fig:compare}.

\section{Conclusions}
\label{sec:conclusion}
We have presented a generative framework (G-LED) for learning the effective dynamics of complex systems and forecasting their statistics. In G-LED a   Bayesian diffusion model is trained on high dimensional simulations  integrates physical information in its prior knowledge. The diffusion model is  combined a flexible attention model that evolves the latent space dynamics. The generative model projects  the latent space dynamics to high dimensional spaces. The capabilities of the framework are demonstrated on three examples of different dimensionality: the KS equation, 2D flow in a backward facing step and 3D turbulent channel flow. We compare the results of G-LED with the LED framework and with forward numerical simulations. We find that G-LED drastically outperforms several related reduced order methods. Moreover, it accurately predict the statistics of key quantities in the challenging problem of  turbulent channel flows. 

We remark that the present results are used only to demonstrate the methodology and the framework can  be applied to other high-dimensional, non-linear systems of interest \cite{price2023gencast,li2023seeds}.
Future research directions include extending G-LED to parametric dynamical systems. By explicitly accounting for parameters in G-LED, the framework could be transferred to new systems without the need for costly retraining. We note that the introduced mechanism to incorporate physical information into the diffusion model could be used to account for the parametric dependencies. Another unsolved challenge is the optimal size of the latent dimension. Currently, we choose the latent dimension based on an a-priori defined subsampling encoder but approaches based on sparsity-enforcing priors could be applicable here and remove the need for any hyperparameter optimization with regards to the latent dimension. We believe that the present work provides a first important step towards integrating generative modeling ideas in the simulation of complex systems.

\section*{Acknowledgments}
H.G. and P.K. acknowledge support by the U.S. Department of Energy, Office of Basic Energy Sciences, under Award Number DE-SC0022199.
S.K. and P.K. acknowledge support by The European High Performance Computing Joint Undertaking
(EuroHPC) Grant DCoMEX (956201-H2020-JTI-EuroHPC-2019-1).

\appendix
\section{Hyperparameters for G-LED}
\begin{table}[htp]
\centering
\begin{tabular}{||c | c | c  |c||} 
	\hline
	 Hyperparameters for       & 1-D KS & 2-D NS & 3-D NS \\ 
	       \hline
	macro dimension ($d_z$) & 16 & $32\times32$& $8\times32\times8$\\
	\hline
		attention cache length ($N_t$) & 512 & 40 & 20\\
	\hline
		number of attention layers & 8 & 8 & 2\\
	\hline	
		number of attention heads  & 4 & 4 & 1\\
	\hline	
	activation function in attention layers & ReLU & ReLU & ReLU\\ 
\hline
layer normalization constant in transformer & $10^{-5}$ & $10^{-5}$ & $10^{-5}$\\ 
\hline
number of convolutional layers in diffusion model & 4 & 4 & 4\\ 
\hline

number of channel per convolution layer & 32 & 32 & 32\\ 
\hline
number of noise step ($N_\epsilon$) & 20 & 20 & 20\\ 
\hline
noise range in diffusion model  & (0.002,80) & (0.002,80) & (0.002,80)\\ 
\hline
Incorporated prior physical information  & None & None & Reynolds stress tensor\\ 
\hline
\end{tabular}
\caption{Important hyperparameters of G-LED for all examples.}
\label{tab:hyper-parameters}
\end{table}

\section{Simulation Details 1D KS equations}
\label{sec:ks_appendix}
The spatial domain is discretized with a grid of $64$ points, and the temporal dynamics are solved using the fourth-order method for stiff PDEs detailed in \cite{kassam2005fourth} with time step $\delta t = 0.025$ at the micro level \cite{menier2023interpretable}. The model is configured to predict for a duration of $T_{\mathrm{prediction}} = 80$, using an initial trajectory of length $T_{\mathrm{warm-up}} = 16$ for the warm-up phase. The data are subsampled with $\Delta t = 0.25$ at the macro level. We used $2.1\times 10^4$ trajectories of length $T=96$  for training and another $1.9\times 10^4$ for validation.

\section{Simulation details and additional results for turbulent flow over backward-facing step}
\label{sec:bfs_appnedix}
The spatial domain is discretized with a $512\times512$ uniform grid
and the temporal domain is discretized with $\delta t = 2\times10^{-4}$. The snapshots are subsampled with a larger time step $\Delta t = 0.05$. $8000$ snapshots are used for training, $500$ snapshots are for validation and $1500$ snapshots are for the testing.

\begin{figure}[htp]
	\centering
	\includegraphics[width=1\textwidth]{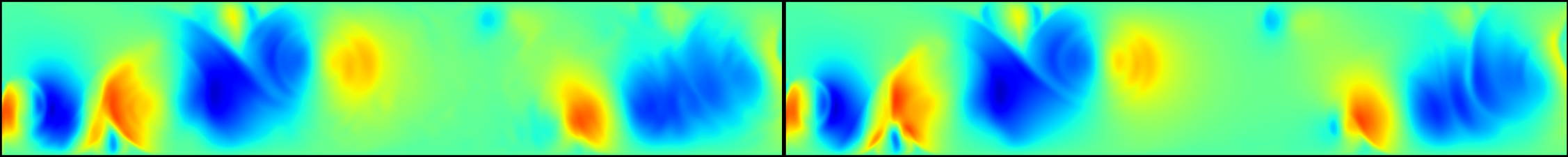}
	\vfil
	\includegraphics[width=1\textwidth]{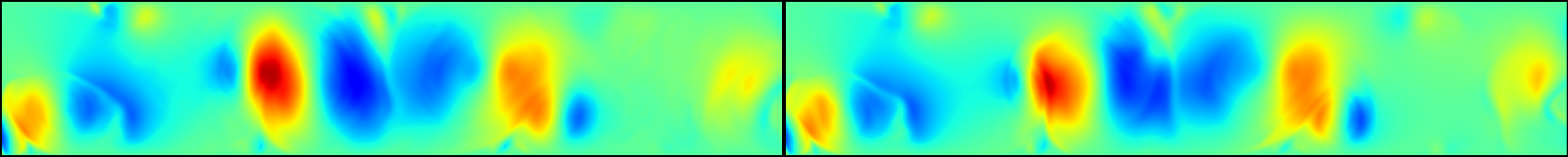}
	\vfil
	\includegraphics[width=1\textwidth]{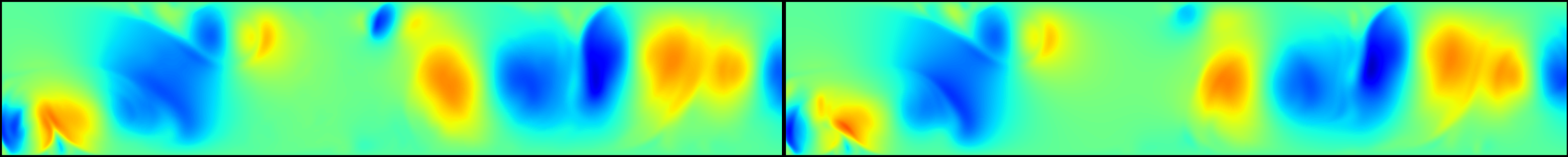}
	\vfil
	\includegraphics[width=1\textwidth]{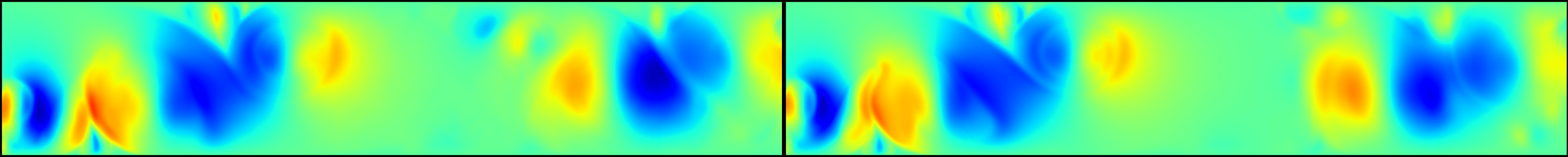}
	\vfil
	\colorbarMatlabJet{-12}{-5.25}{1.5}{8.25}{15}
	\caption{The wallnormal velocity rollout of LED (\textit{left}) and LES (\textit{right}) given the same initial state at $t=0.05, 0.50, 0.95, 1.25$ (\textit{from top to bottom})}
\label{fig:bfs_v_contour}
\end{figure}
\begin{figure}[htp]
	\centering
	\includegraphics[width=0.32\textwidth]{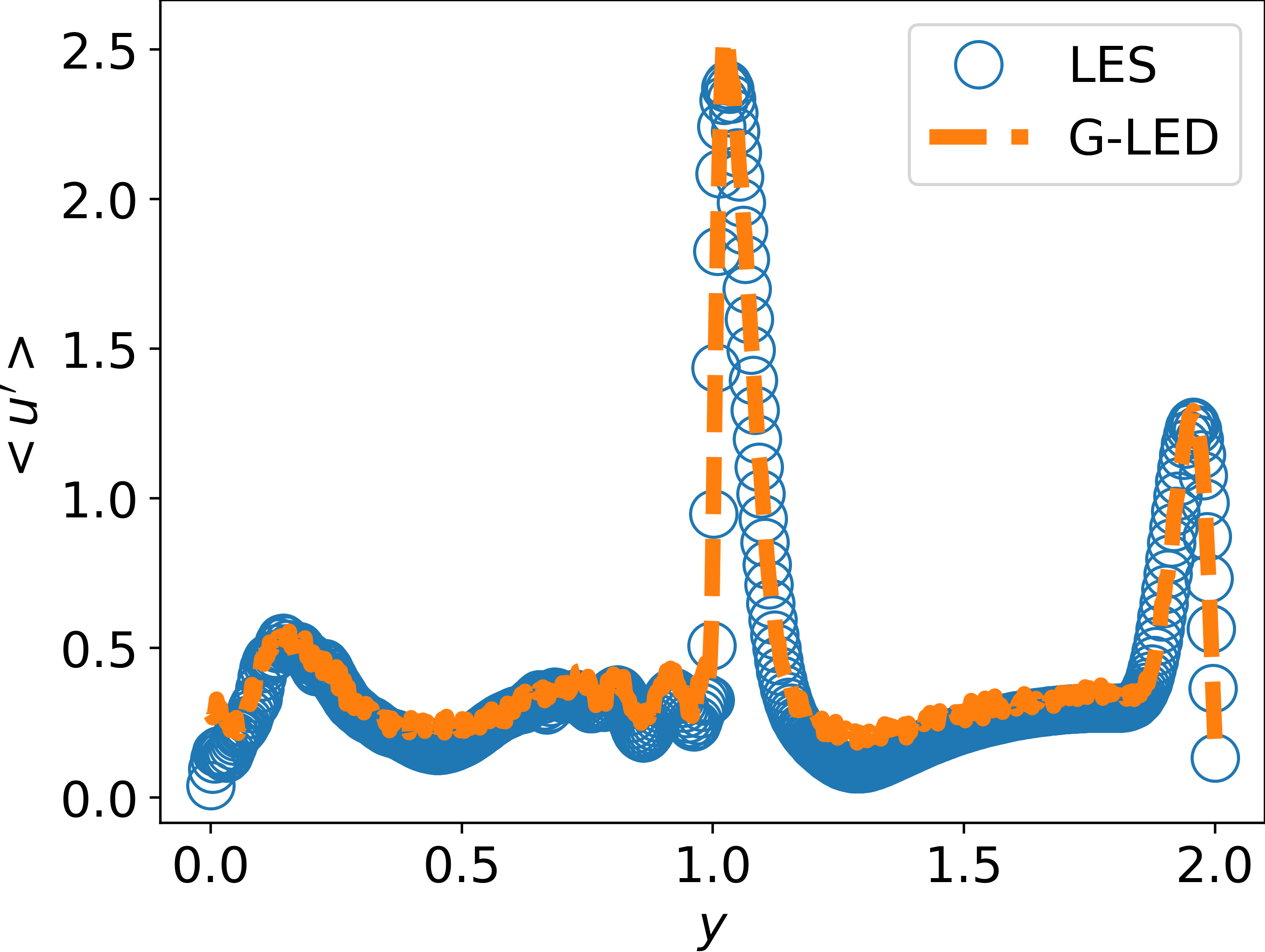}
	\includegraphics[width=0.32\textwidth]{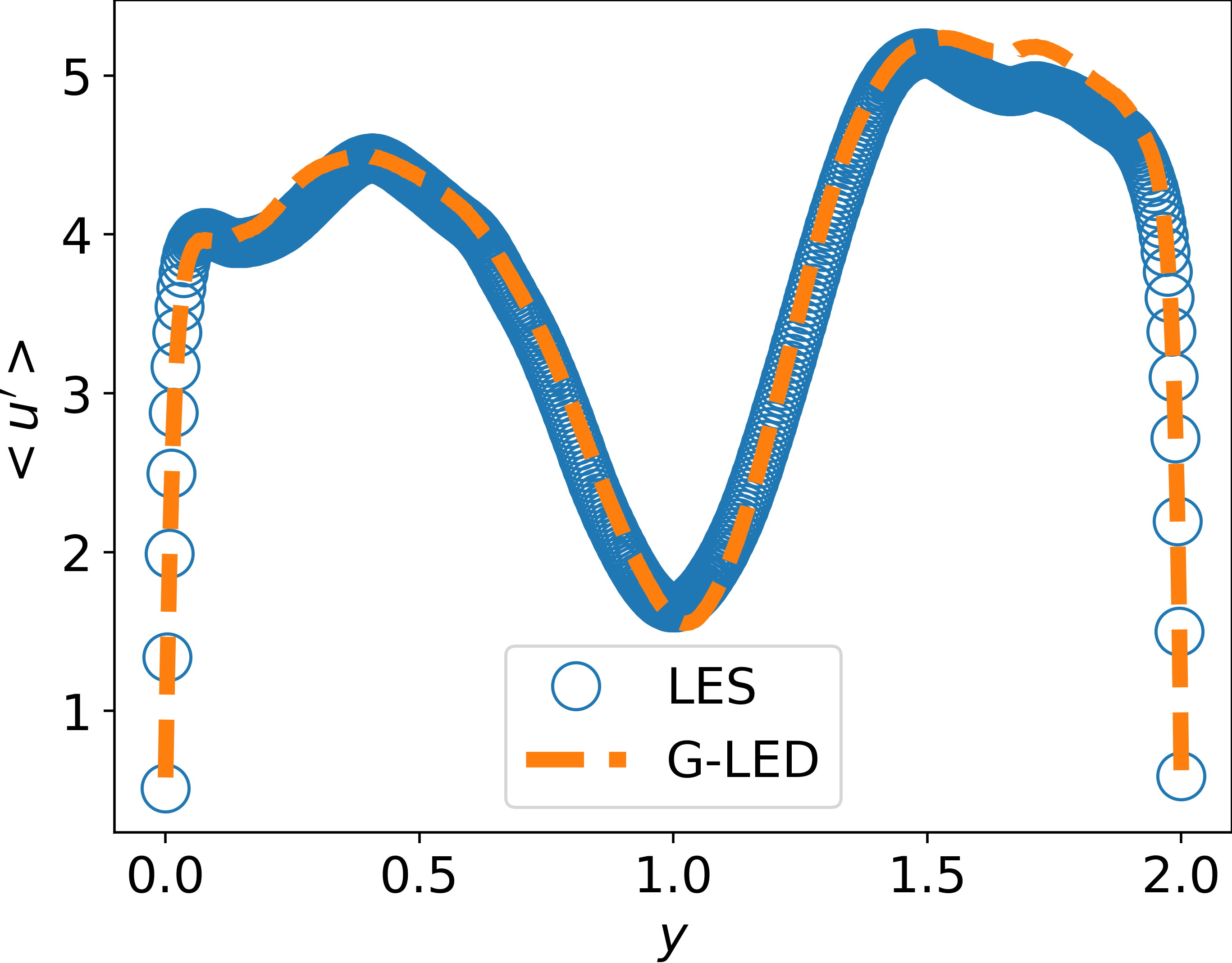}
	\includegraphics[width=0.32\textwidth]{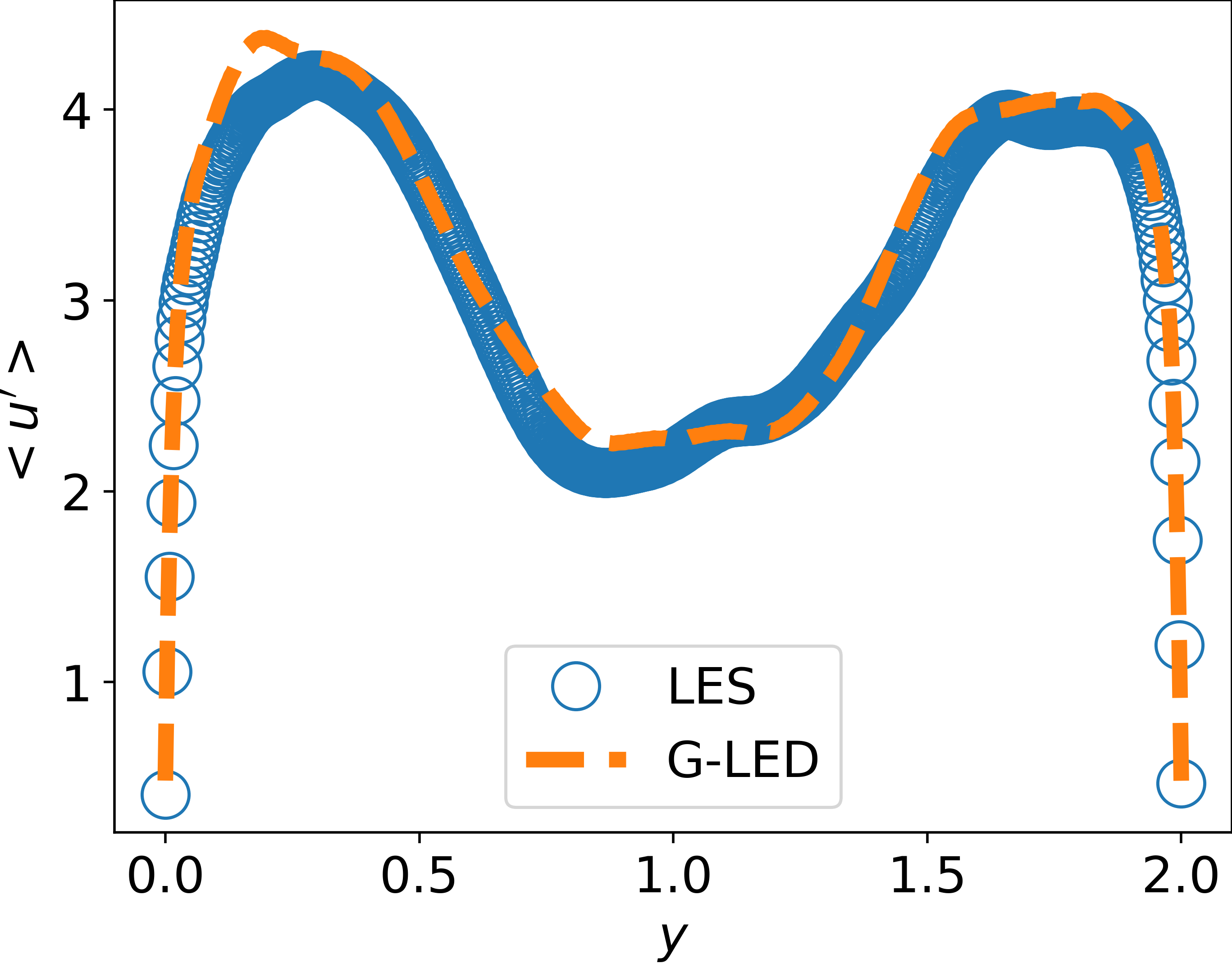}
	\caption{Root mean square of streamwise velcotiy fluctuation at $x=0, 4, 8$ (\textit{from left to right}). } 
\label{fig:u_rms_bfs}
\end{figure}

\begin{figure}[htp]
	\centering
	\includegraphics[width=0.32\textwidth]{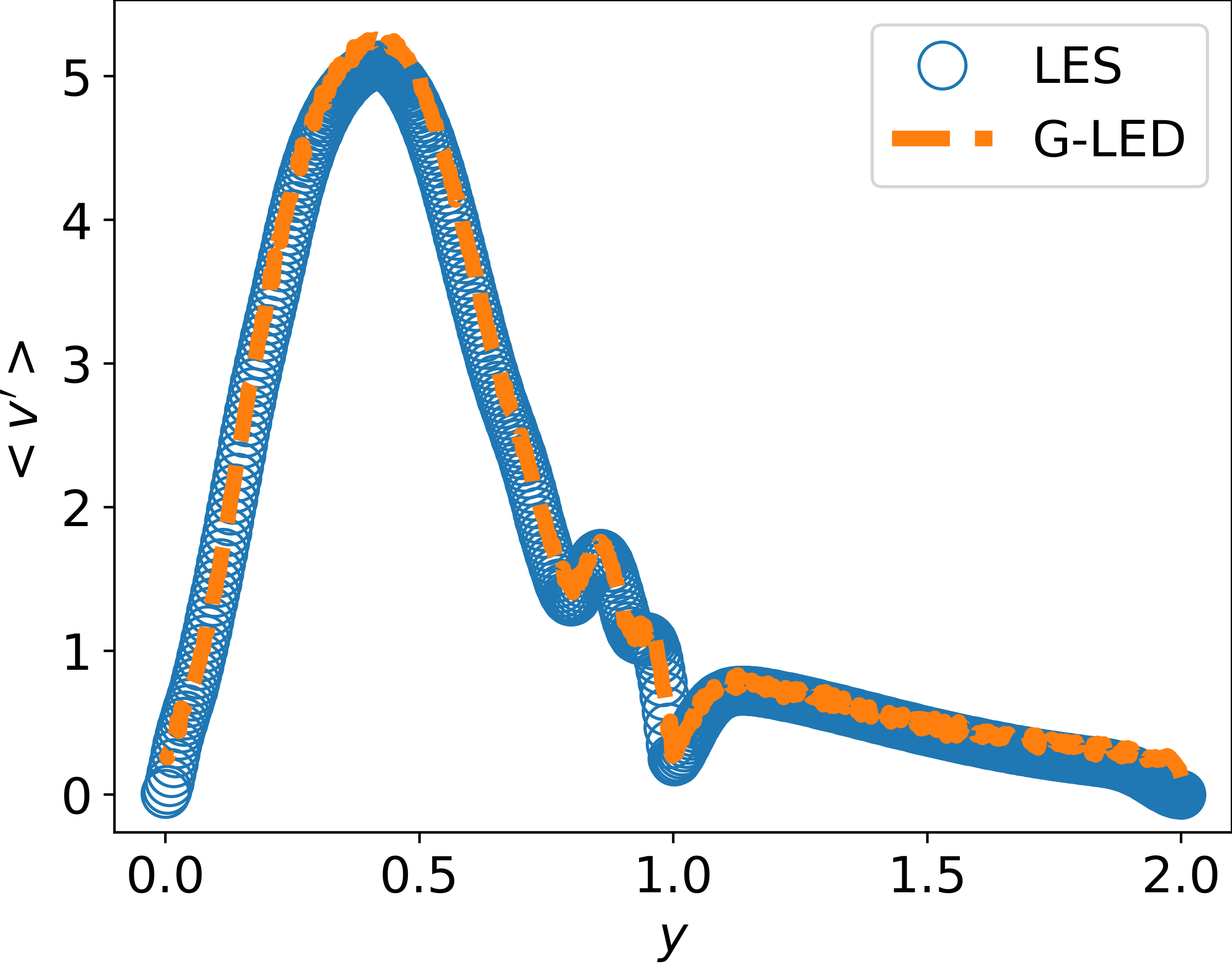}
	\includegraphics[width=0.32\textwidth]{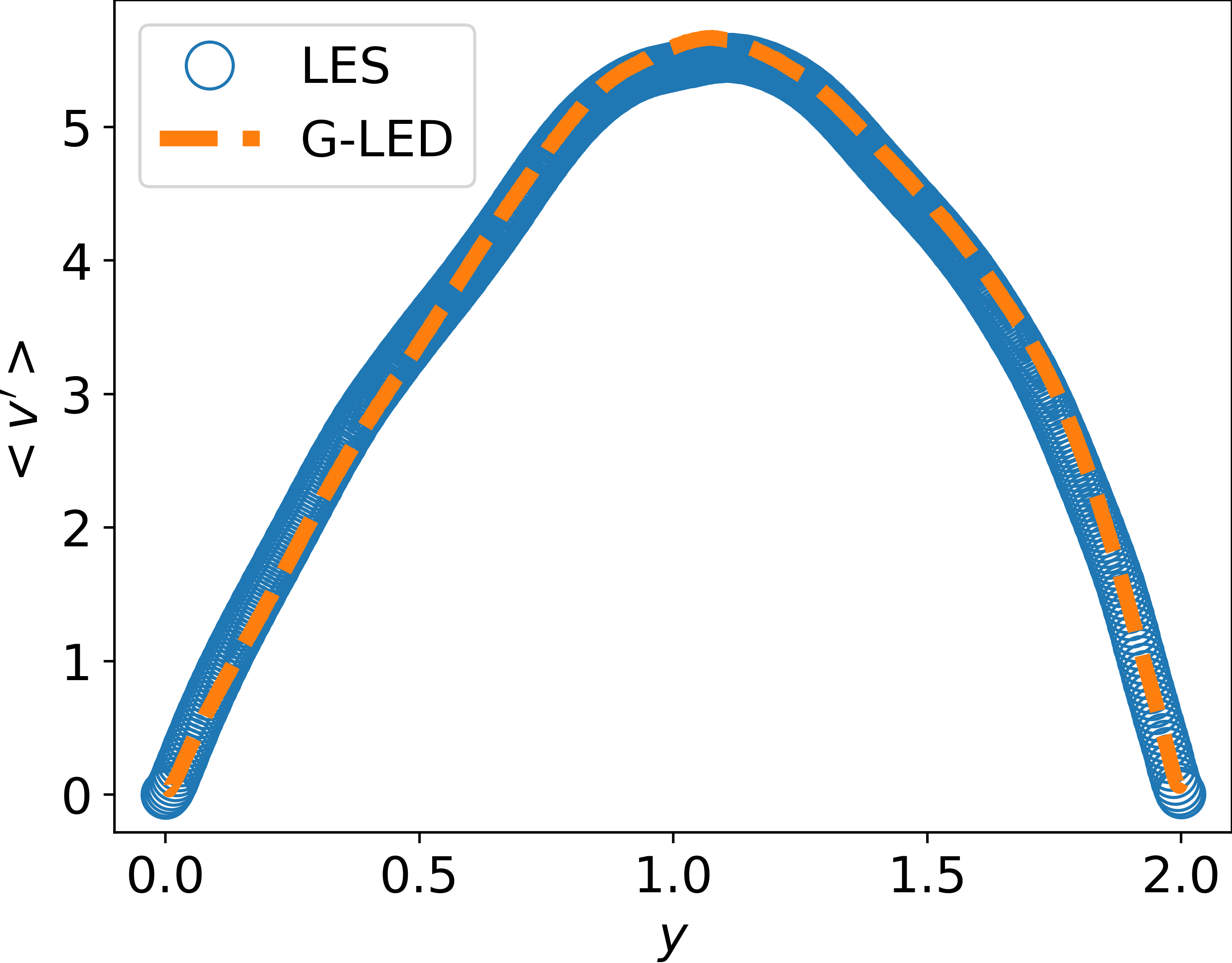}
	\includegraphics[width=0.32\textwidth]{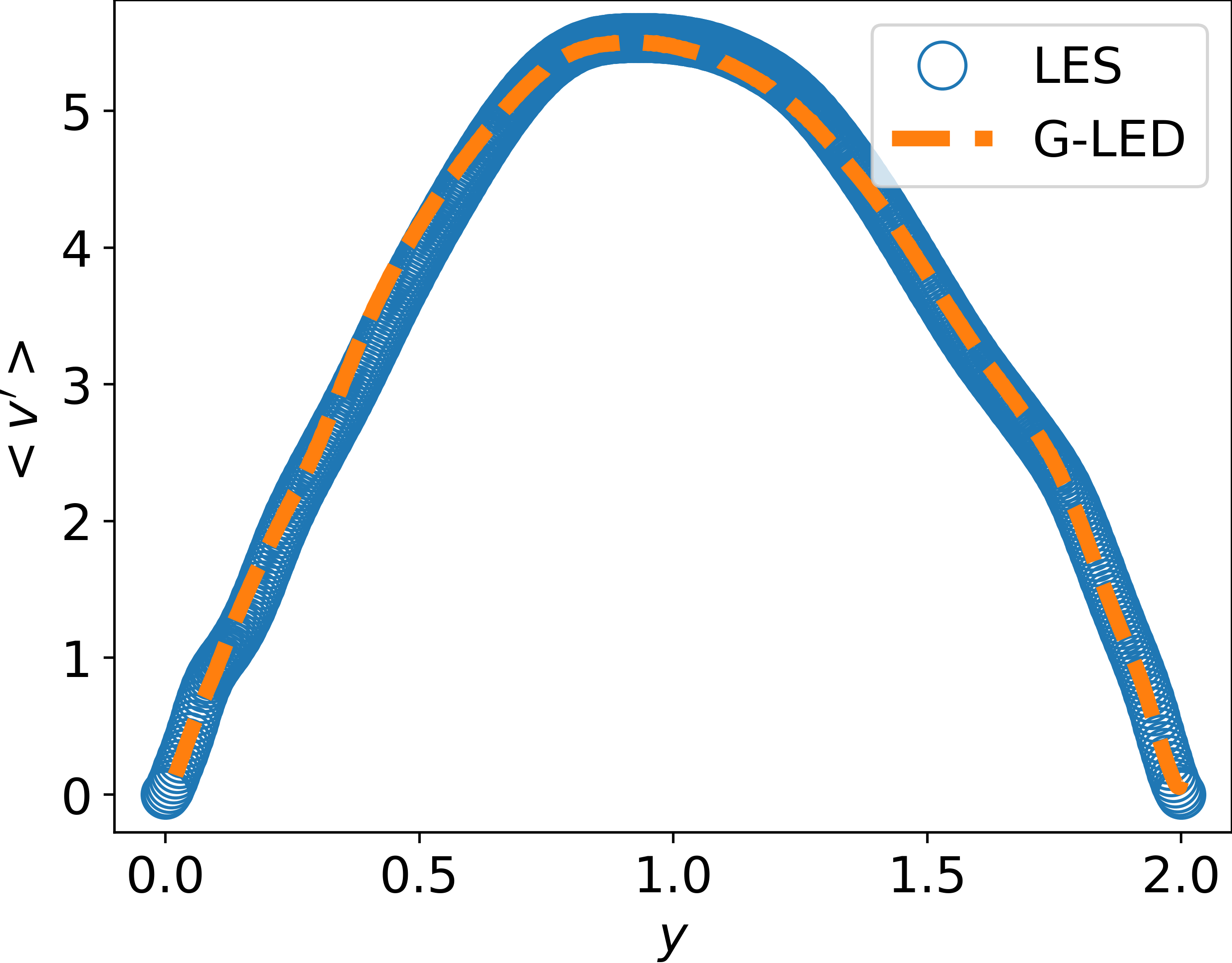}
	\caption{Root mean square of wallnormal velocity fluctuation at $x=0, 4, 8$ (\textit{from left to right}). } 
\label{fig:v_rms_bfs}
\end{figure}

\begin{figure}[htp]
	\centering
	\includegraphics[width=0.32\textwidth]{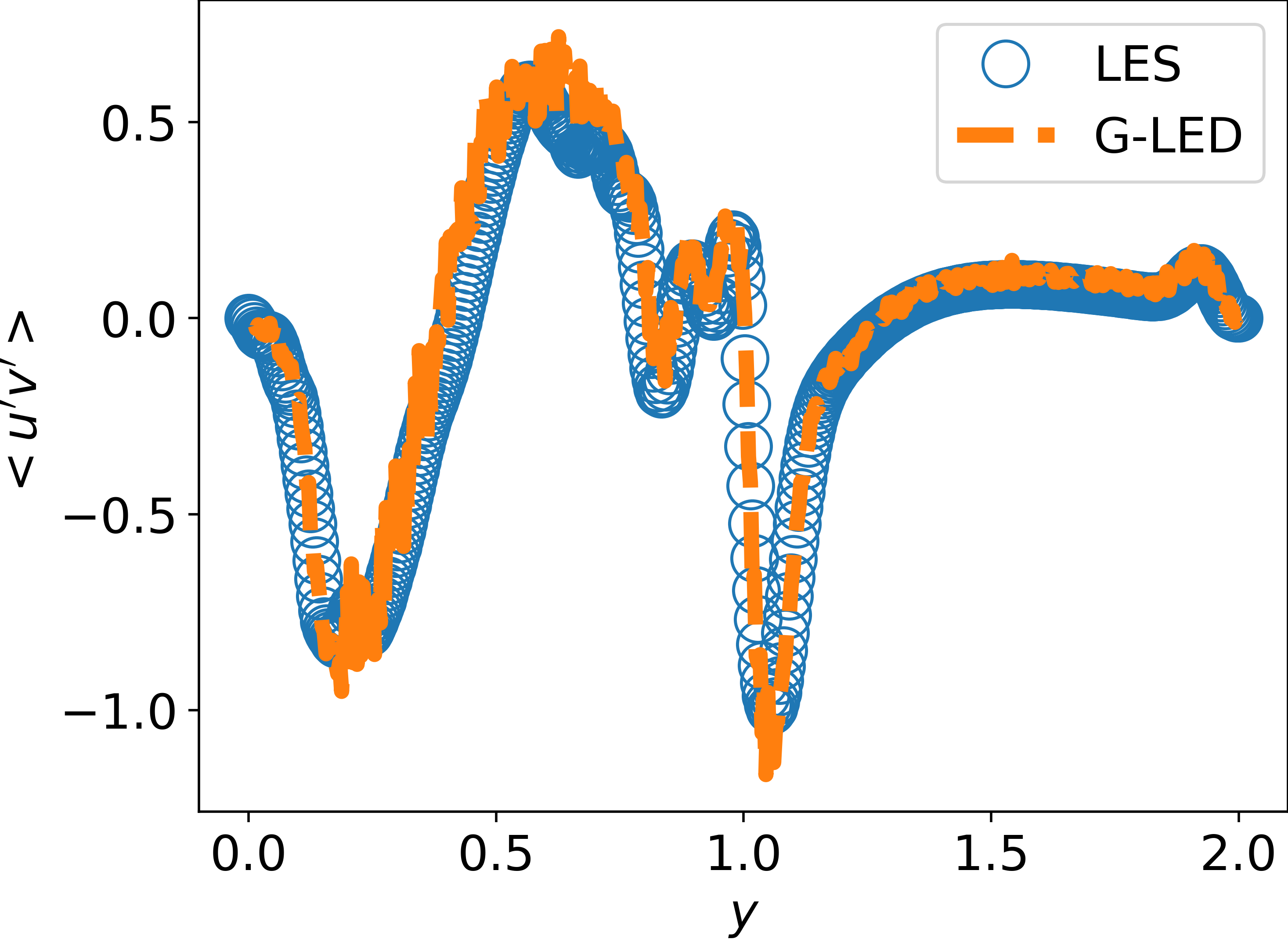}
	\includegraphics[width=0.32\textwidth]{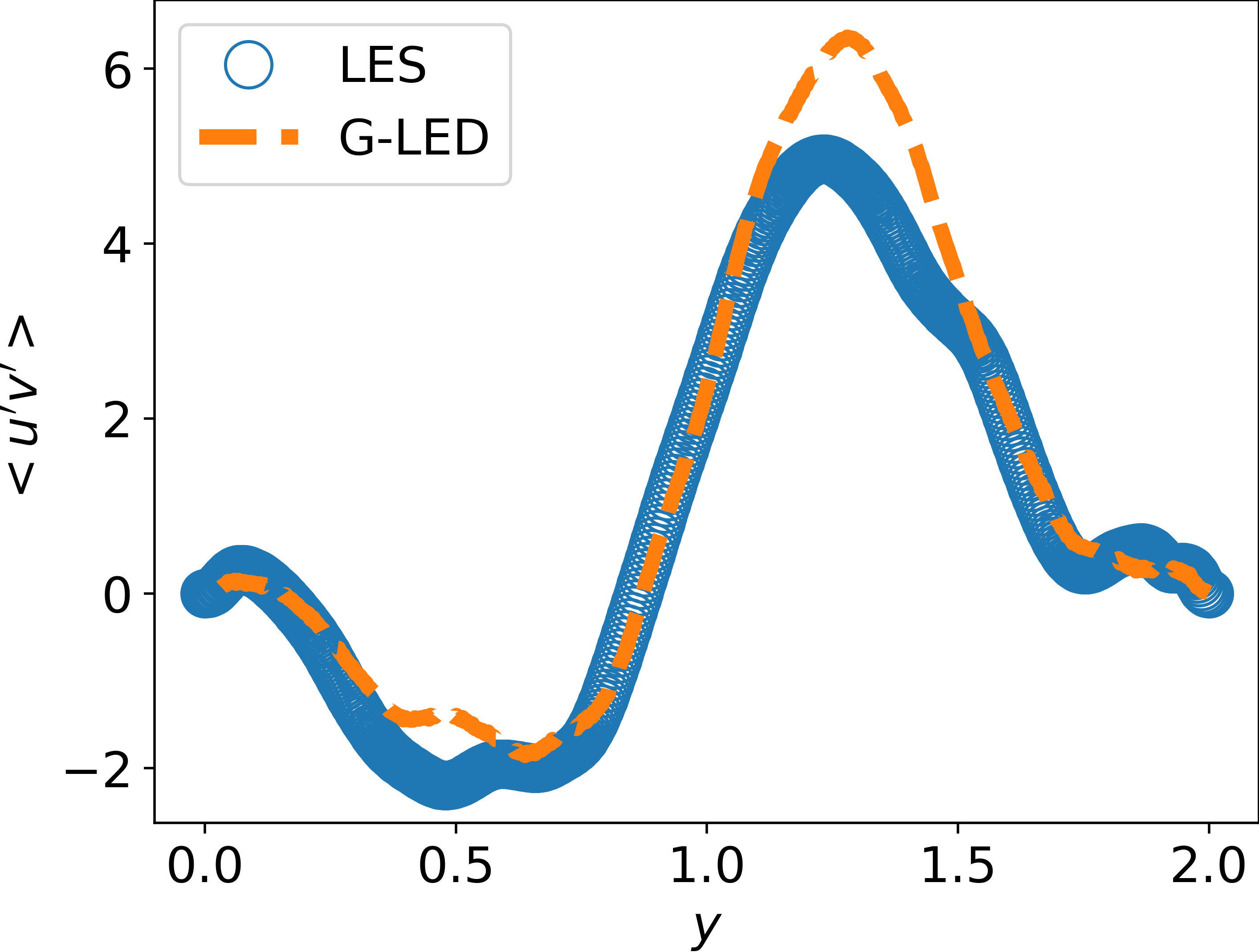}
	\includegraphics[width=0.32\textwidth]{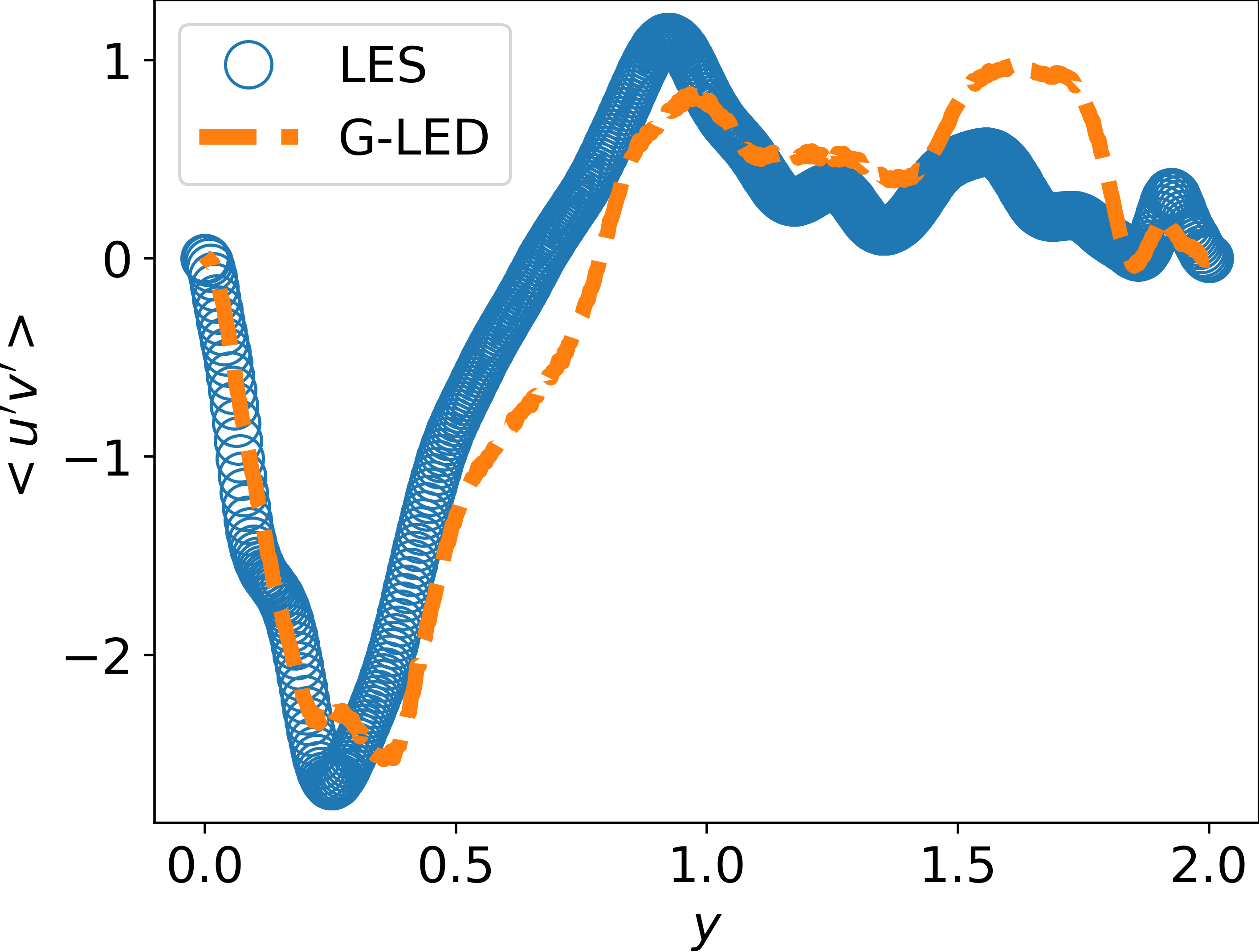}
	\caption{Mean stress of streamwise-wallnormal velocity at $x=0, 4, 8$ (\textit{from left to right}). }
  \label{fig:uv_rms_bfs}
\end{figure}

\begin{figure}[htp]
	\centering
	\includegraphics[width=0.32\textwidth]{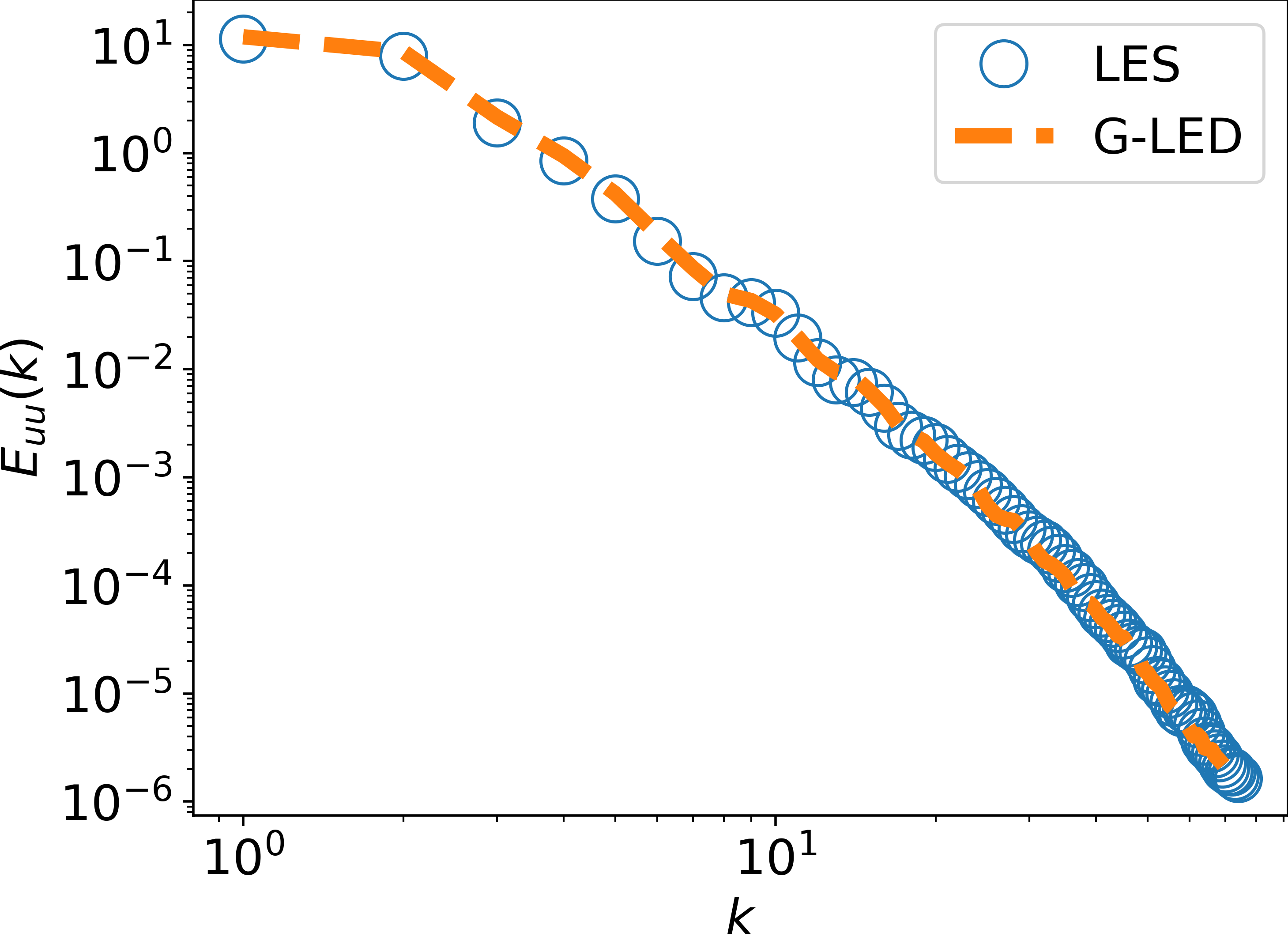}
	\includegraphics[width=0.32\textwidth]{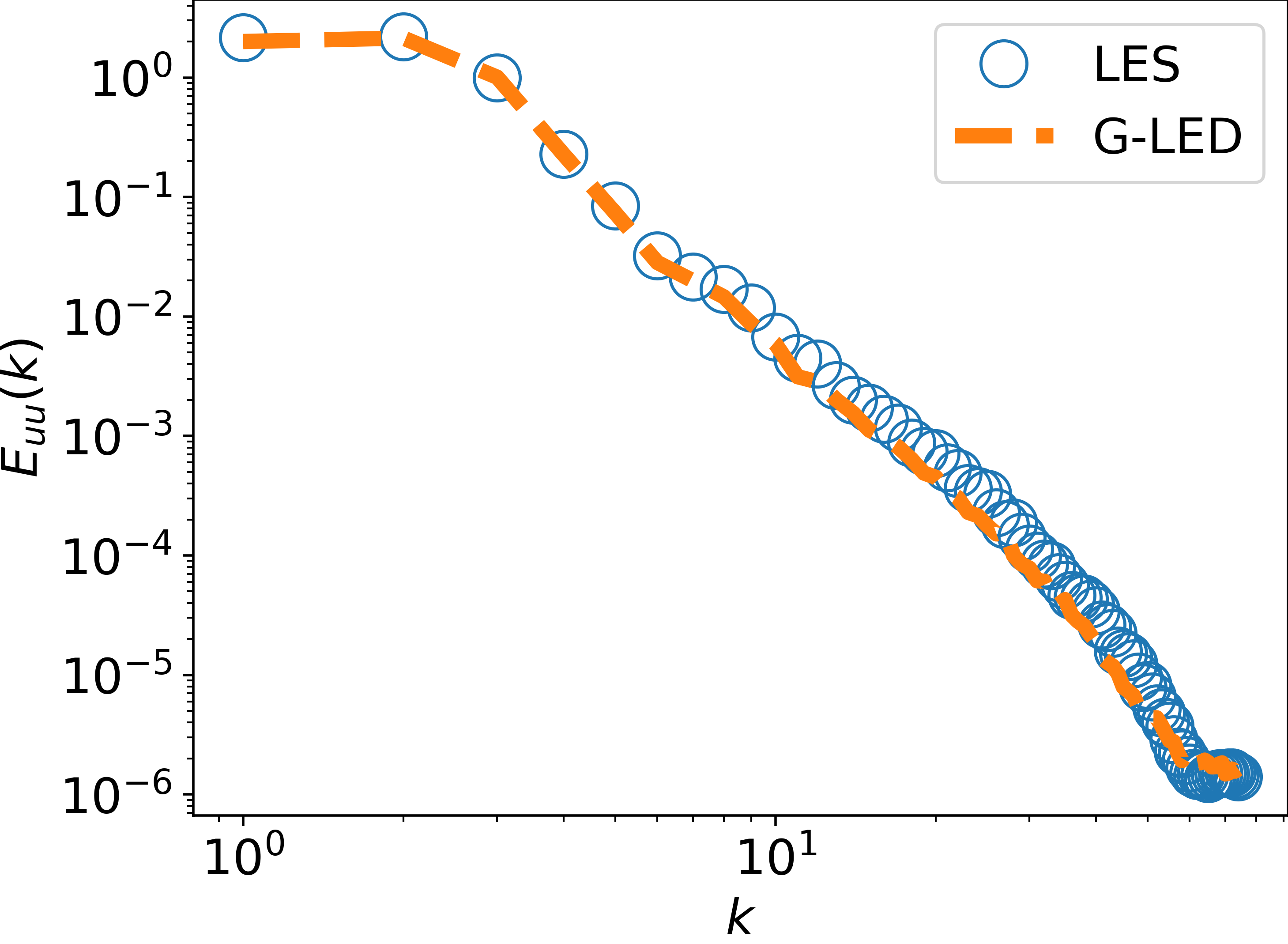}
	\includegraphics[width=0.32\textwidth]{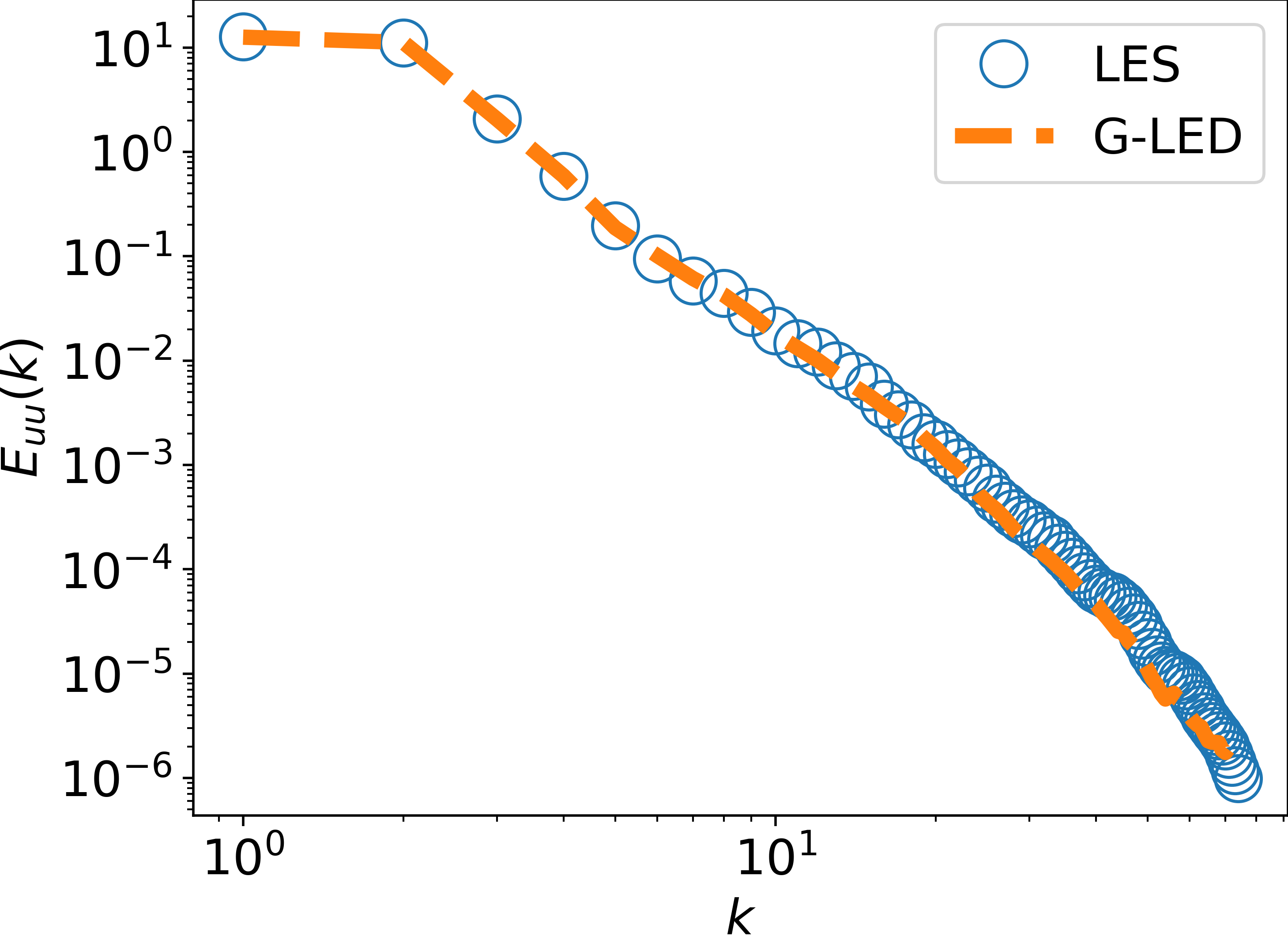}
	\caption{Energy spectrum of streamwise velocity at $y=0.2, 1, 1.8$ (\textit{from left to right}). } 
 \label{fig:u_energy_bfs}
\end{figure}

\begin{figure}[htp]
	\centering
	\includegraphics[width=0.32\textwidth]{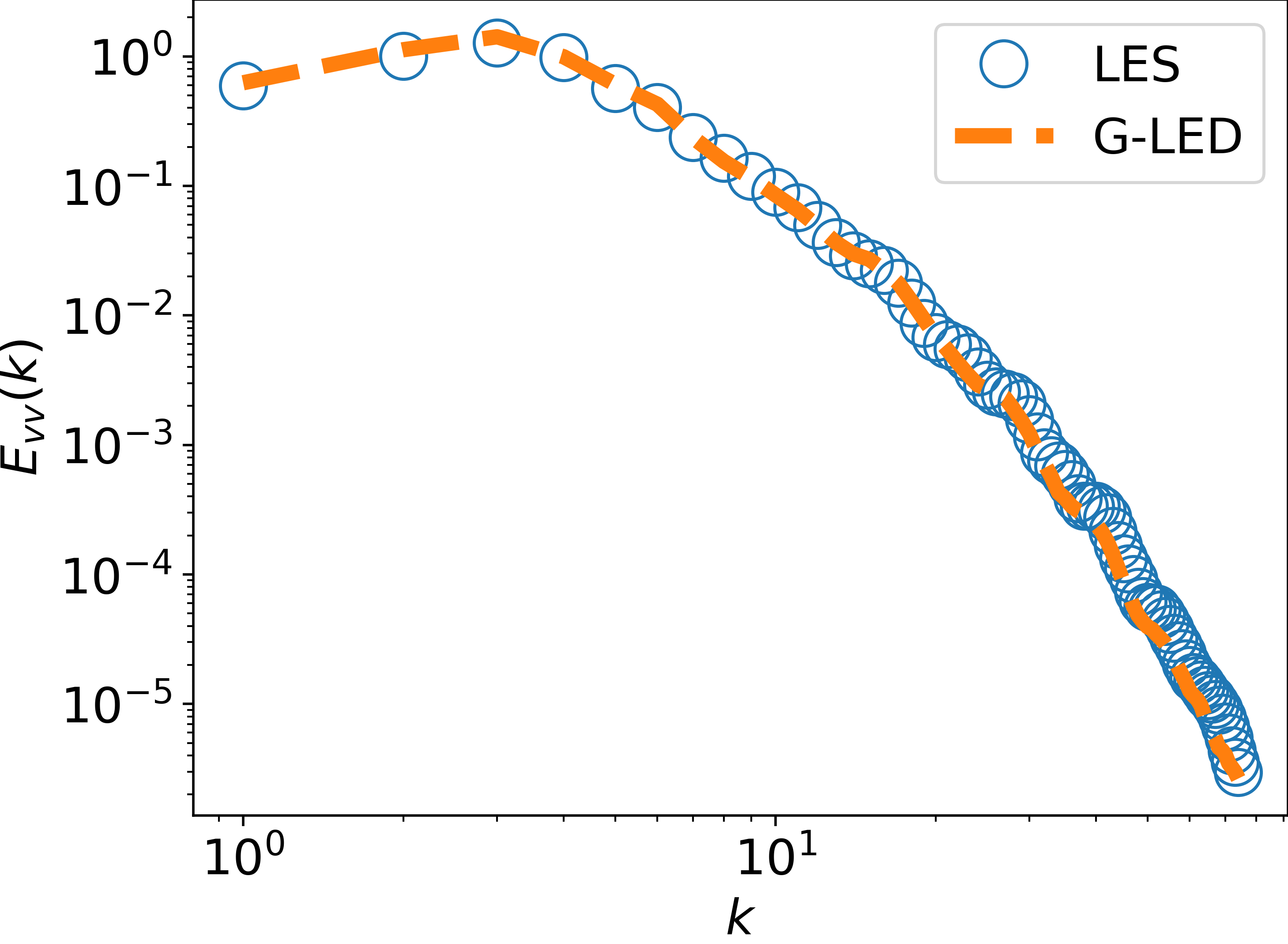}
	\includegraphics[width=0.32\textwidth]{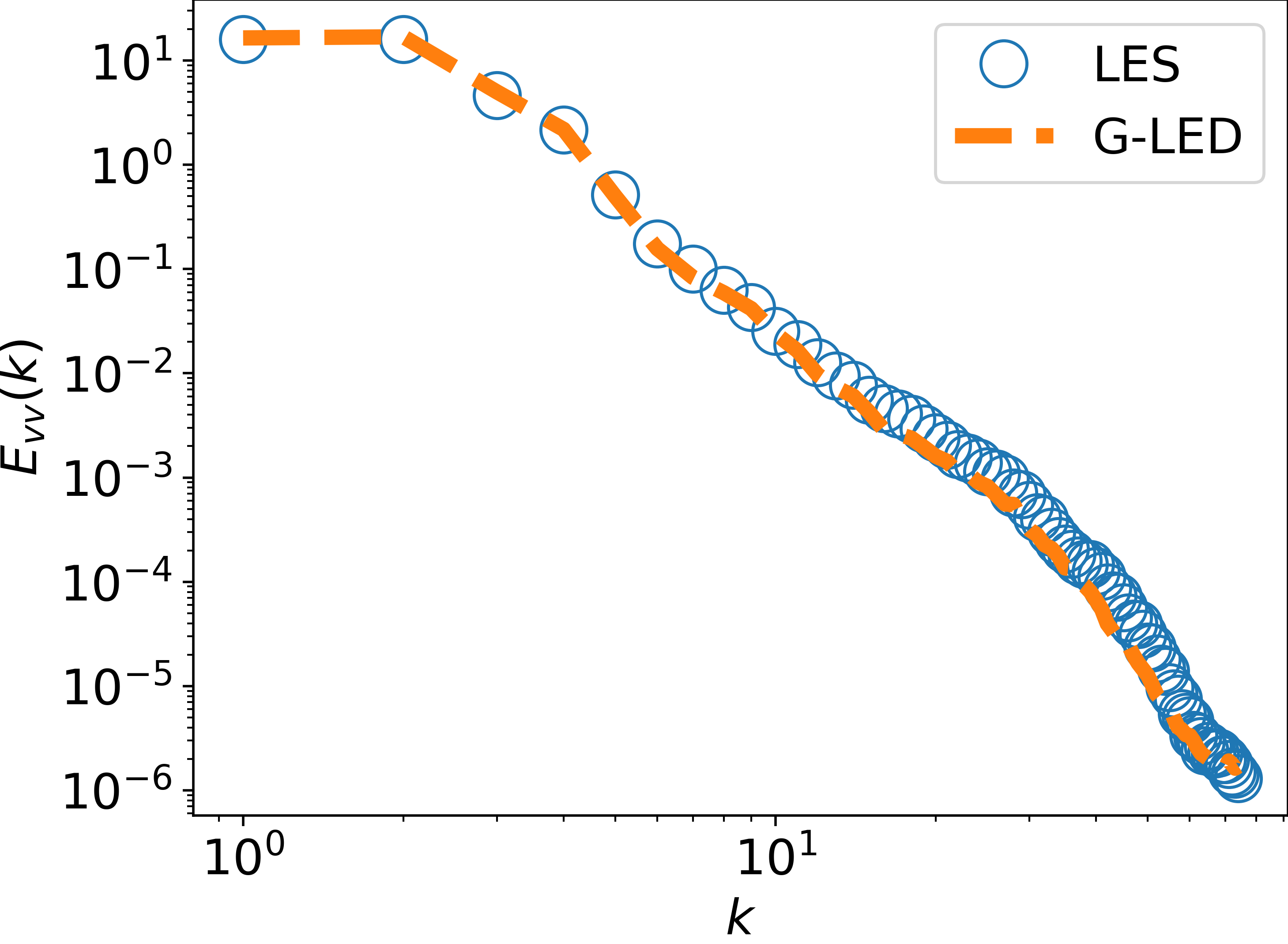}
	\includegraphics[width=0.32\textwidth]{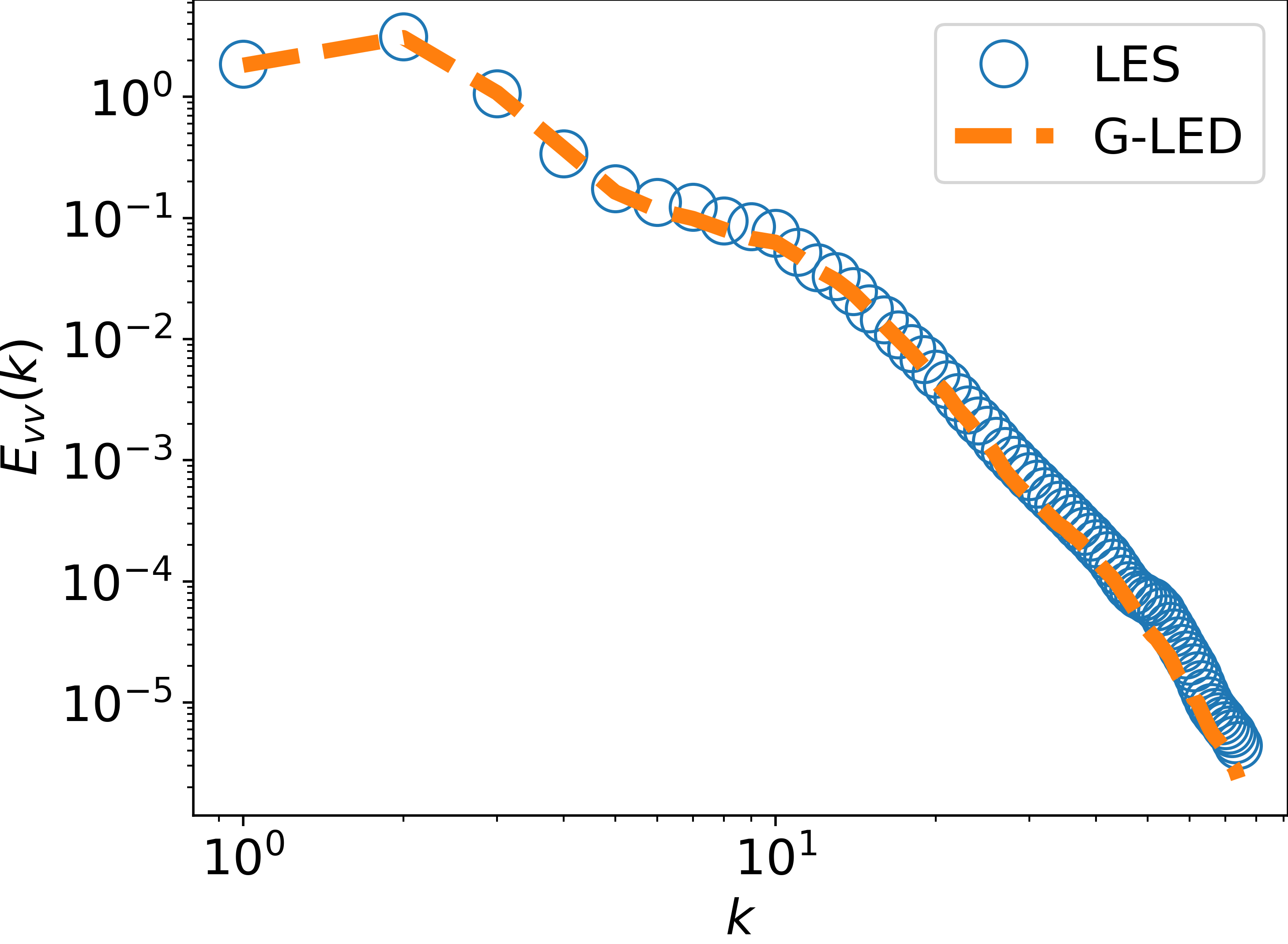}
	\caption{Energy spectrum of wallnormal velocity at $y=0.2, 1,  1.8$ (\textit{from left to right}). }
 \label{fig:v_energy_bfs}
\end{figure}

\begin{figure}[htp]
	\centering
	\includegraphics[width=0.32\textwidth]{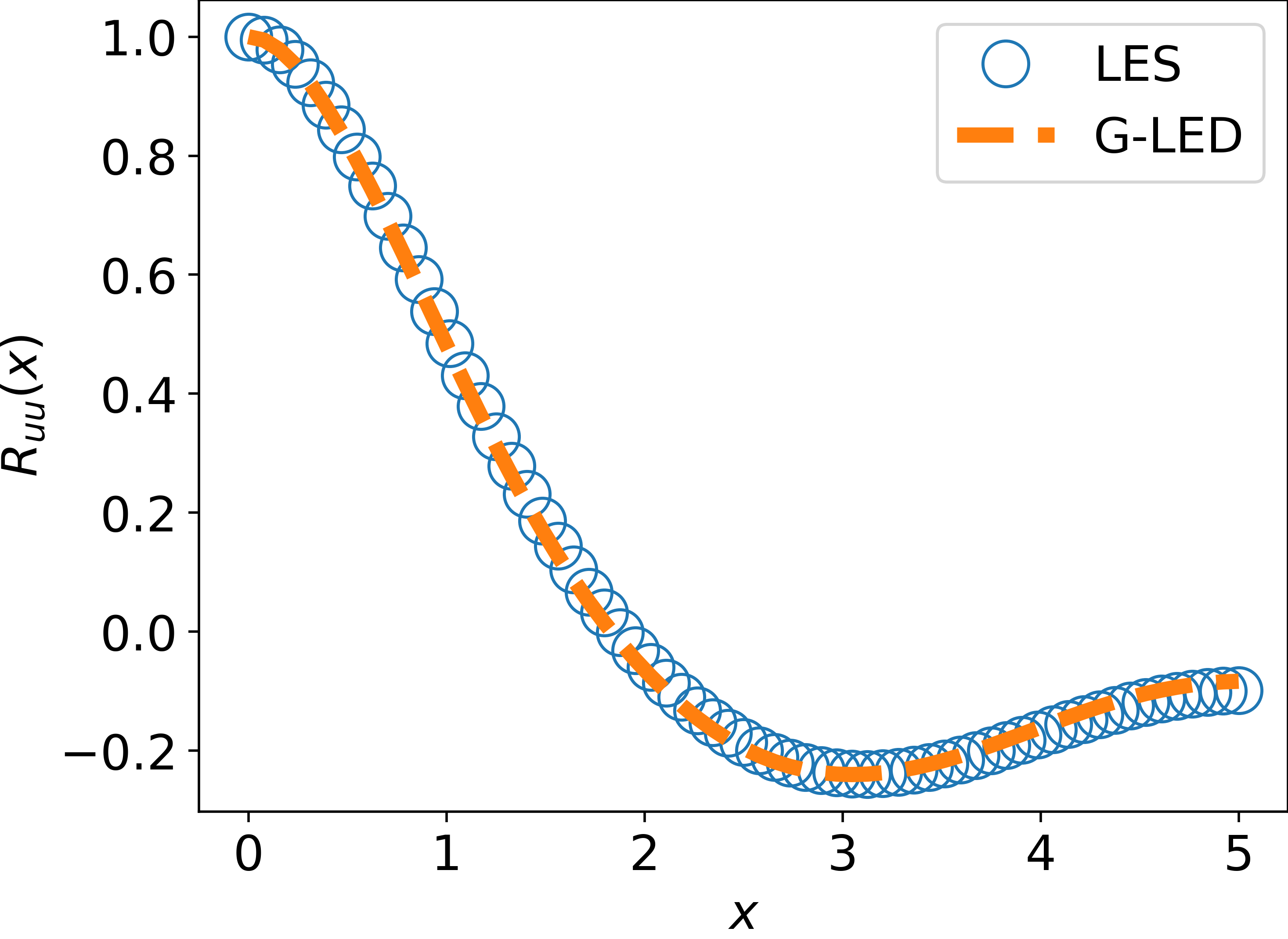}
	\includegraphics[width=0.32\textwidth]{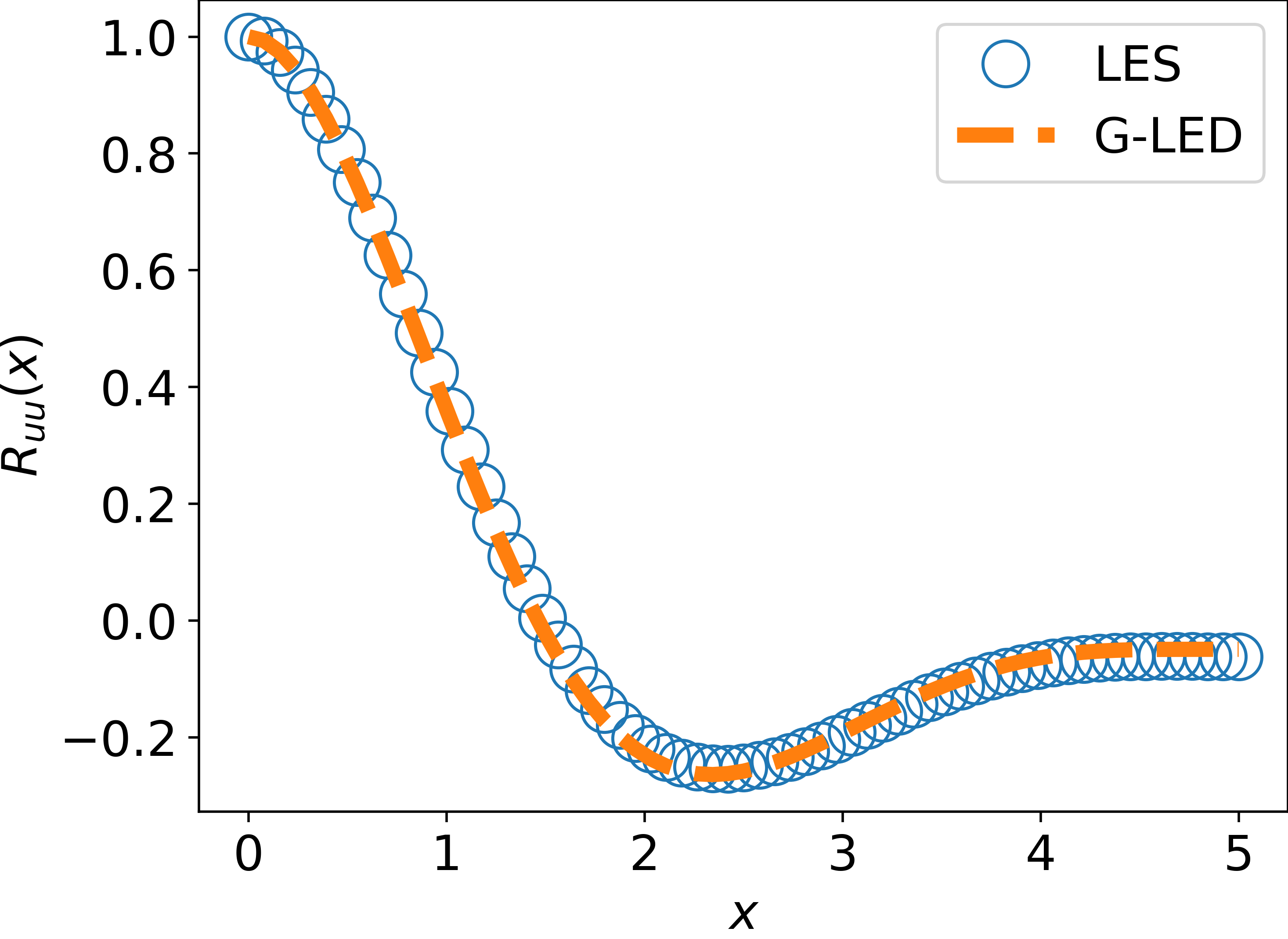}
	\includegraphics[width=0.32\textwidth]{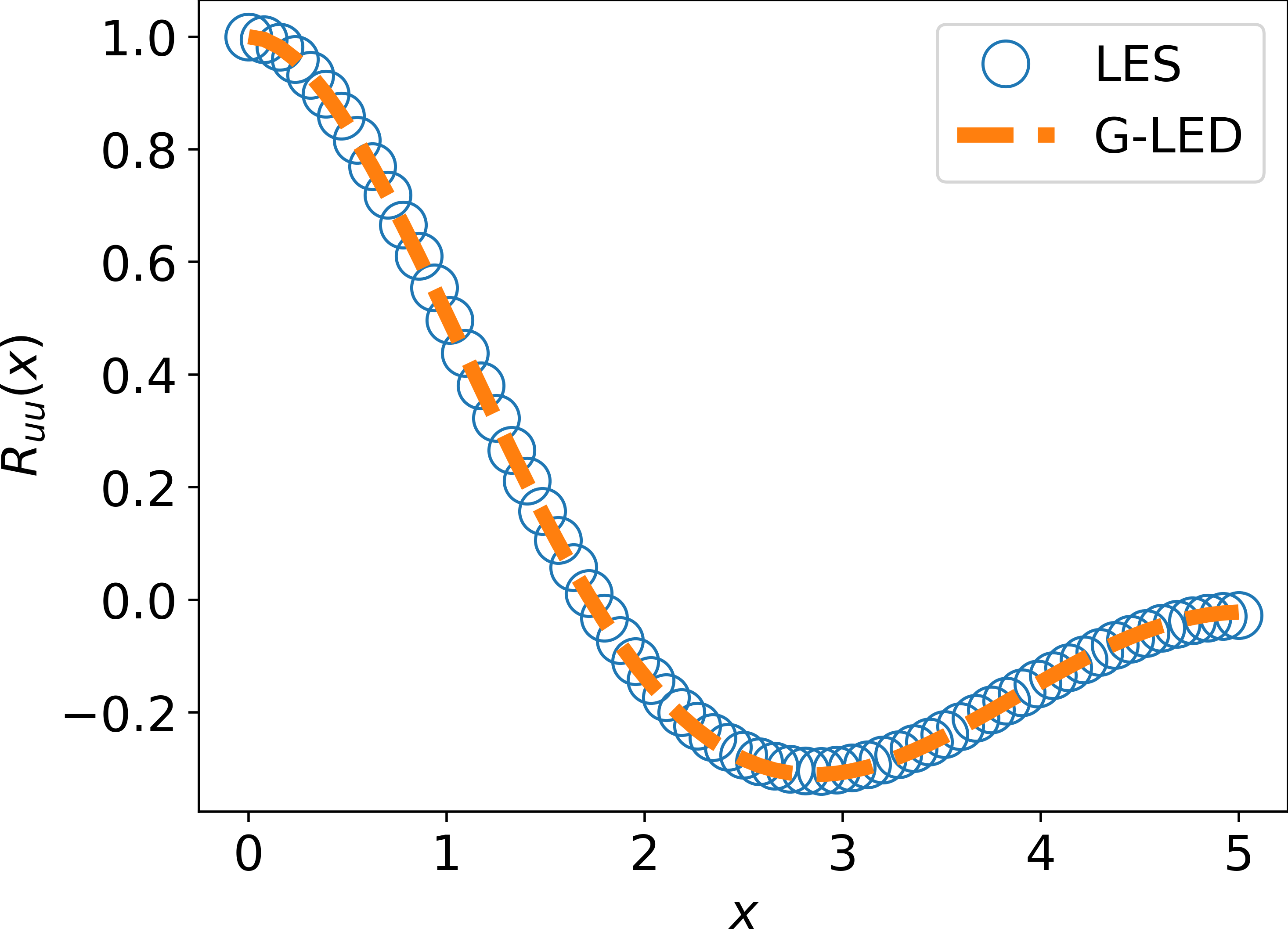}
	\caption{Spatial correlation of streamwise veloctiy at $y=0.2 1, 1.8$ (\textit{from left to right}). } 
 \label{fig:spacor_u_bfs}
\end{figure}

\begin{figure}[htp]
	\centering
	\includegraphics[width=0.32\textwidth]{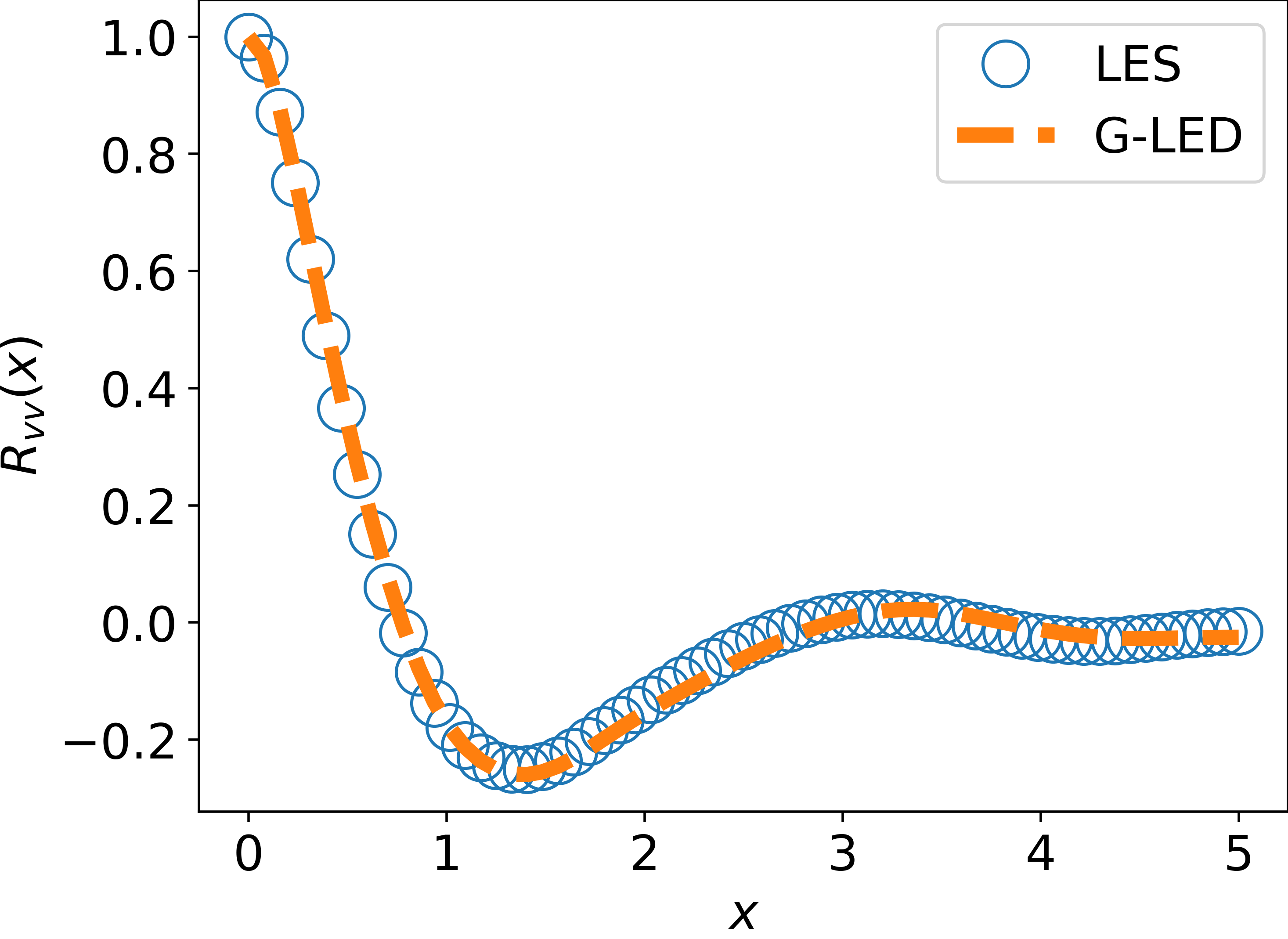}
	\includegraphics[width=0.32\textwidth]{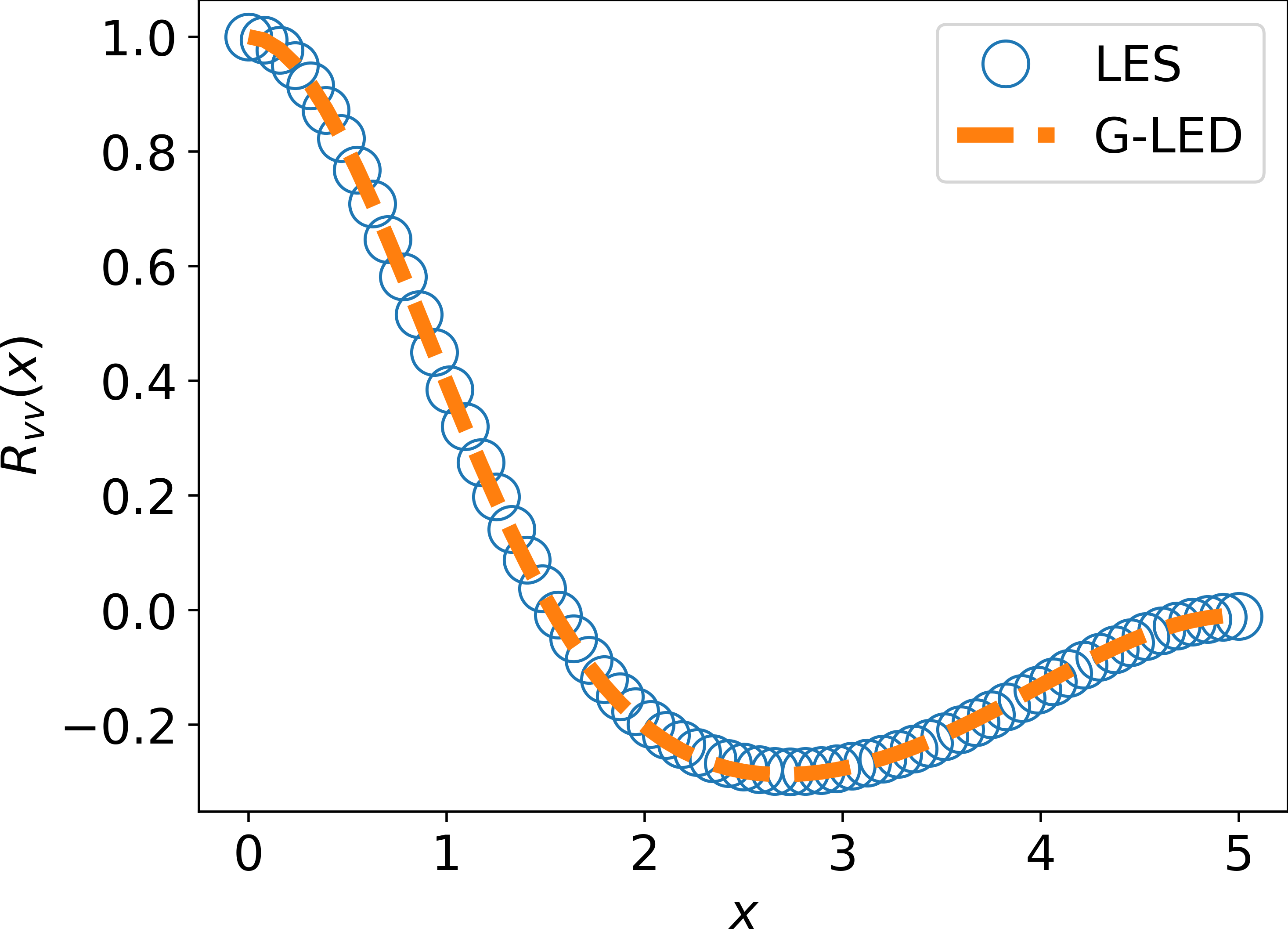}
	\includegraphics[width=0.32\textwidth]{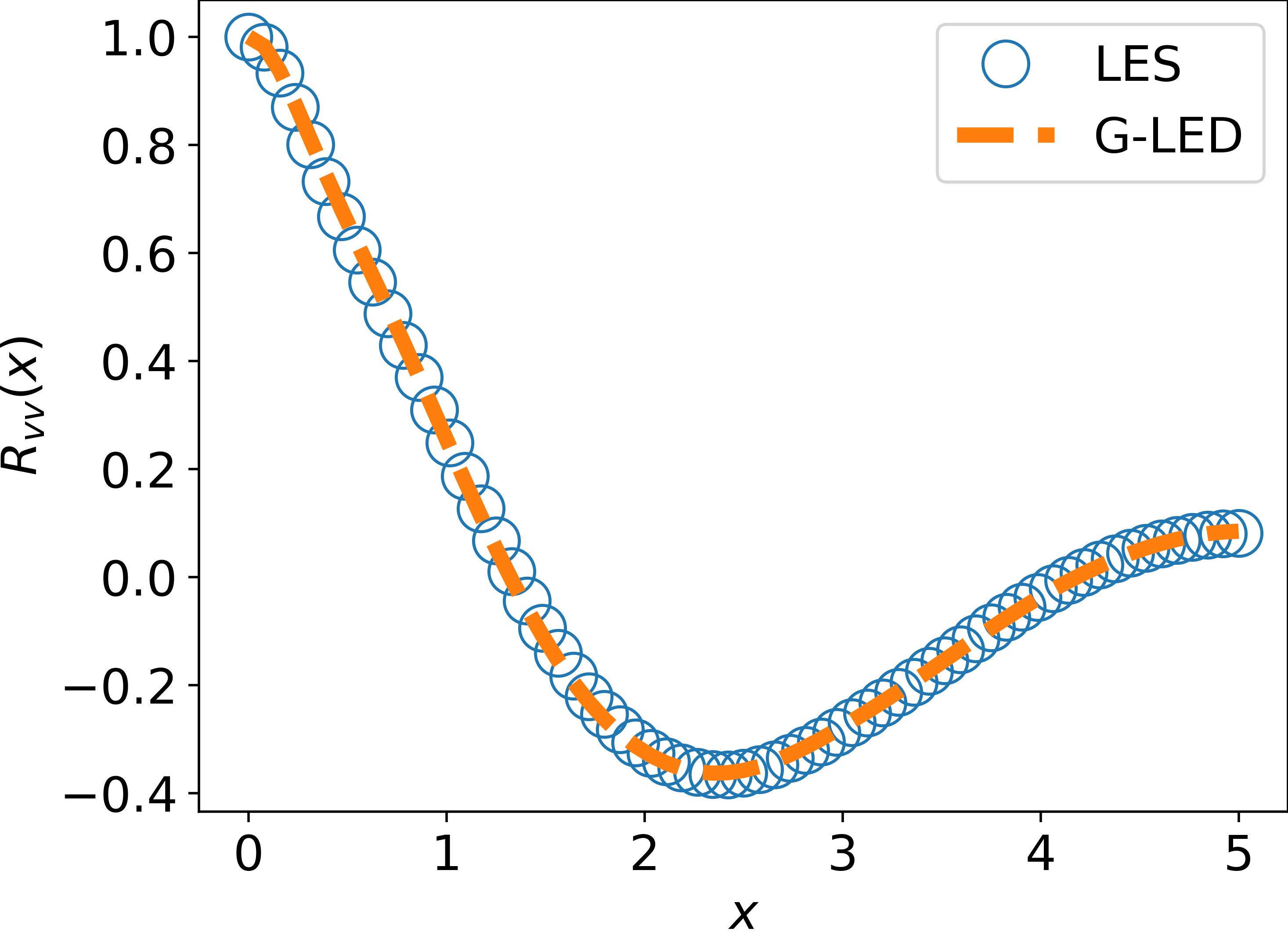}
	\caption{Spatial correlation of wallnormal velcotiy at $y=0.2, 1, 1.8$ (\textit{from left to right}). } 
  \label{fig:spacor_v_bfs}
\end{figure}

\begin{figure}[htp]
	\centering
	\includegraphics[width=0.32\textwidth]{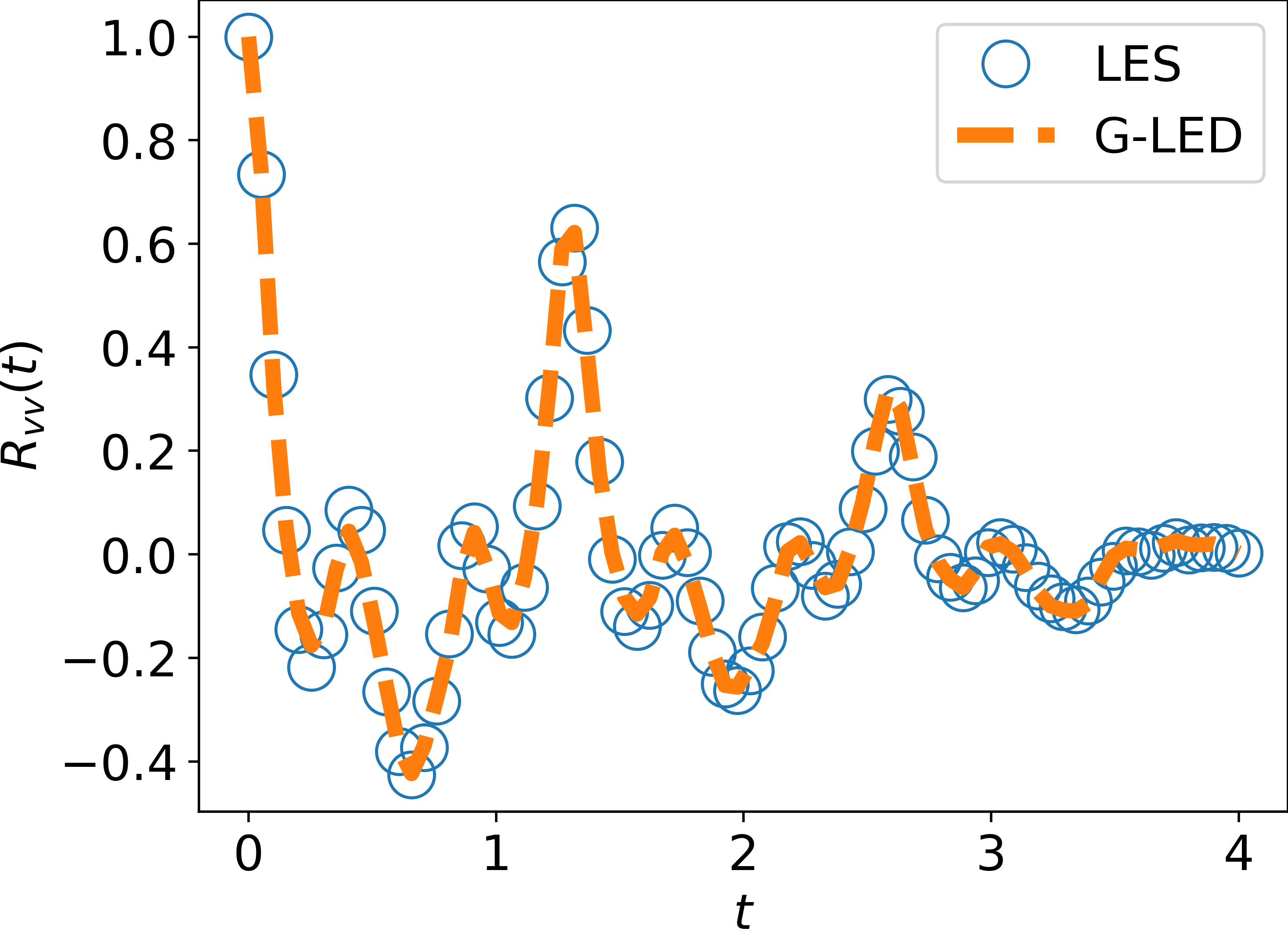}
	\includegraphics[width=0.32\textwidth]{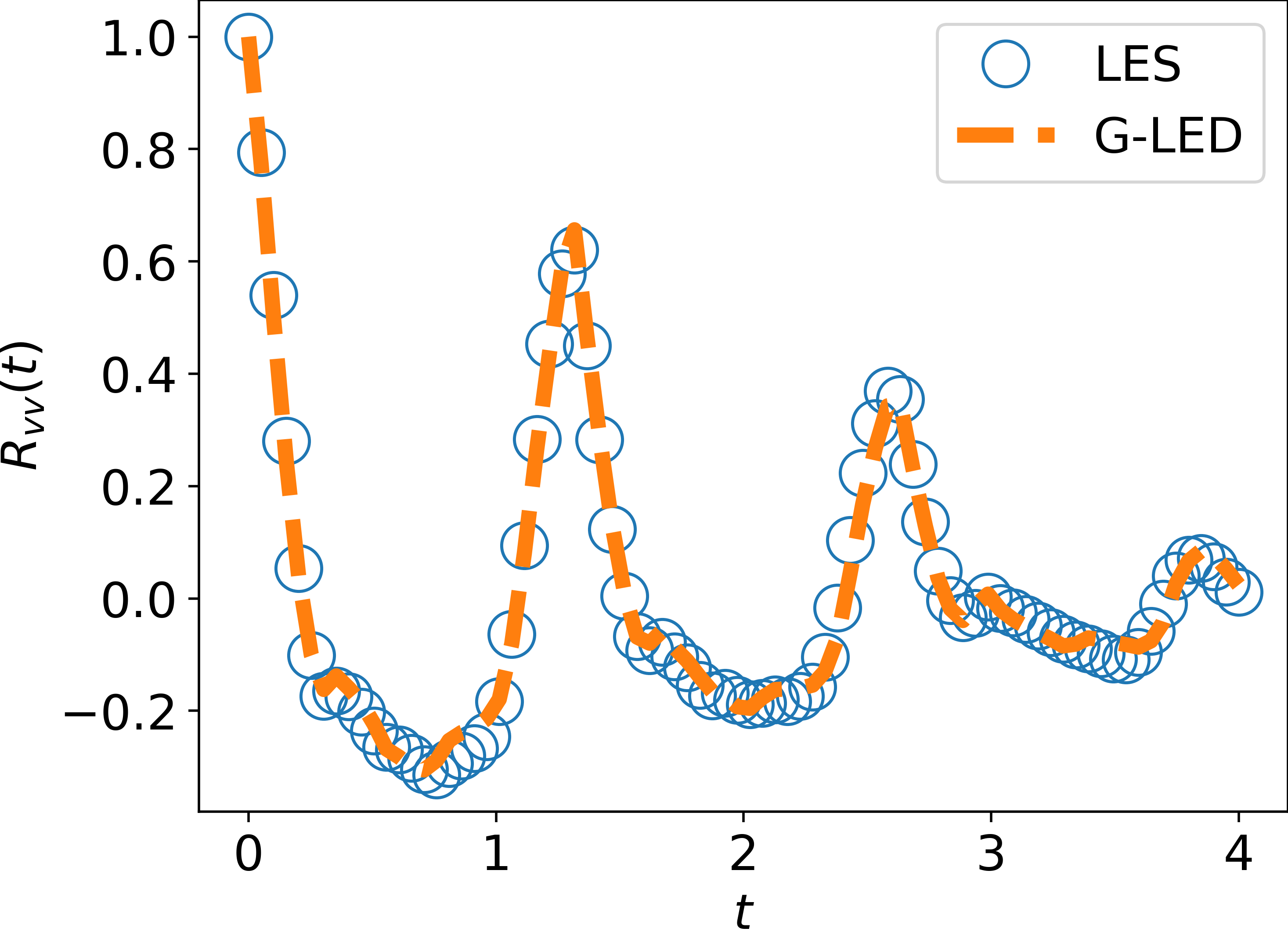}
	\includegraphics[width=0.32\textwidth]{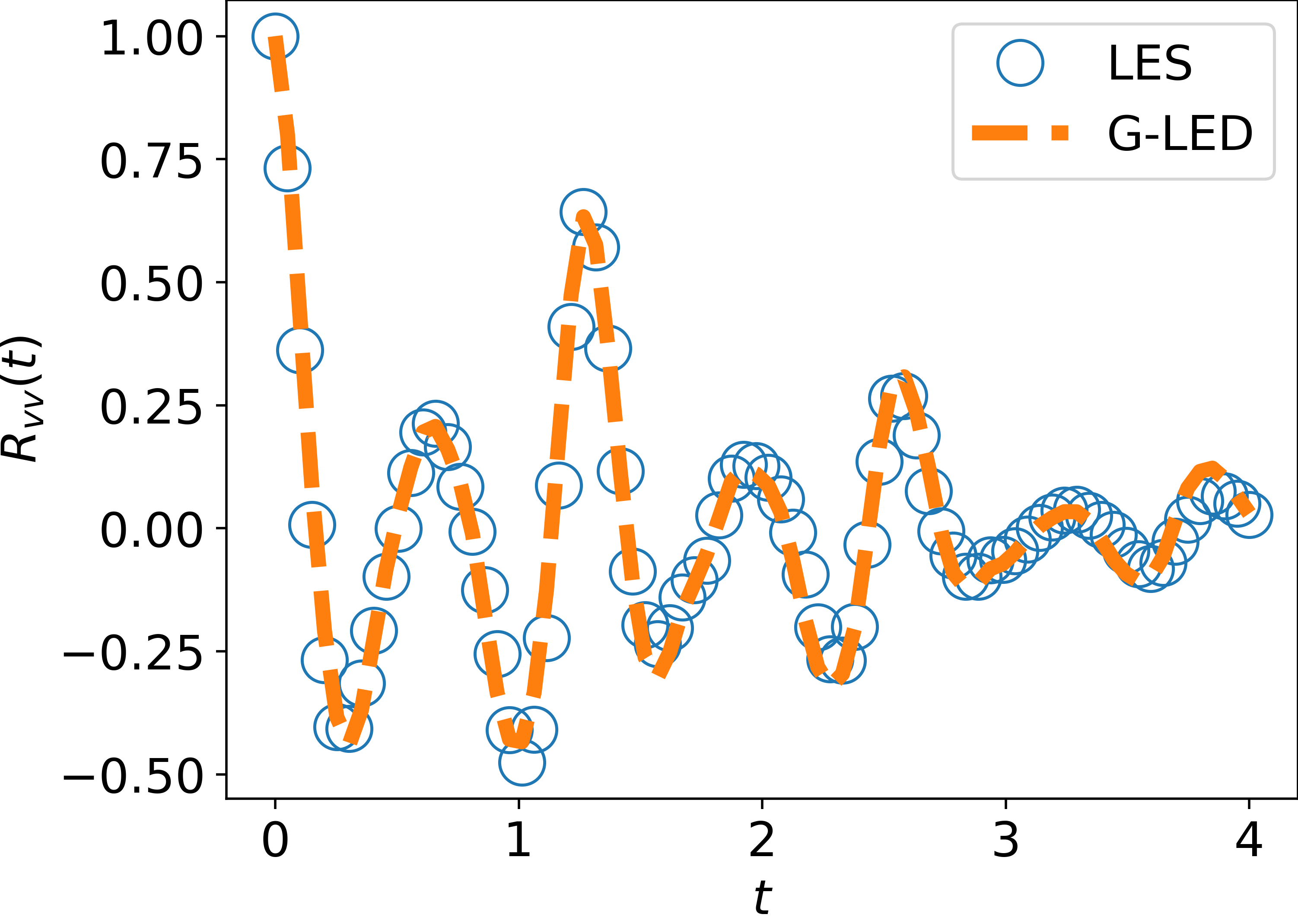}
	\caption{Temporal correlation of wallnormal velcotiy at $(x,y)=(0.6,0.6),(0.6,1.8), (1.8, 1.8)$ (\textit{from left to right}). } 
 \label{fig:tmpcor_v_bfs}
\end{figure}

\section{Simulation details and additional results for turbulent channel flow}
\label{sec:channel_appendix}
The domain of interest is the entire $4\times2\times2$ cuboid, which is discretized by $40\times 50 \times 30$ spatially and $\delta t = 0.2$ temporally. The LES is detailed in \cite{it:2015-014}, which serves the truth of the micro-level dynamic. The training, validation, and test data sets consist of $50000$, $1000$, and $5000$ snapshots, respectively, with a macro time-step size of $\Delta t = 4$. 

\begin{figure}[htp]
	\centering
	\includegraphics[width = 0.15\textwidth]{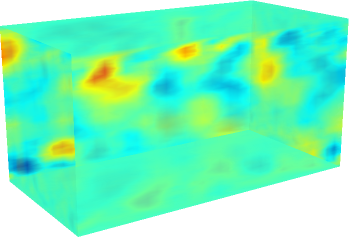}
    \includegraphics[width = 0.15\textwidth]{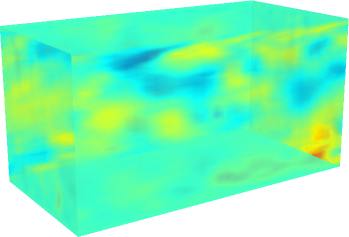}
    \includegraphics[width = 0.15\textwidth]{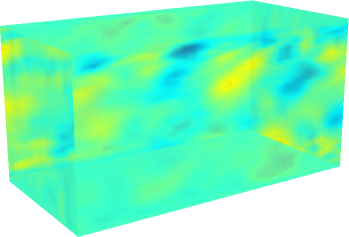}
    \includegraphics[width = 0.15\textwidth]{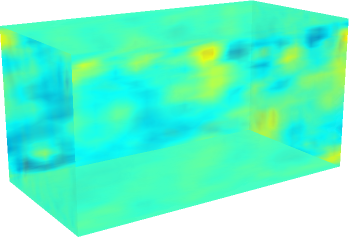}
    \includegraphics[width = 0.15\textwidth]{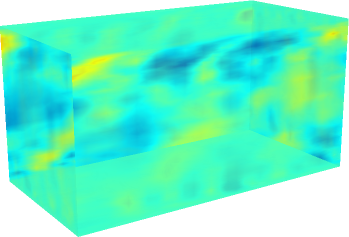}
    \includegraphics[width = 0.15\textwidth]{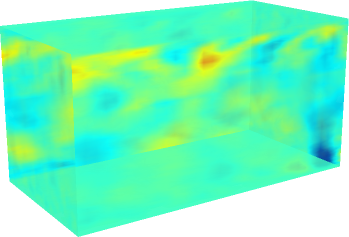}
    \vfil
    \includegraphics[width = 0.15\textwidth]{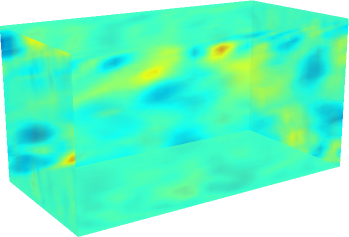}
    \includegraphics[width = 0.15\textwidth]{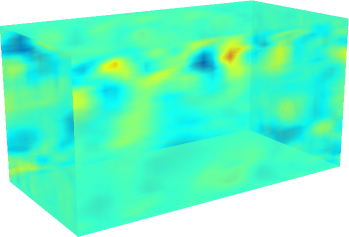}
    \includegraphics[width = 0.15\textwidth]{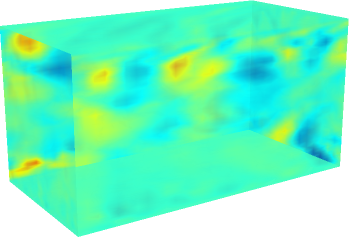}
    \includegraphics[width = 0.15\textwidth]{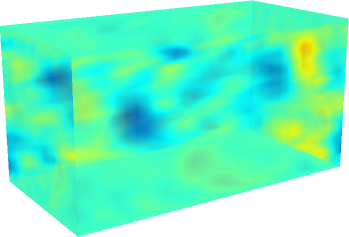}
    \includegraphics[width = 0.15\textwidth]{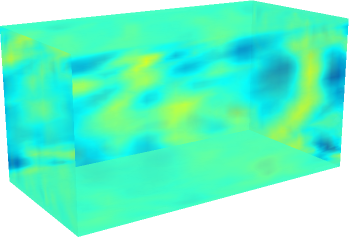}
    \includegraphics[width = 0.15\textwidth]{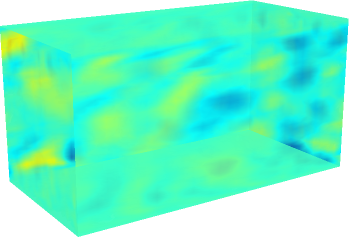}
    \colorbarMatlabJet{-0.025}{-0.0125}{0}{0.0125}{0.025}
\caption{Volume plots of wallnormal velocity for the turbulent channel flow from G-LED (\textit{top row}) and LES (\textit{bottom row}) at $t=60,120,180,240,300,360$ which is equivalent to flow-through time at $t_{\mathrm{ft}}=2.27, 4.55, 6.82, 9.10, 11.3, 13.65$ (\textit{from left to right}). }
\label{fig:channel_flow_v_contour}
\end{figure}

\begin{figure}[htp]
	\centering
    \includegraphics[width = 0.15\textwidth]{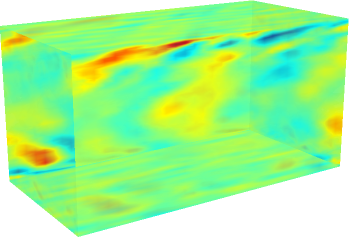}
    \includegraphics[width = 0.15\textwidth]{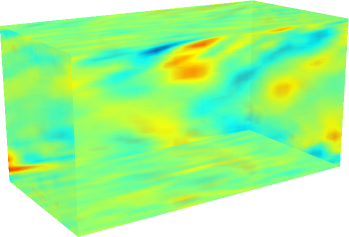}
    \includegraphics[width = 0.15\textwidth]{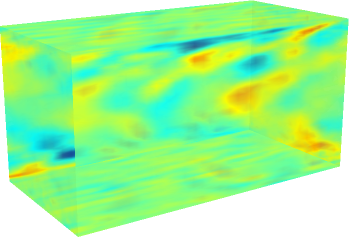}
    \includegraphics[width = 0.15\textwidth]{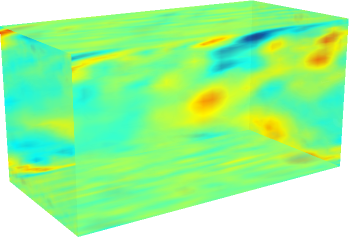}
    \includegraphics[width = 0.15\textwidth]{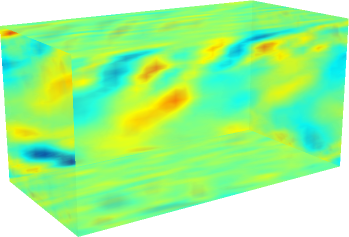}
    \includegraphics[width = 0.15\textwidth]{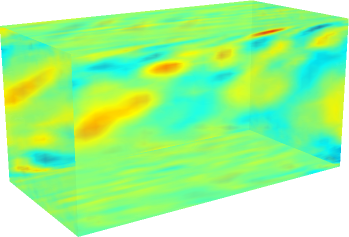}
	\vfil
	\includegraphics[width = 0.15\textwidth]{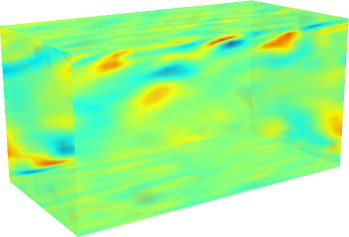}
    \includegraphics[width = 0.15\textwidth]{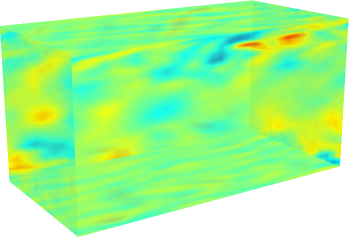}
    \includegraphics[width = 0.15\textwidth]{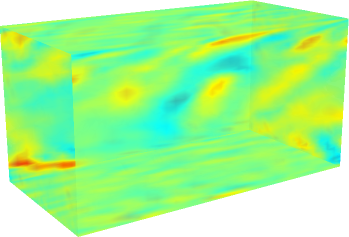}
    \includegraphics[width = 0.15\textwidth]{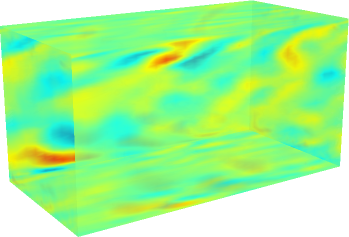}
    \includegraphics[width = 0.15\textwidth]{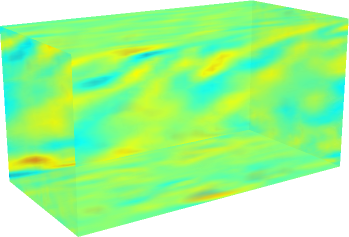}
    \includegraphics[width = 0.15\textwidth]{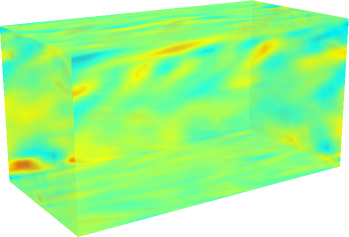}
	\vfil
	\colorbarMatlabJet{-0.04}{-0.02}{0}{0.02}{0.04}
\caption{Volume plots of spanwise velocity for the turbulent channel flow from G-LED (\textit{top row}) and LES (\textit{bottom row}) at $t=60,120,180,240,300,360$ which is equivalent to flow-through time at $t_{\mathrm{ft}}=2.27, 4.55, 6.82, 9.10, 11.3, 13.65$ (\textit{from left to right}). }
\label{fig:channel_flow_w_contour}
\end{figure}

\begin{figure}[htp]
	\centering
	\includegraphics[width=0.32\textwidth]{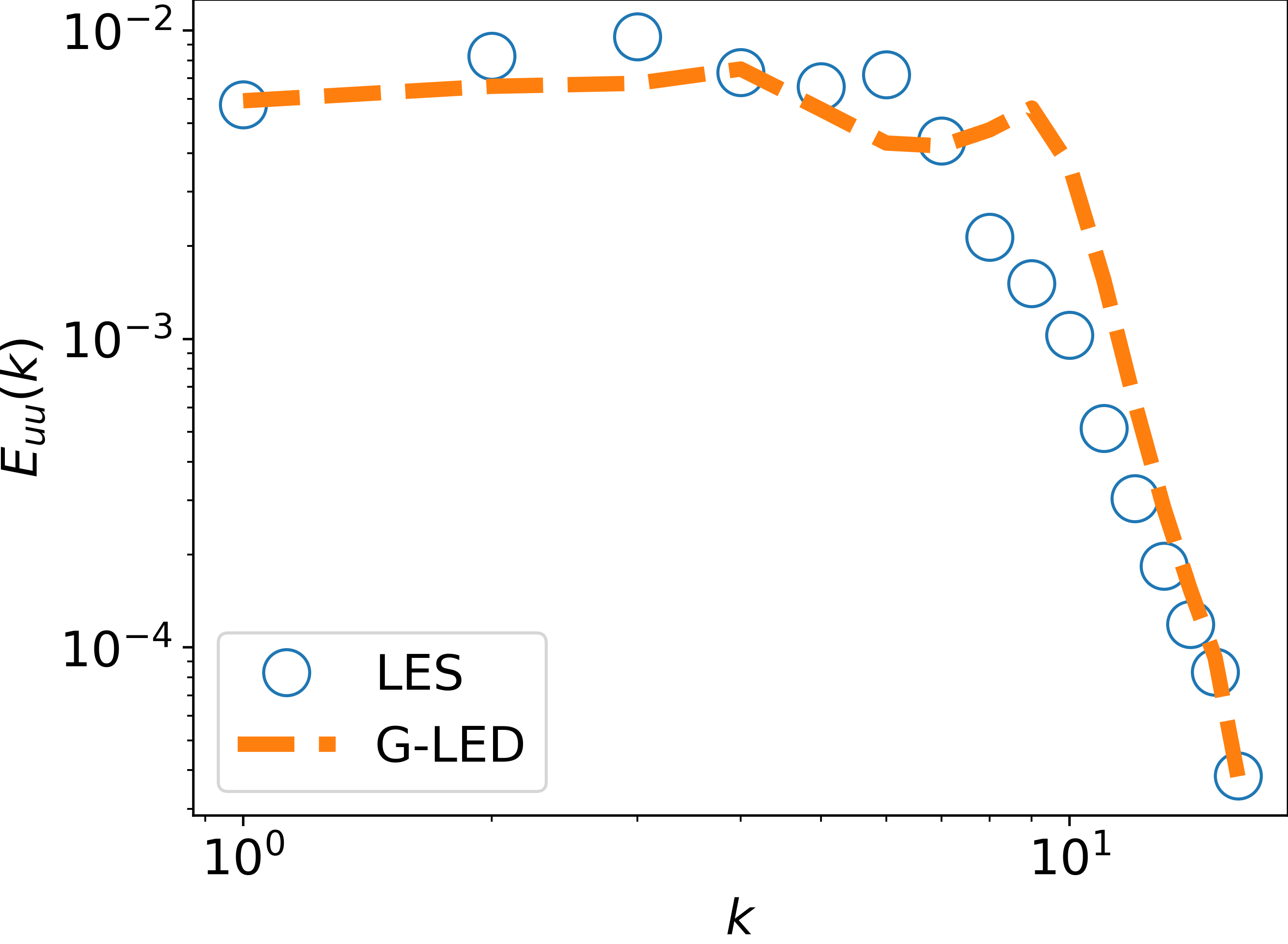}
	\includegraphics[width=0.32\textwidth]{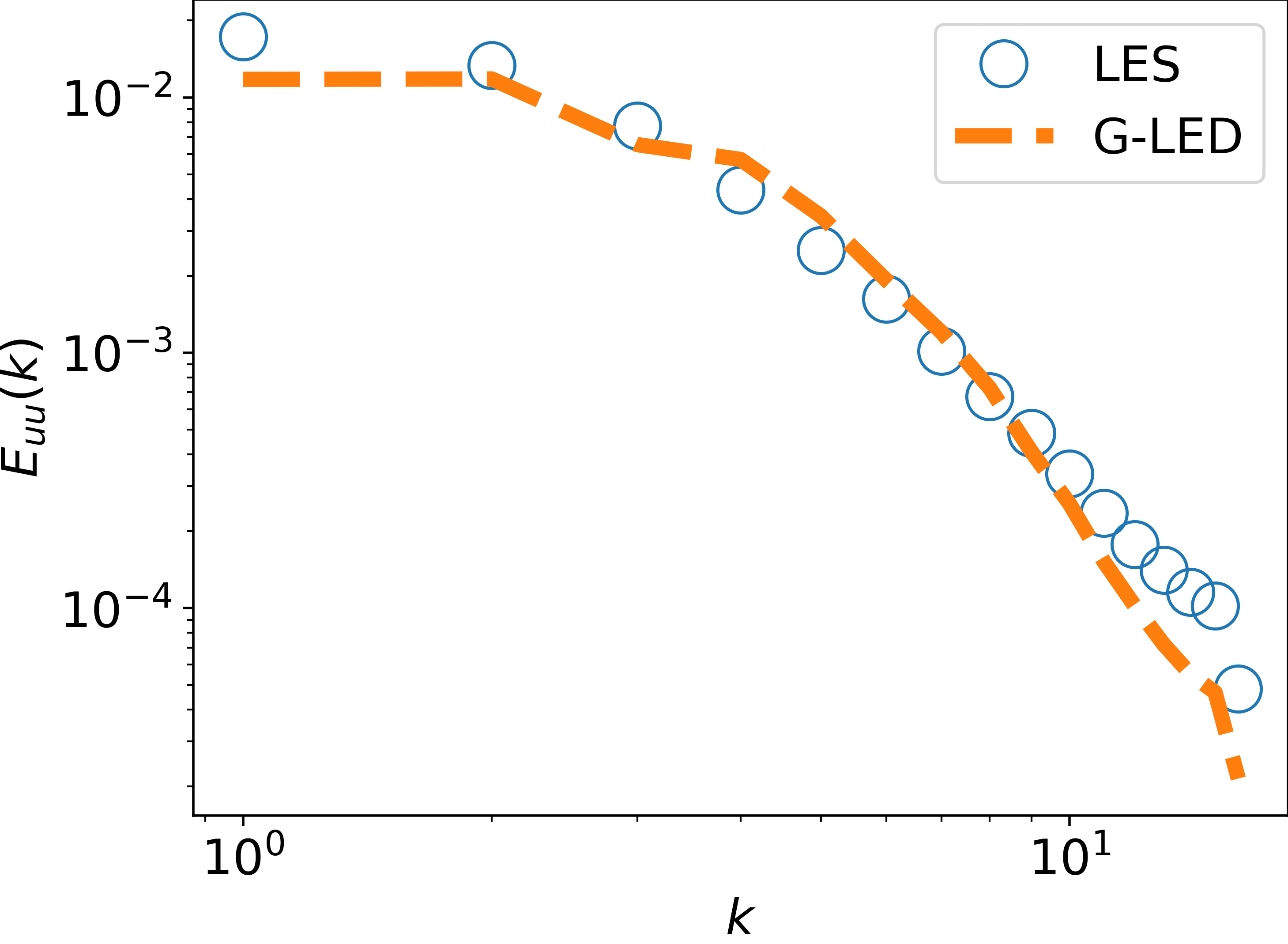}
	\includegraphics[width=0.32\textwidth]{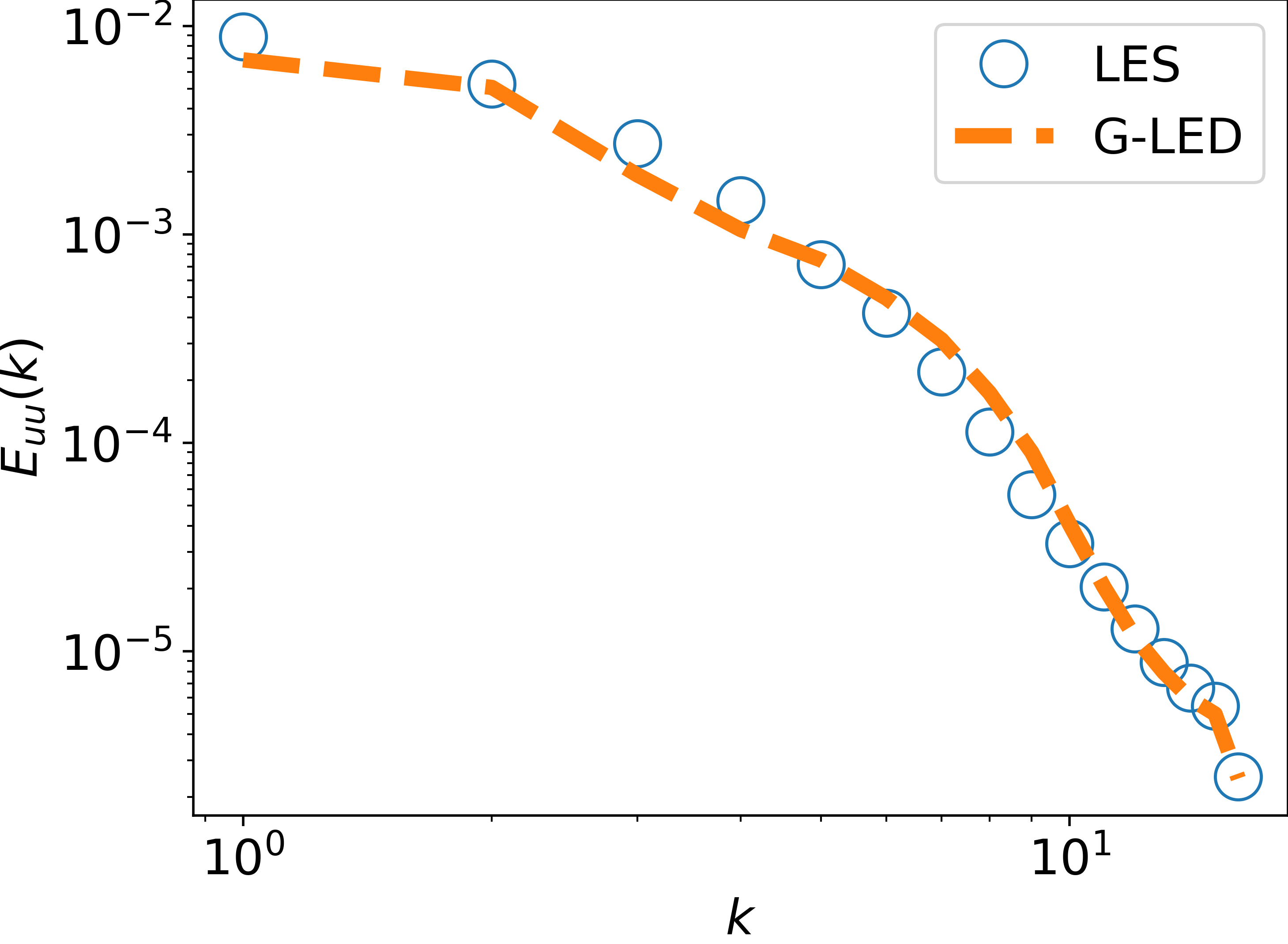}
	\vfil
	\includegraphics[width=0.32\textwidth]{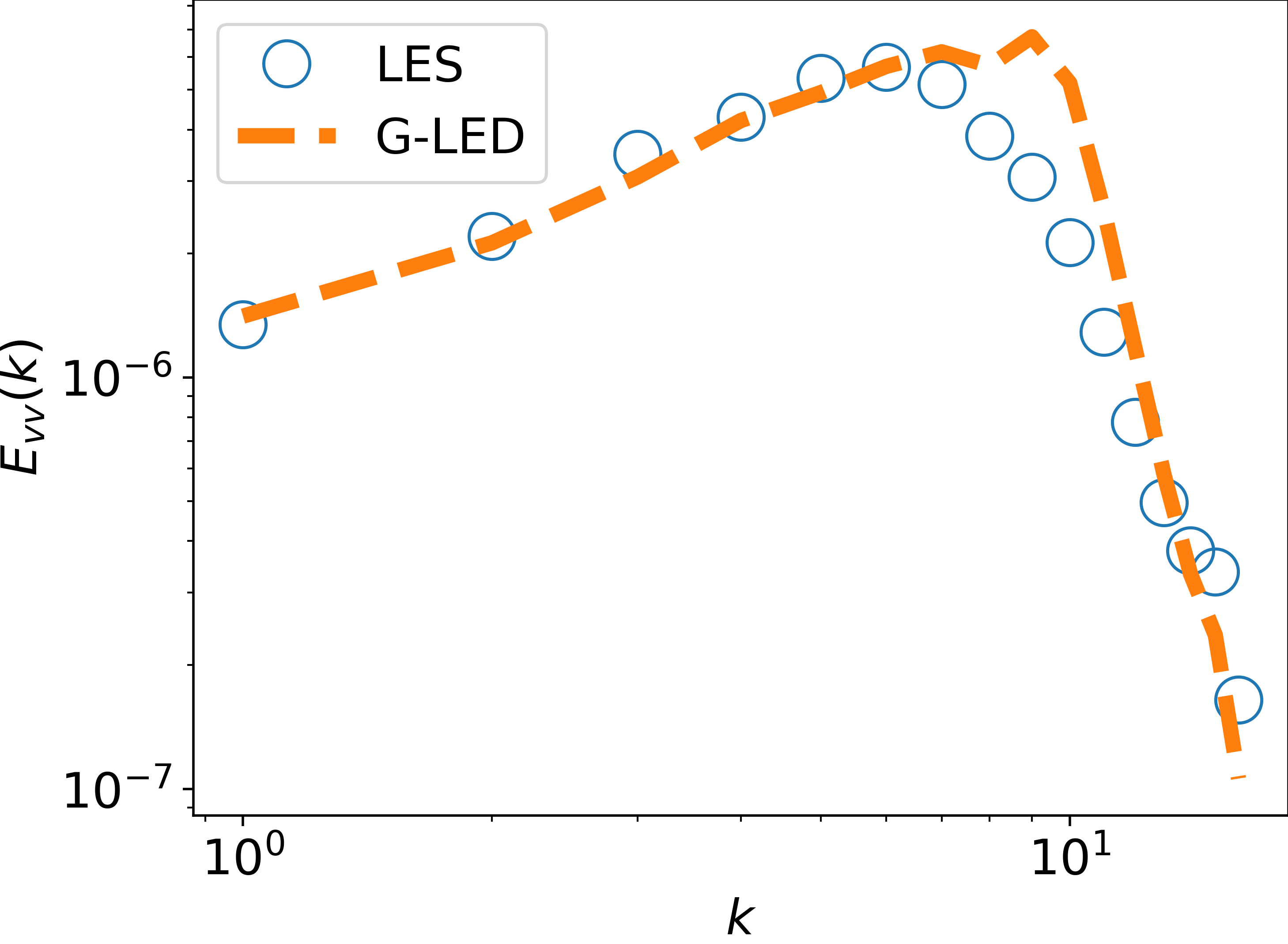}
	\includegraphics[width=0.32\textwidth]{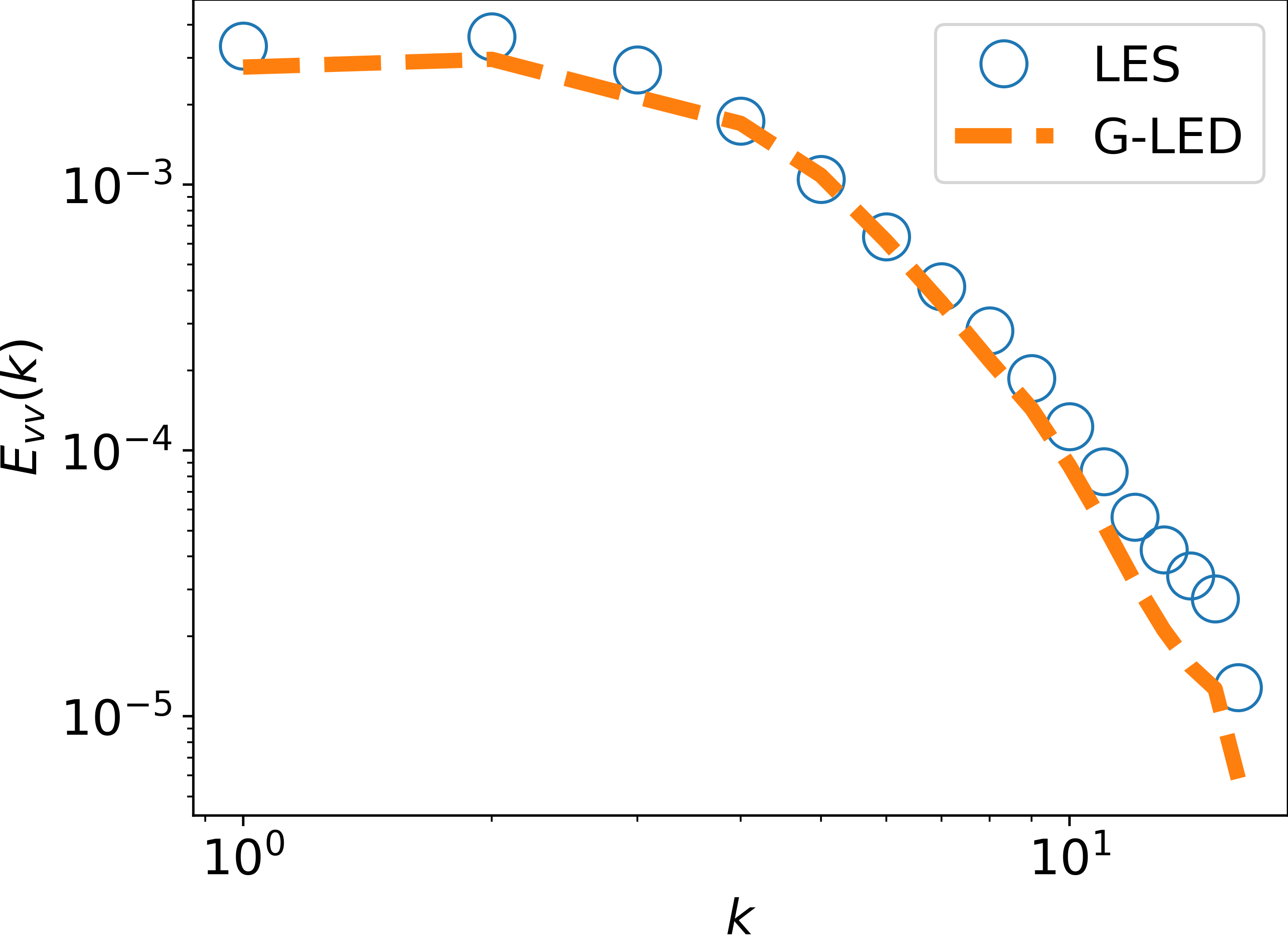}
	\includegraphics[width=0.32\textwidth]{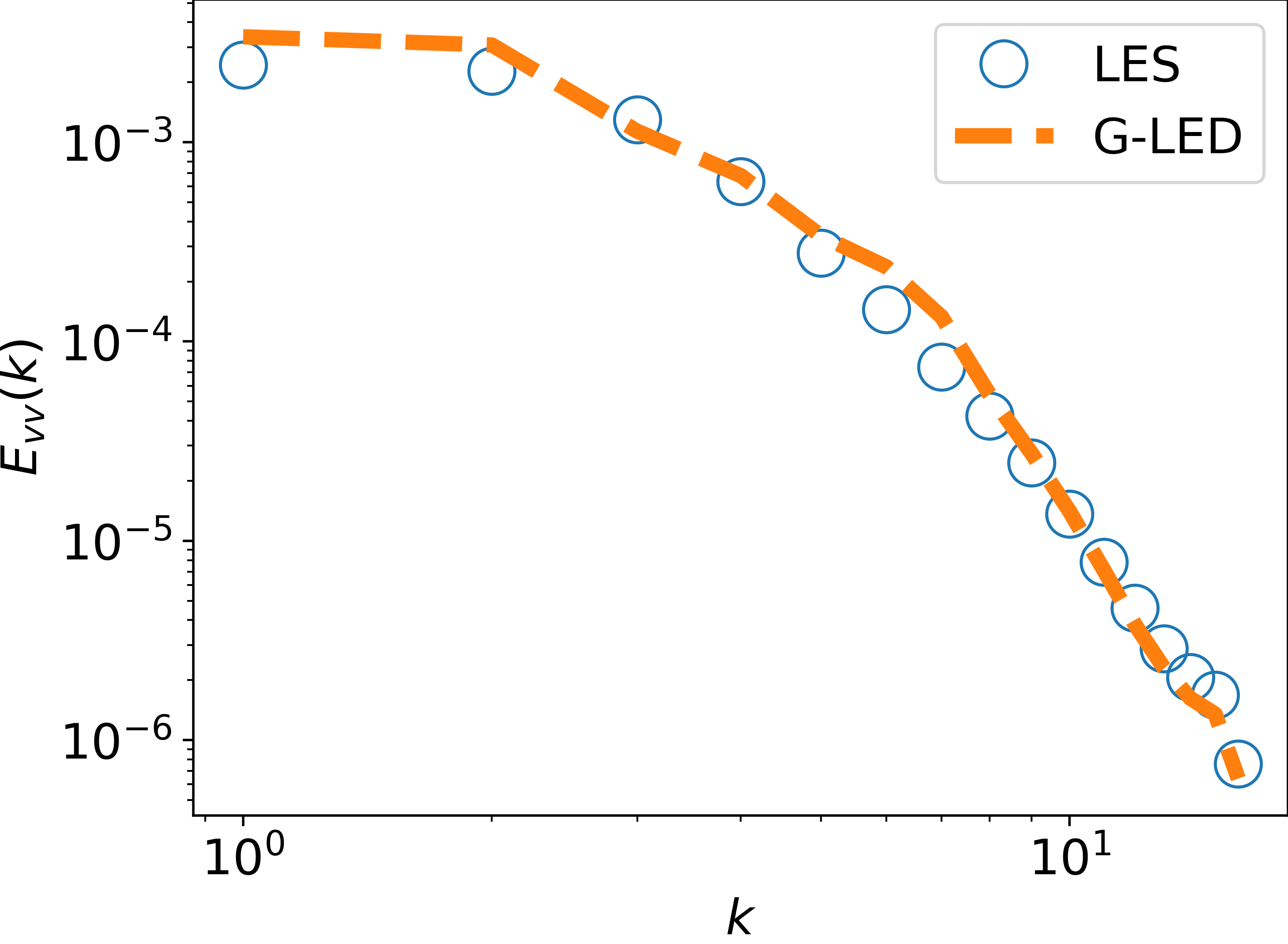}
	\vfil
	\includegraphics[width=0.32\textwidth]{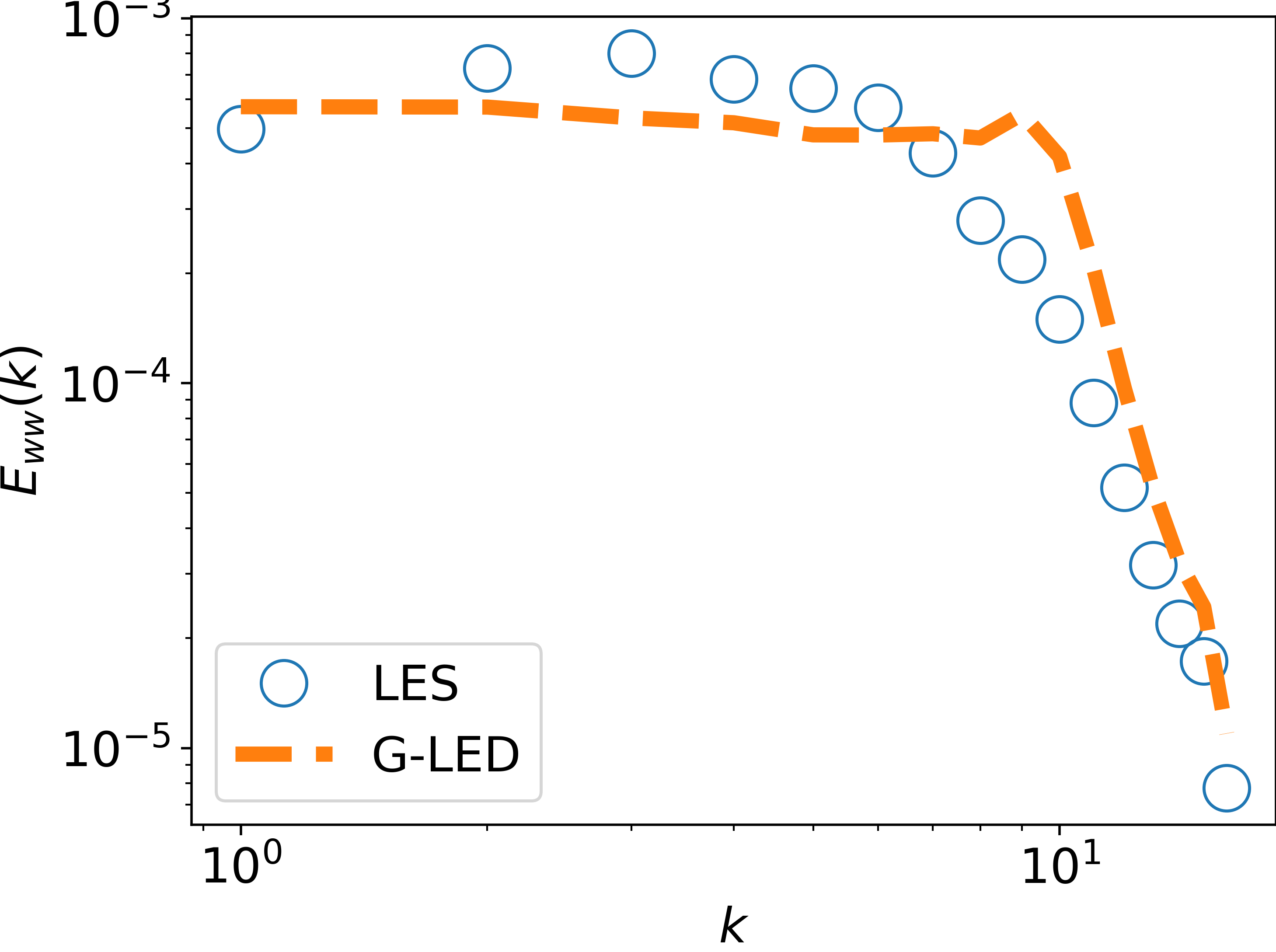}
	\includegraphics[width=0.32\textwidth]{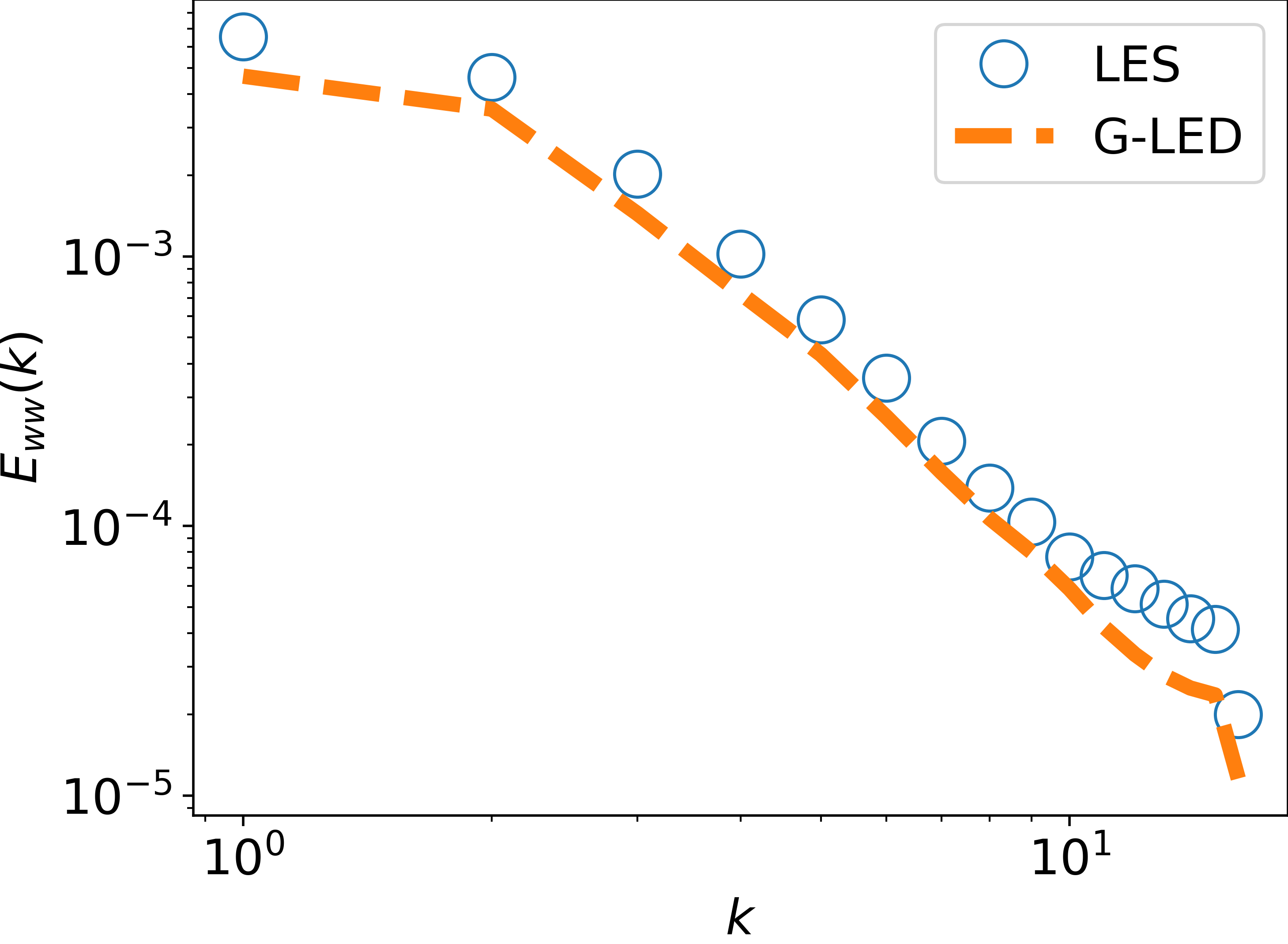}
	\includegraphics[width=0.32\textwidth]{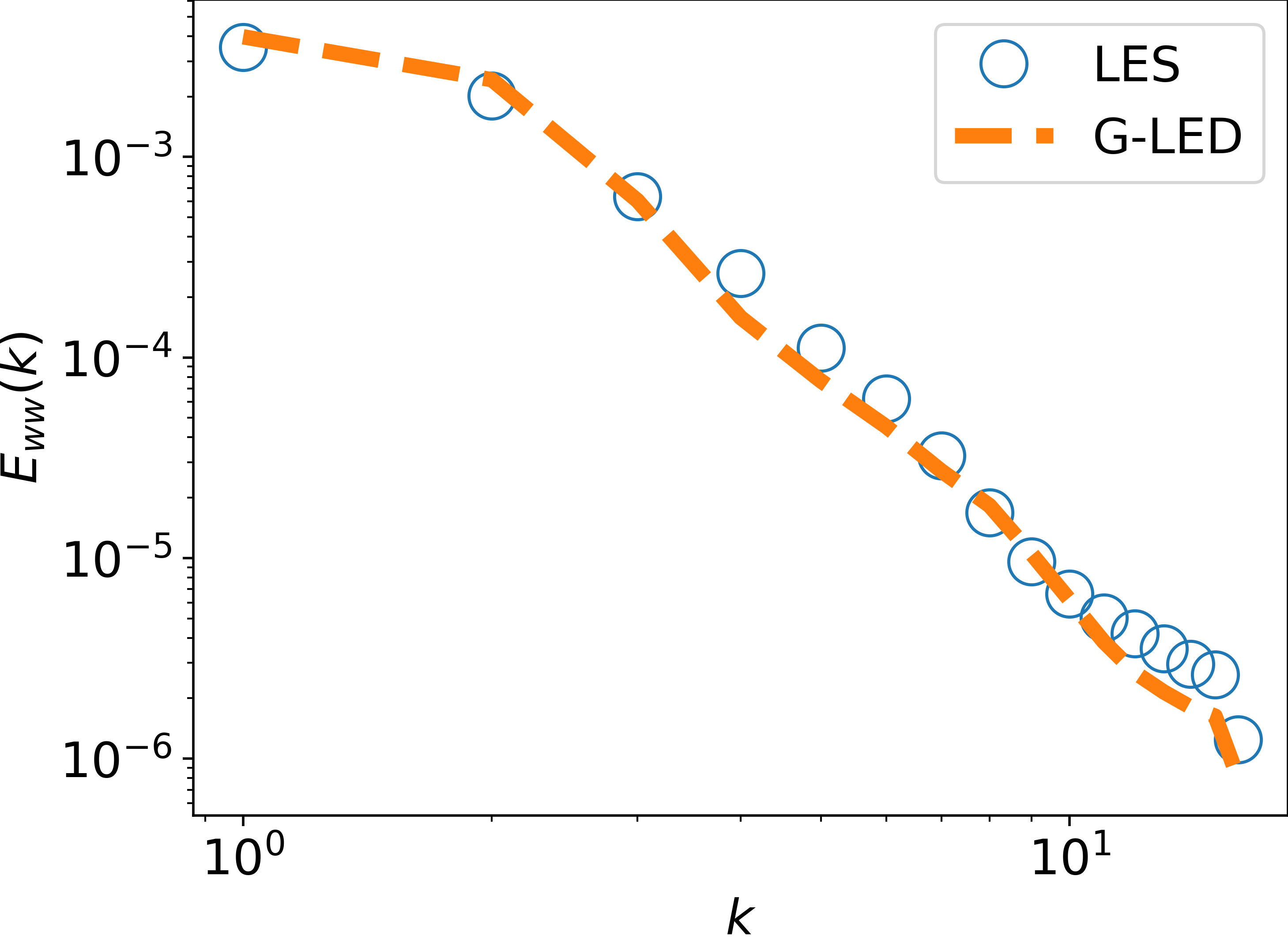}
	\caption{Energy spectrum at $y+=4.57, 171, 320$ (\textit{from left to right}) for streamwise, wallnormall and spanwise velocity (\textit{from top to bottom}). } 
 \label{fig:channel_energy_spectrum}
\end{figure}

\begin{figure}[htp]
	\centering
	\includegraphics[width=0.32\textwidth]{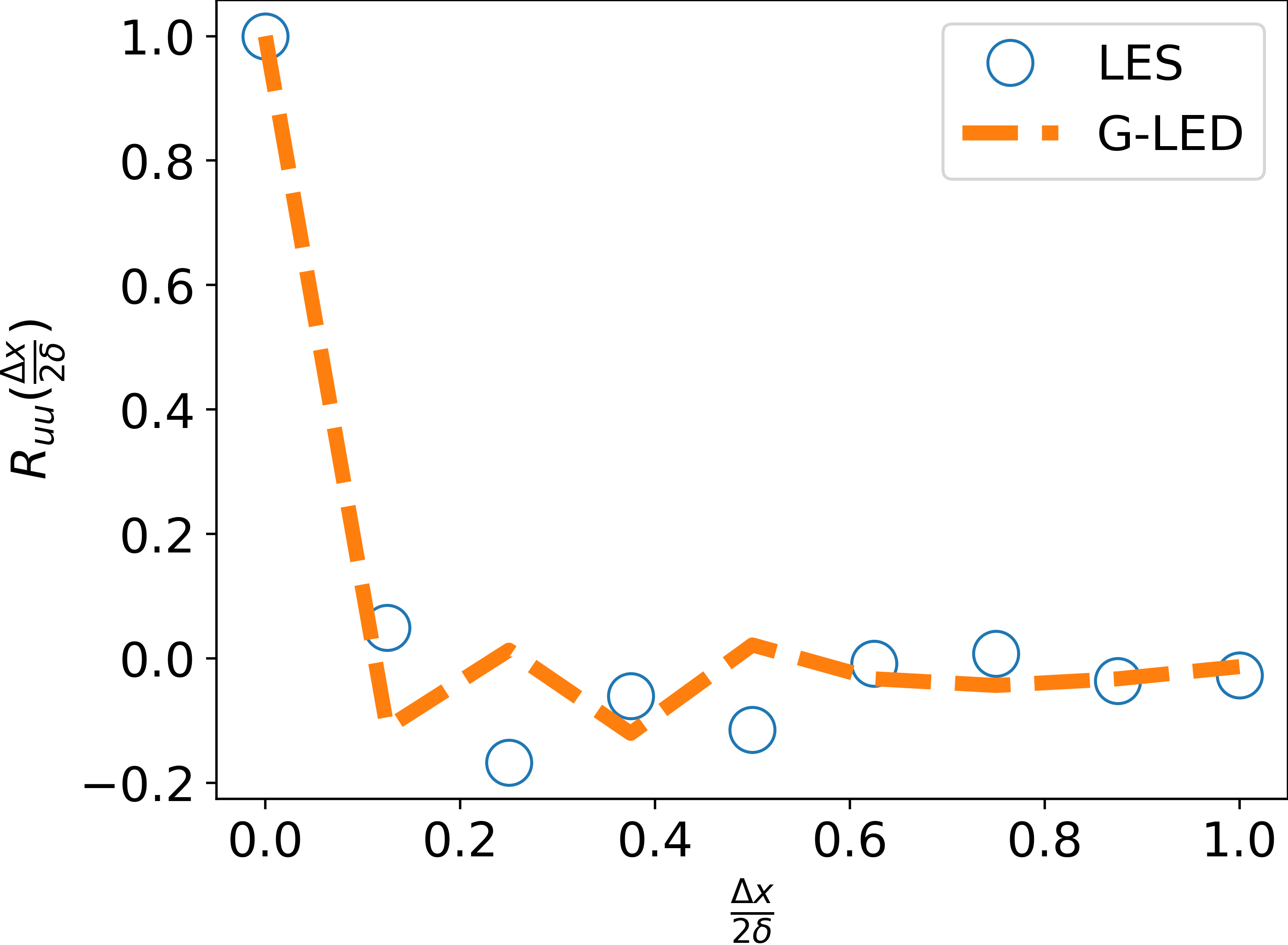}
	\includegraphics[width=0.32\textwidth]{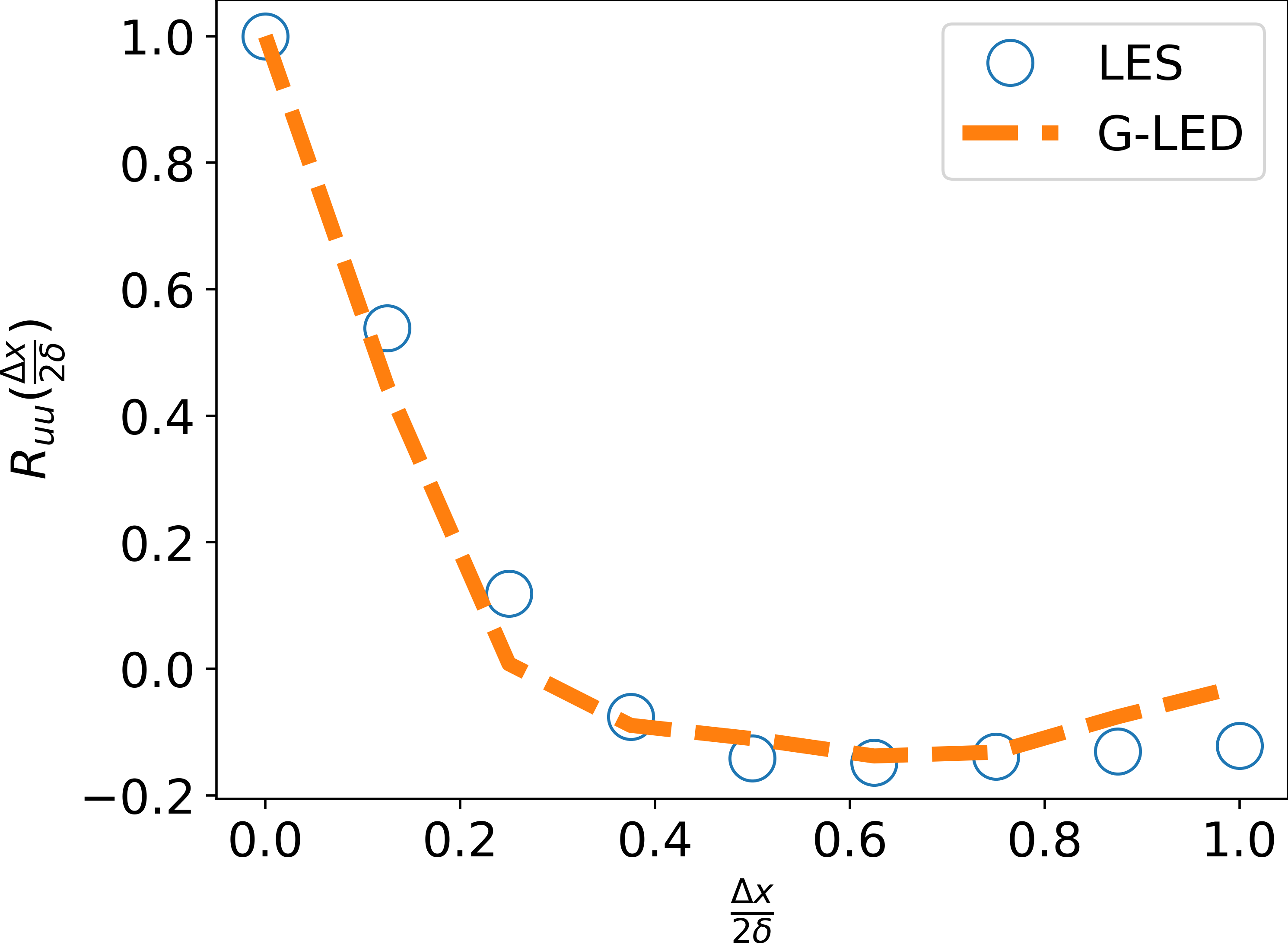}
	\includegraphics[width=0.32\textwidth]{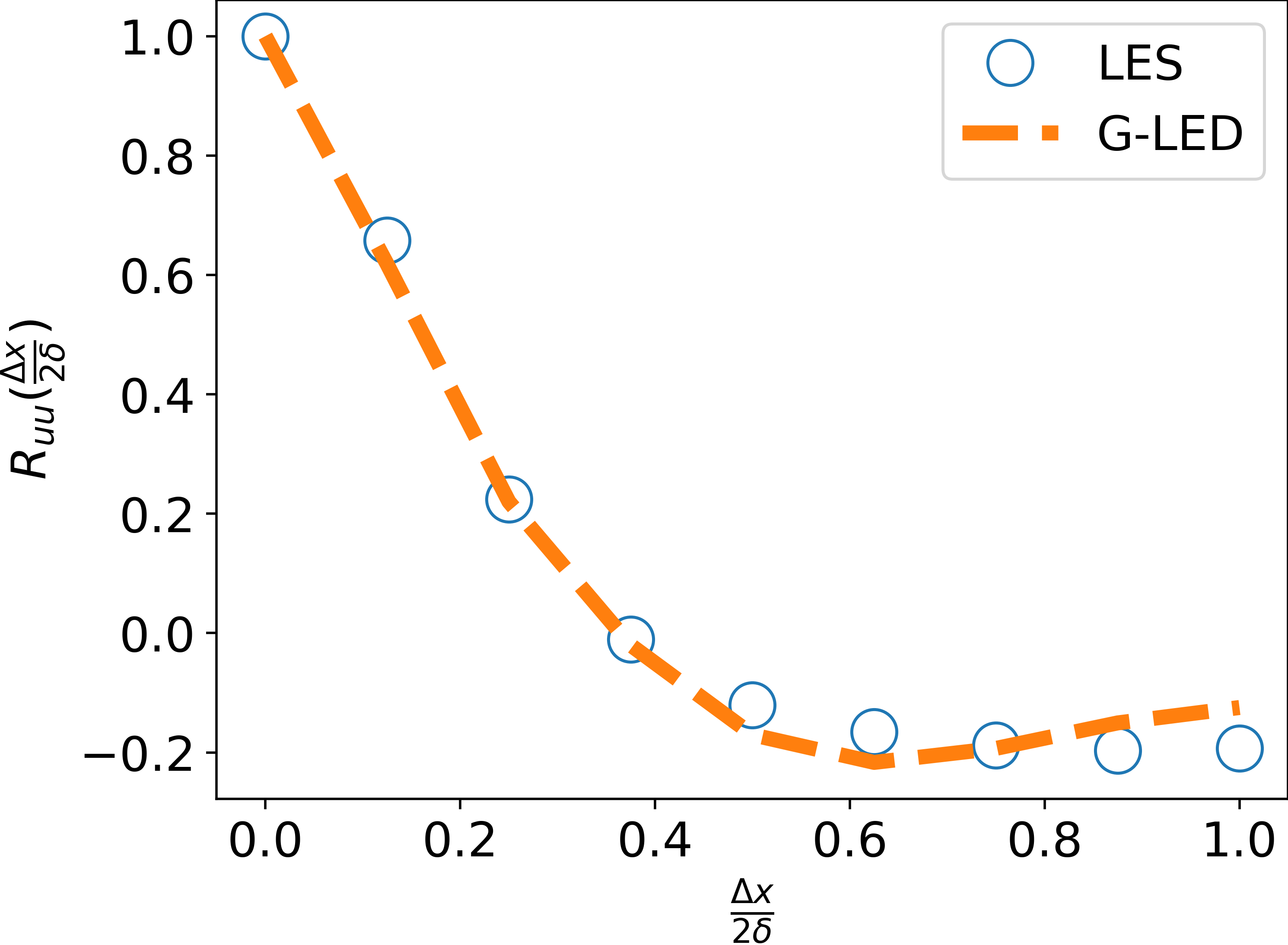}
	\vfil
	\includegraphics[width=0.32\textwidth]{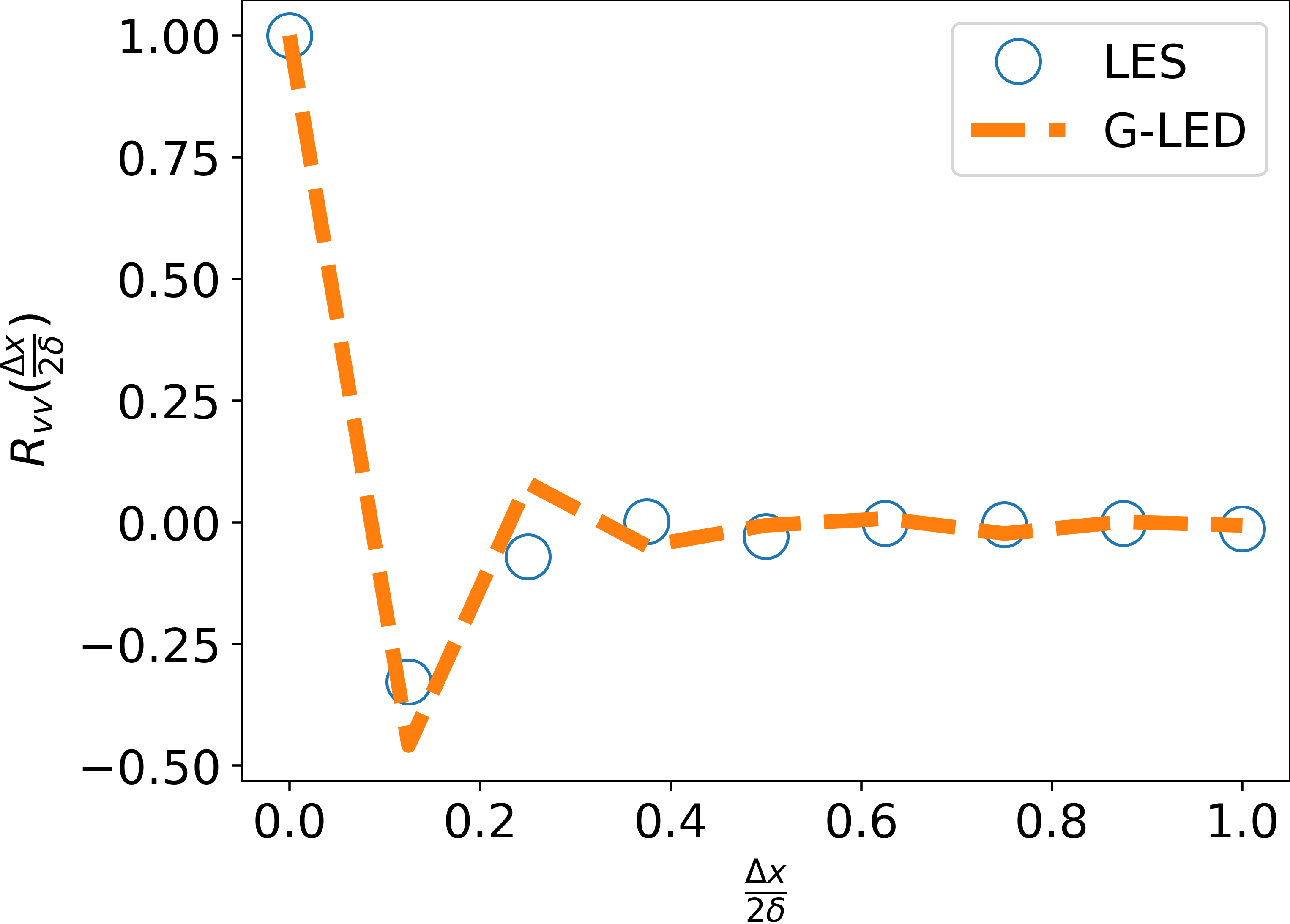}
	\includegraphics[width=0.32\textwidth]{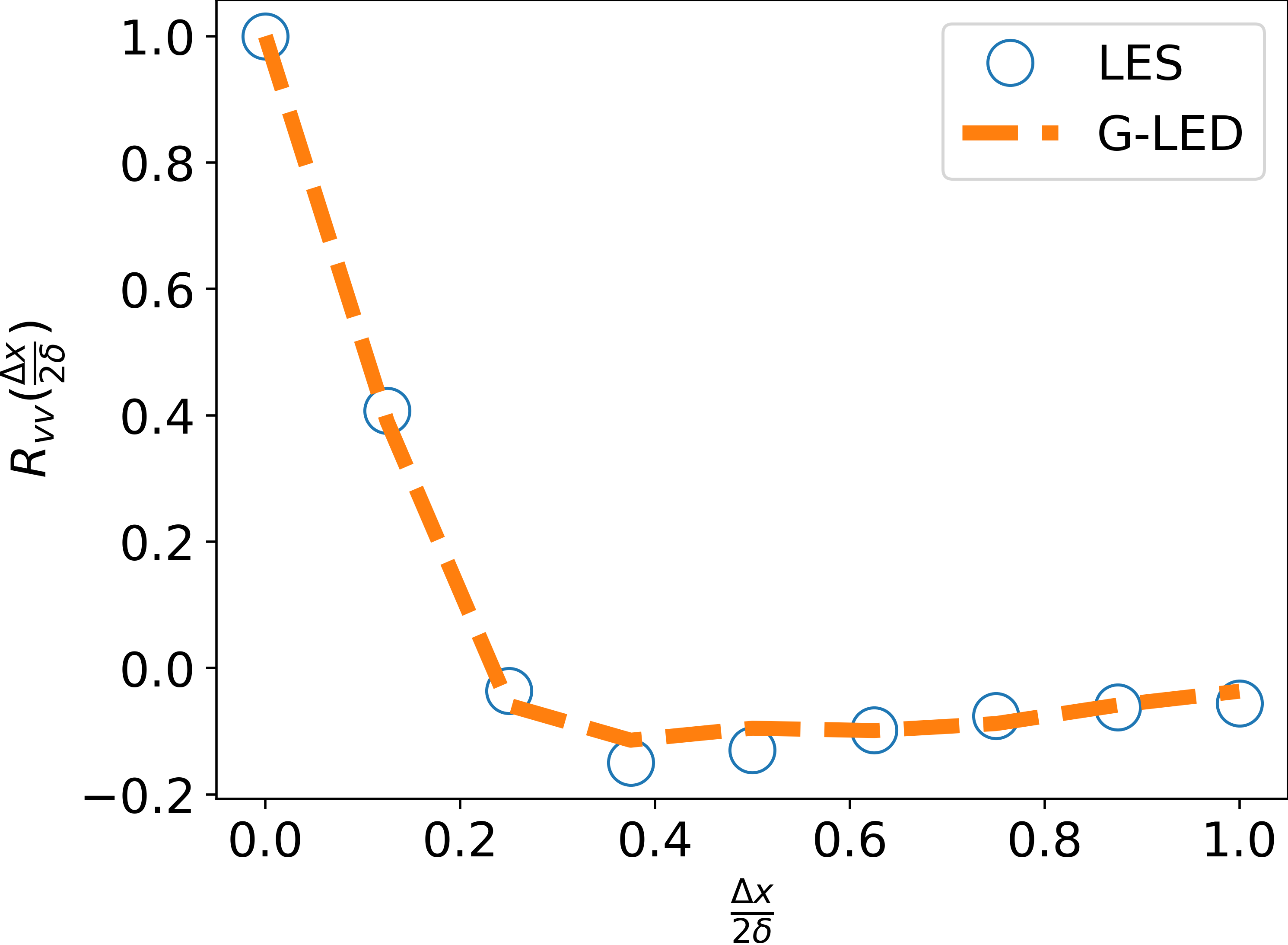}
	\includegraphics[width=0.32\textwidth]{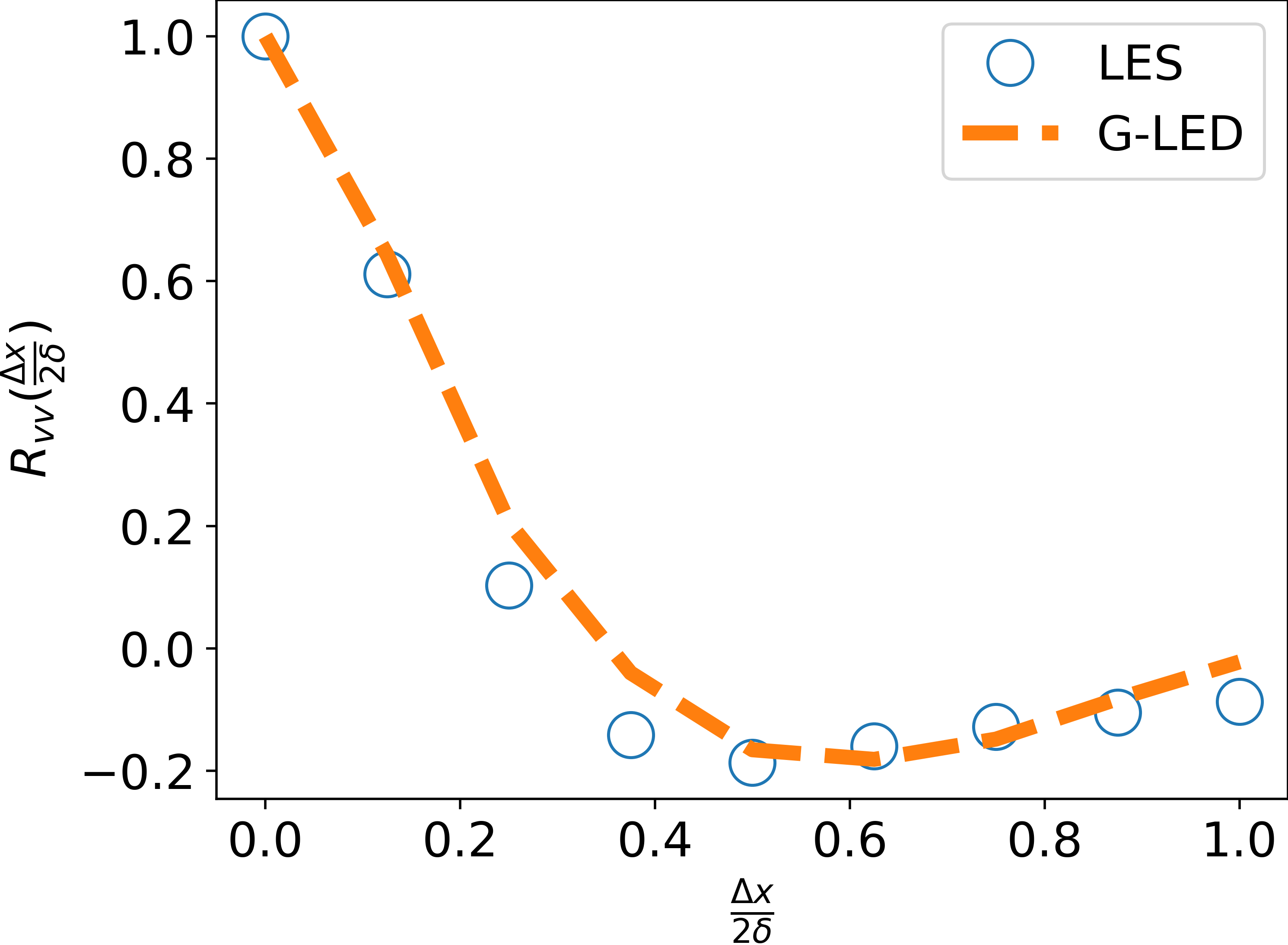}
	\vfil
	\includegraphics[width=0.32\textwidth]{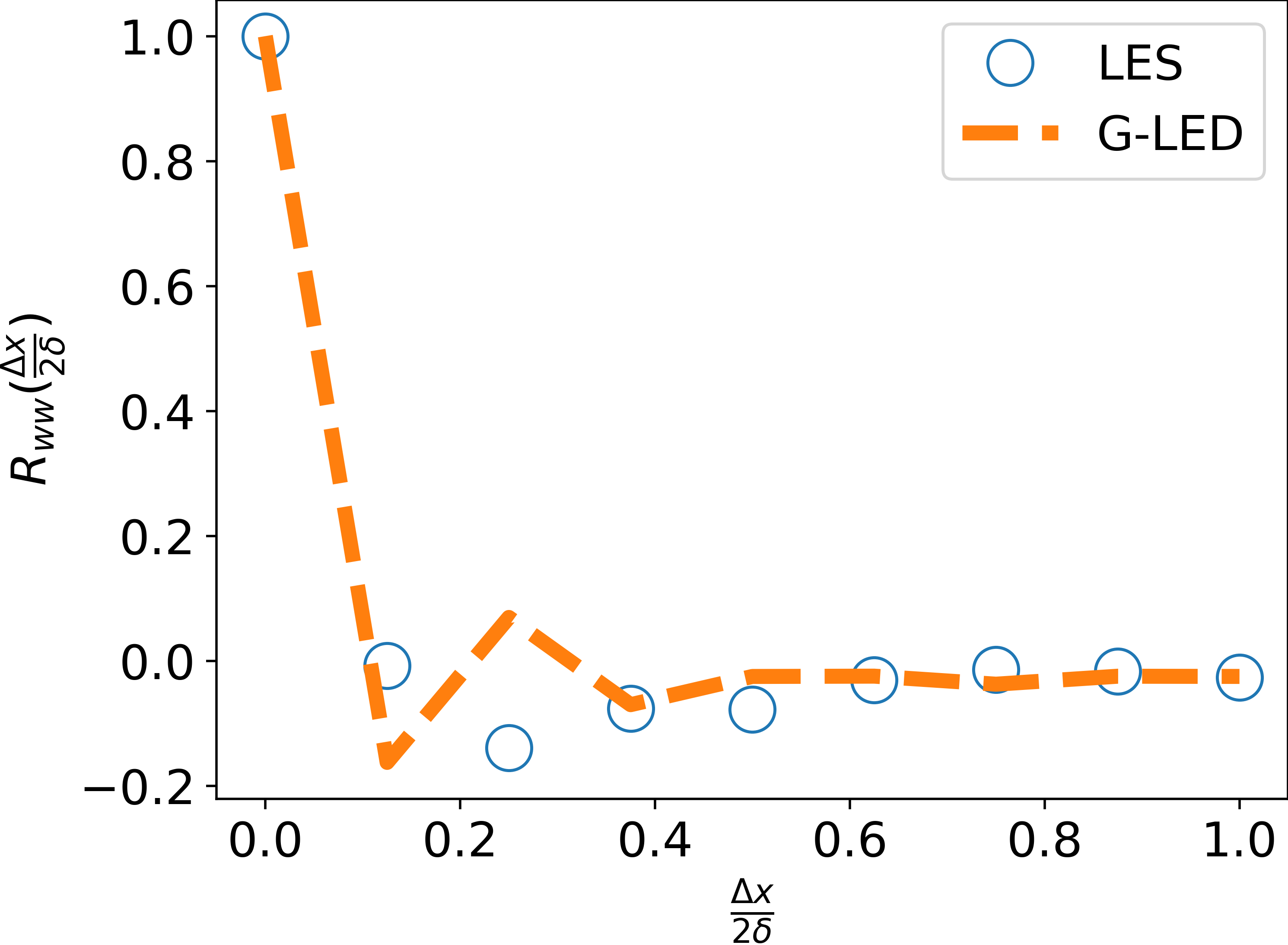}
	\includegraphics[width=0.32\textwidth]{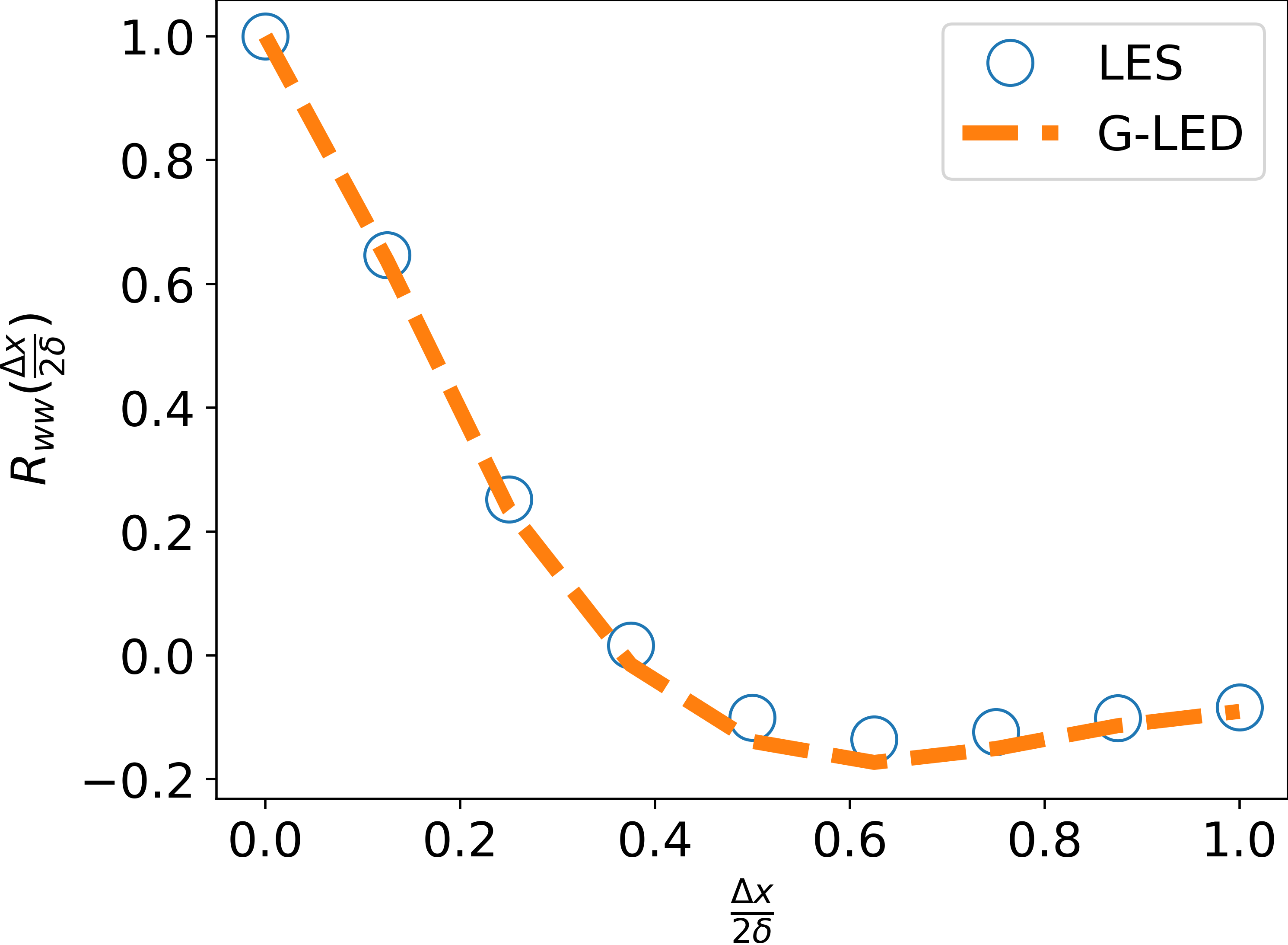}
	\includegraphics[width=0.32\textwidth]{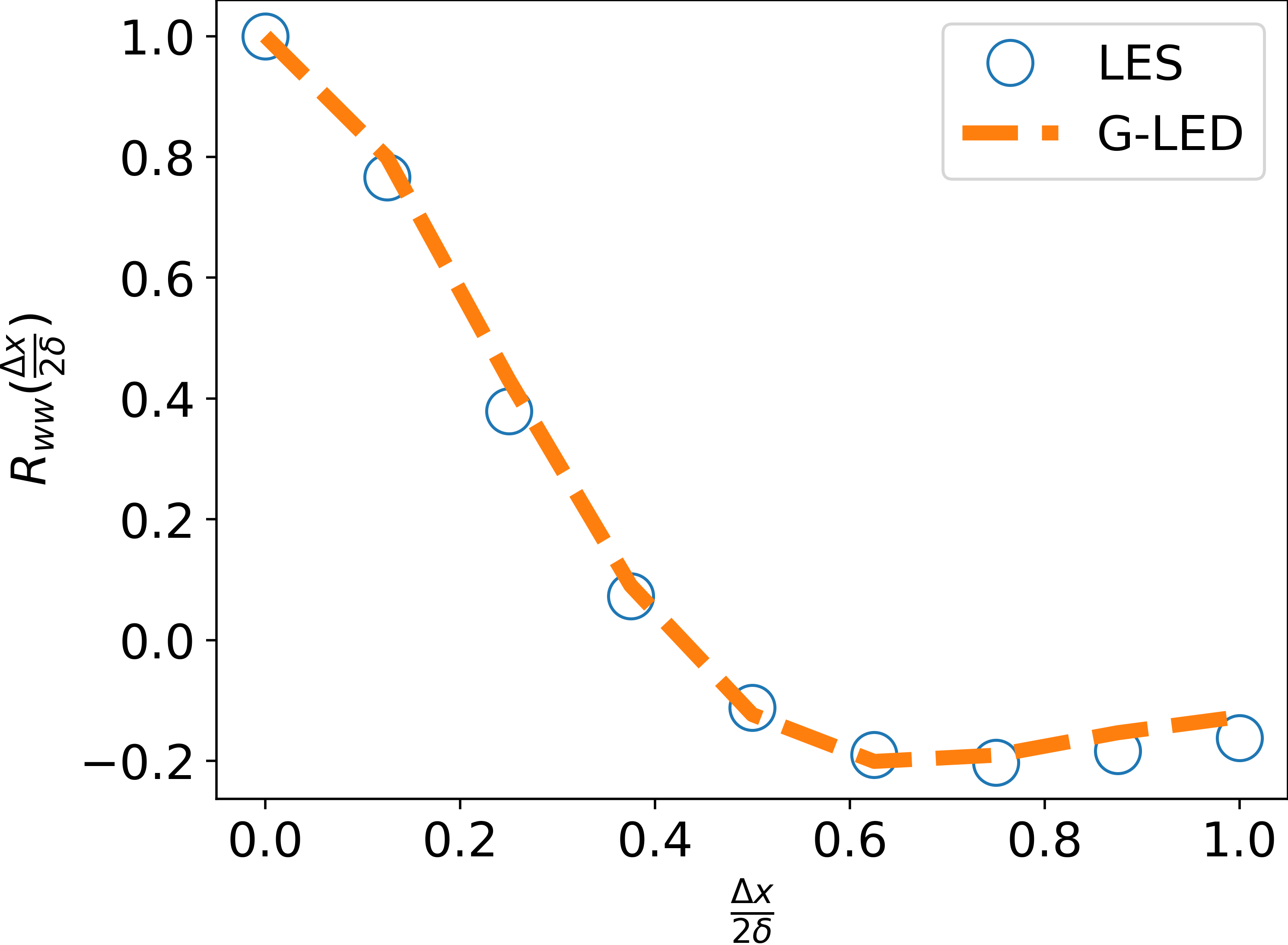}
	\caption{Spatial correlation at $y+=4.57, 171, 320$ (\textit{from left to right}) for streamwise, wallnormall and spanwise velocity (\textit{from top to bottom}). } 
\label{fig:channel_spatial_cor}
\end{figure}

\begin{figure}[htp]
	\centering
	\includegraphics[width=0.32\textwidth]{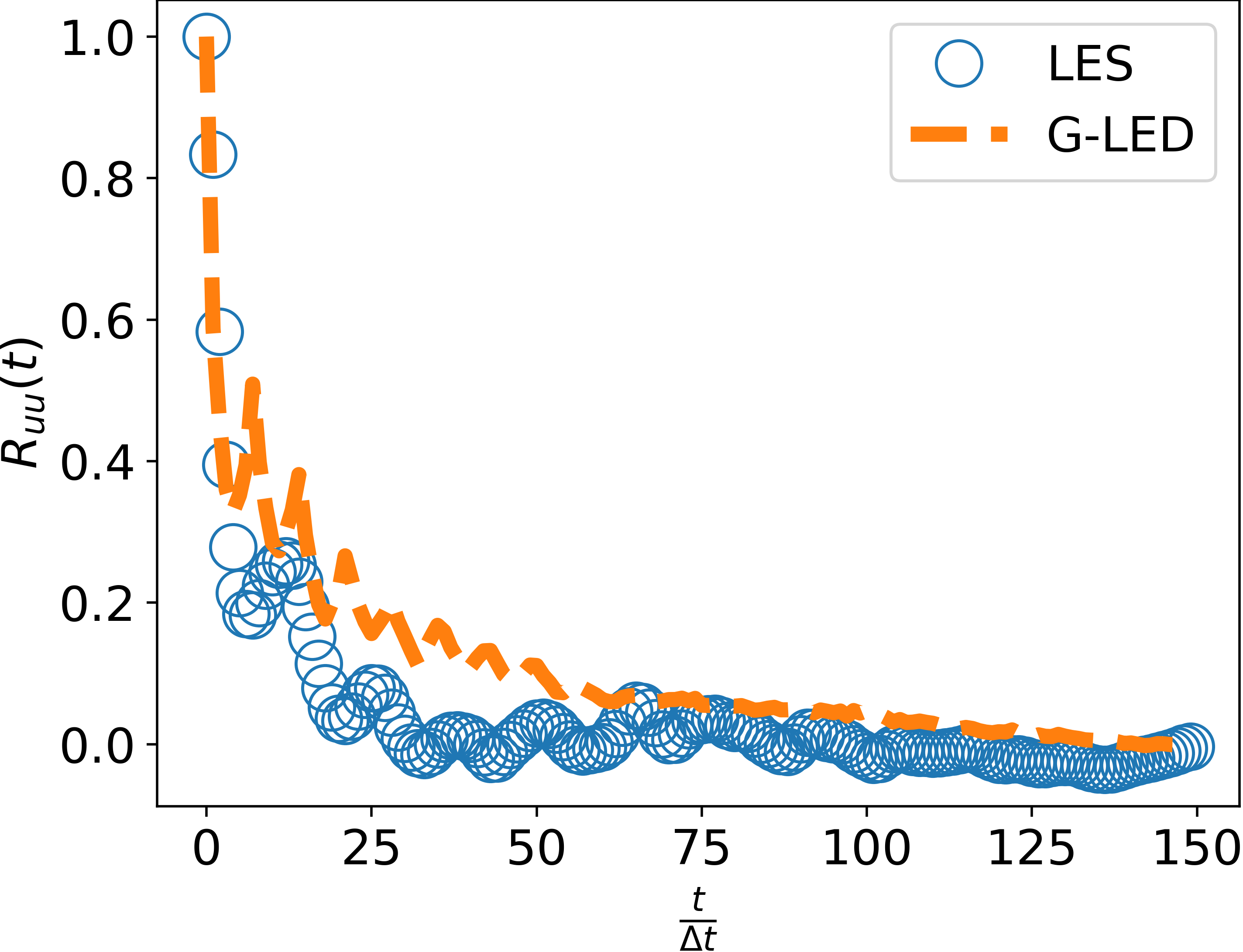}
	\includegraphics[width=0.32\textwidth]{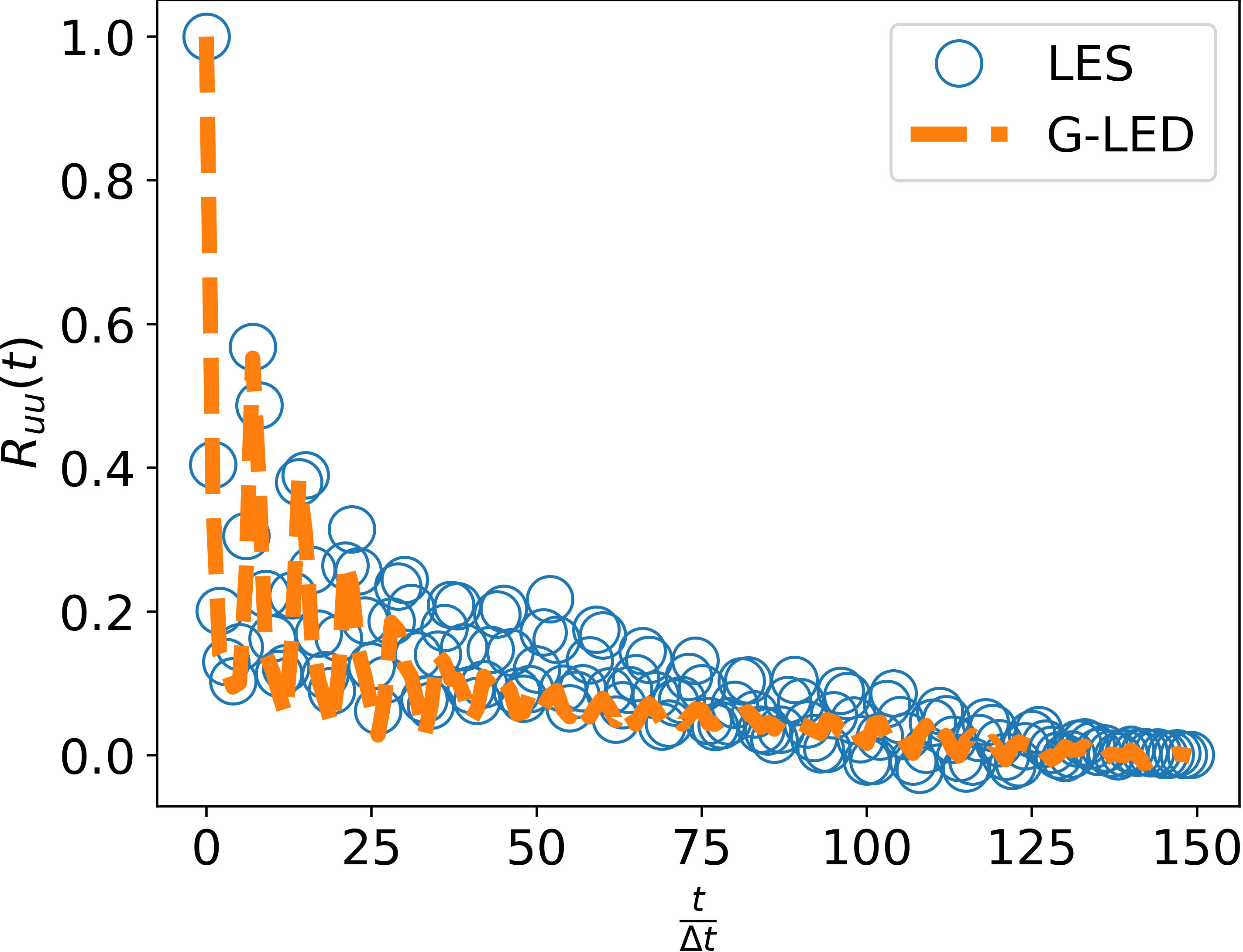}
	\includegraphics[width=0.32\textwidth]{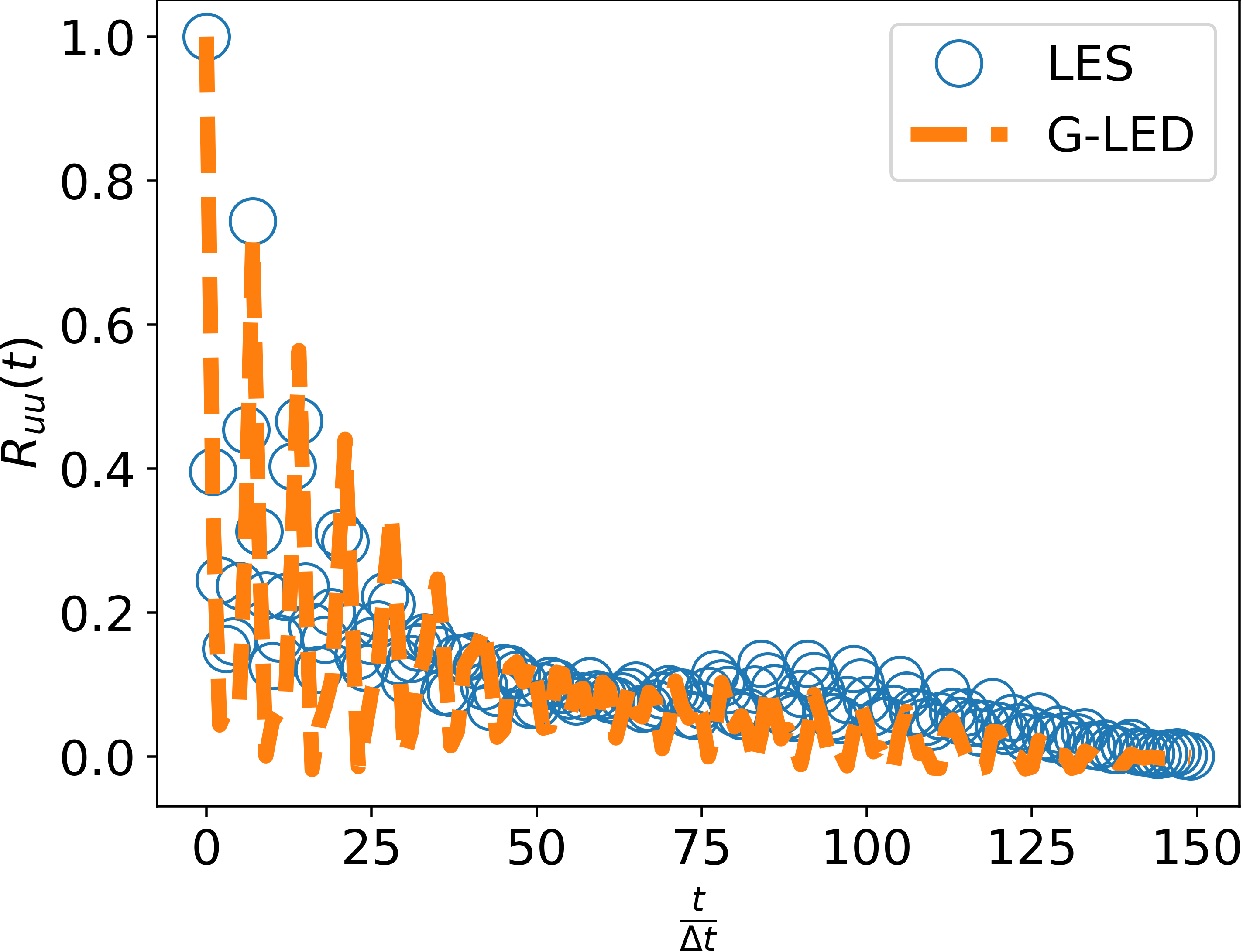}
	\vfil
	\includegraphics[width=0.32\textwidth]{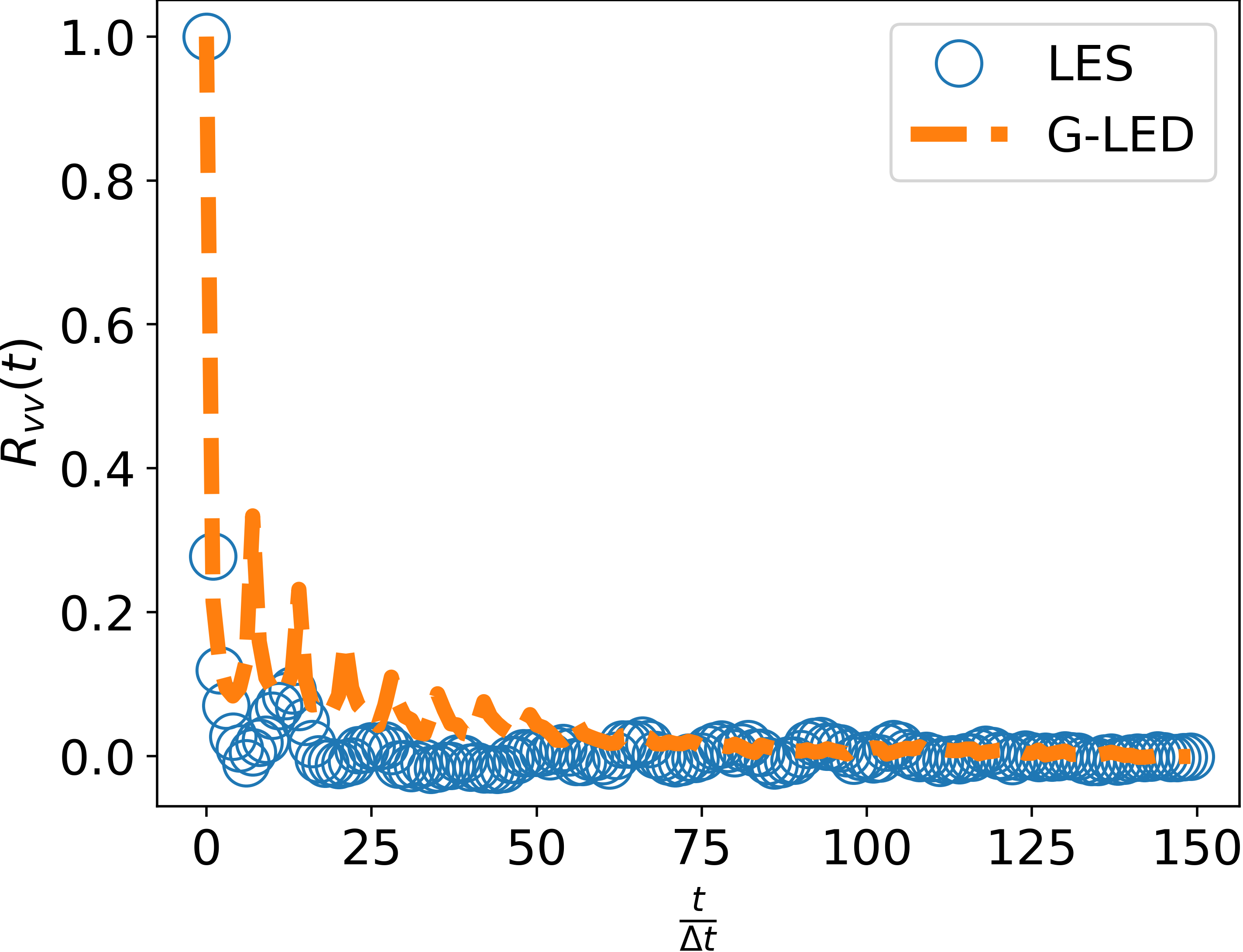}
	\includegraphics[width=0.32\textwidth]{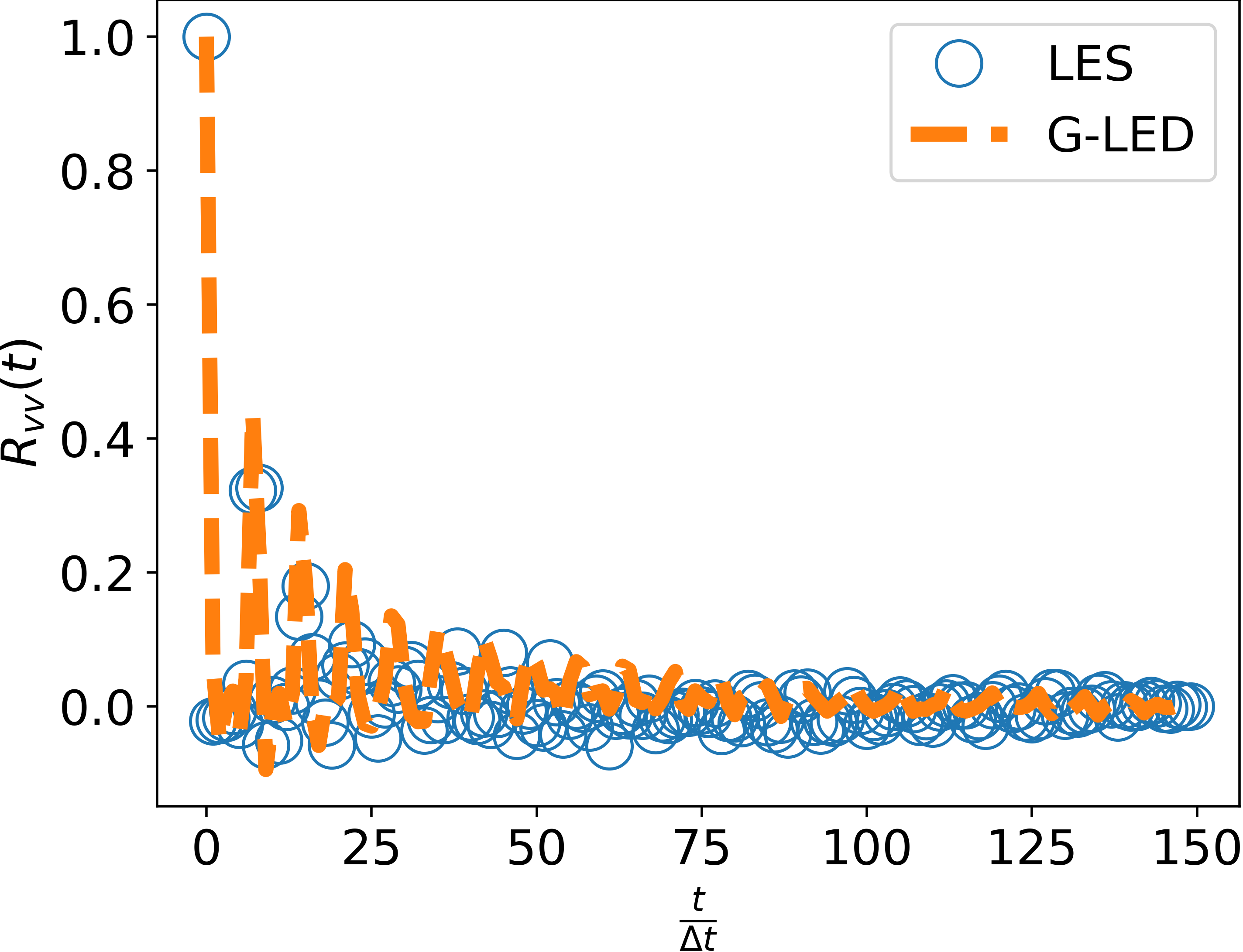}
	\includegraphics[width=0.32\textwidth]{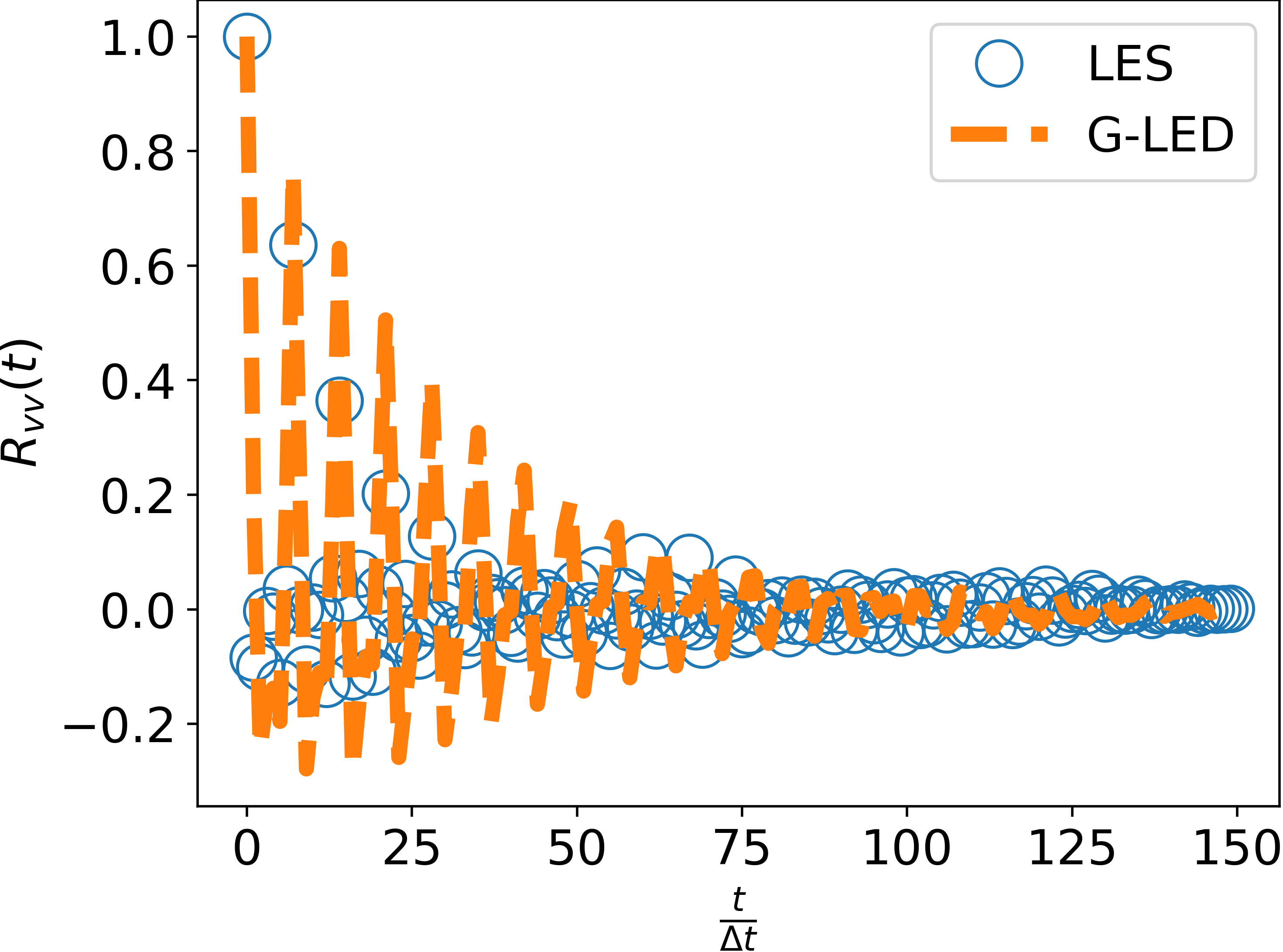}
	\vfil
	\includegraphics[width=0.32\textwidth]{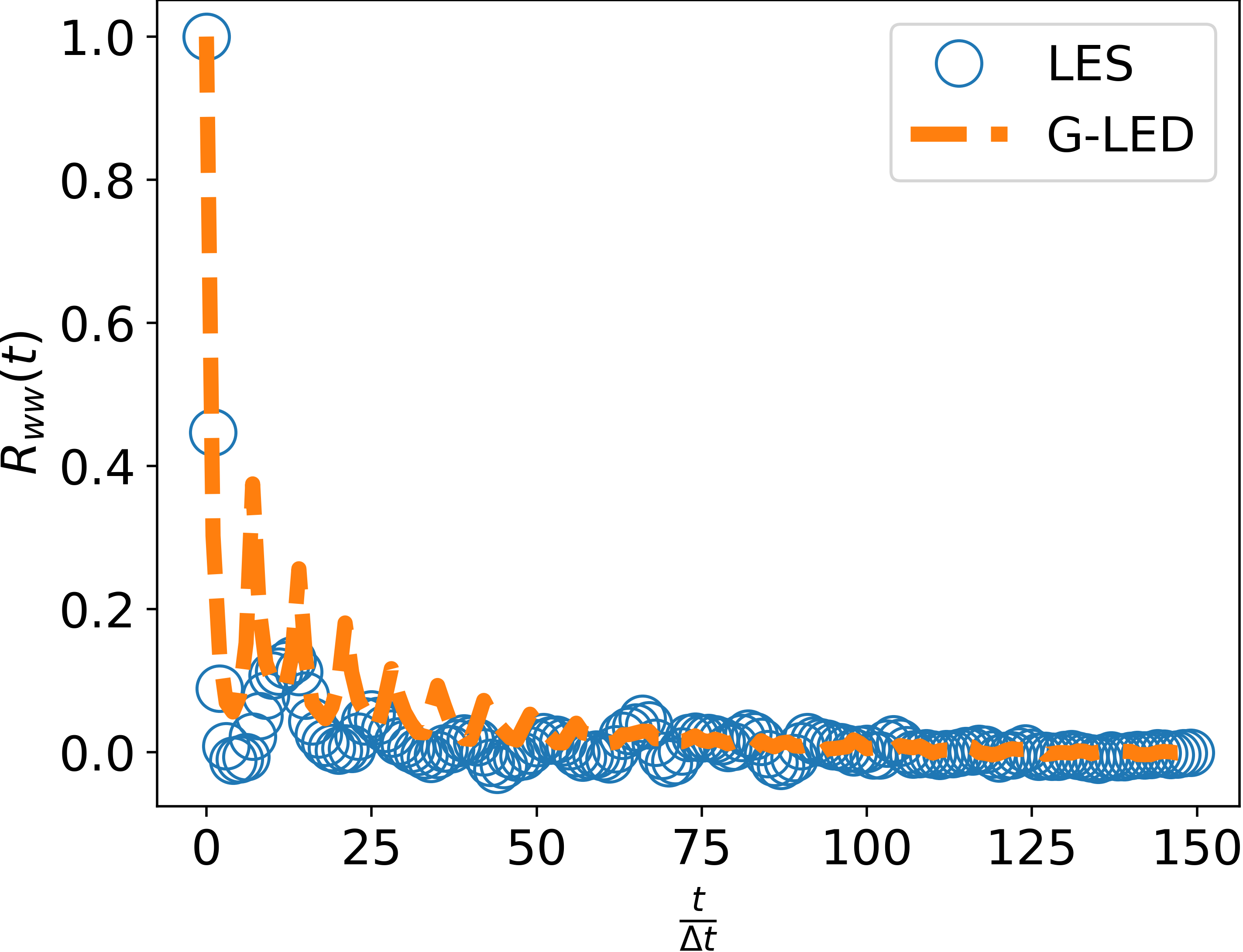}
	\includegraphics[width=0.32\textwidth]{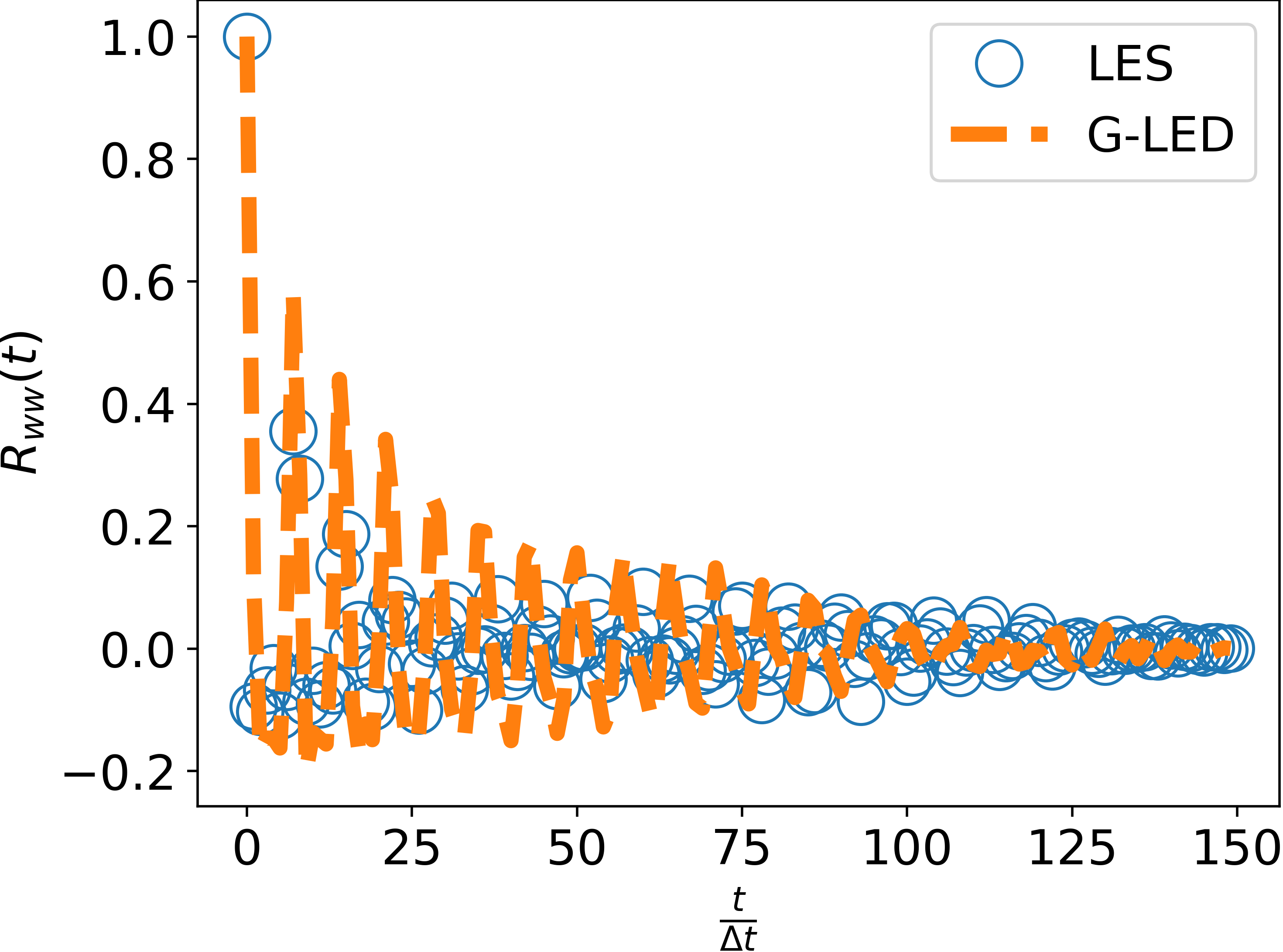}
	\includegraphics[width=0.32\textwidth]{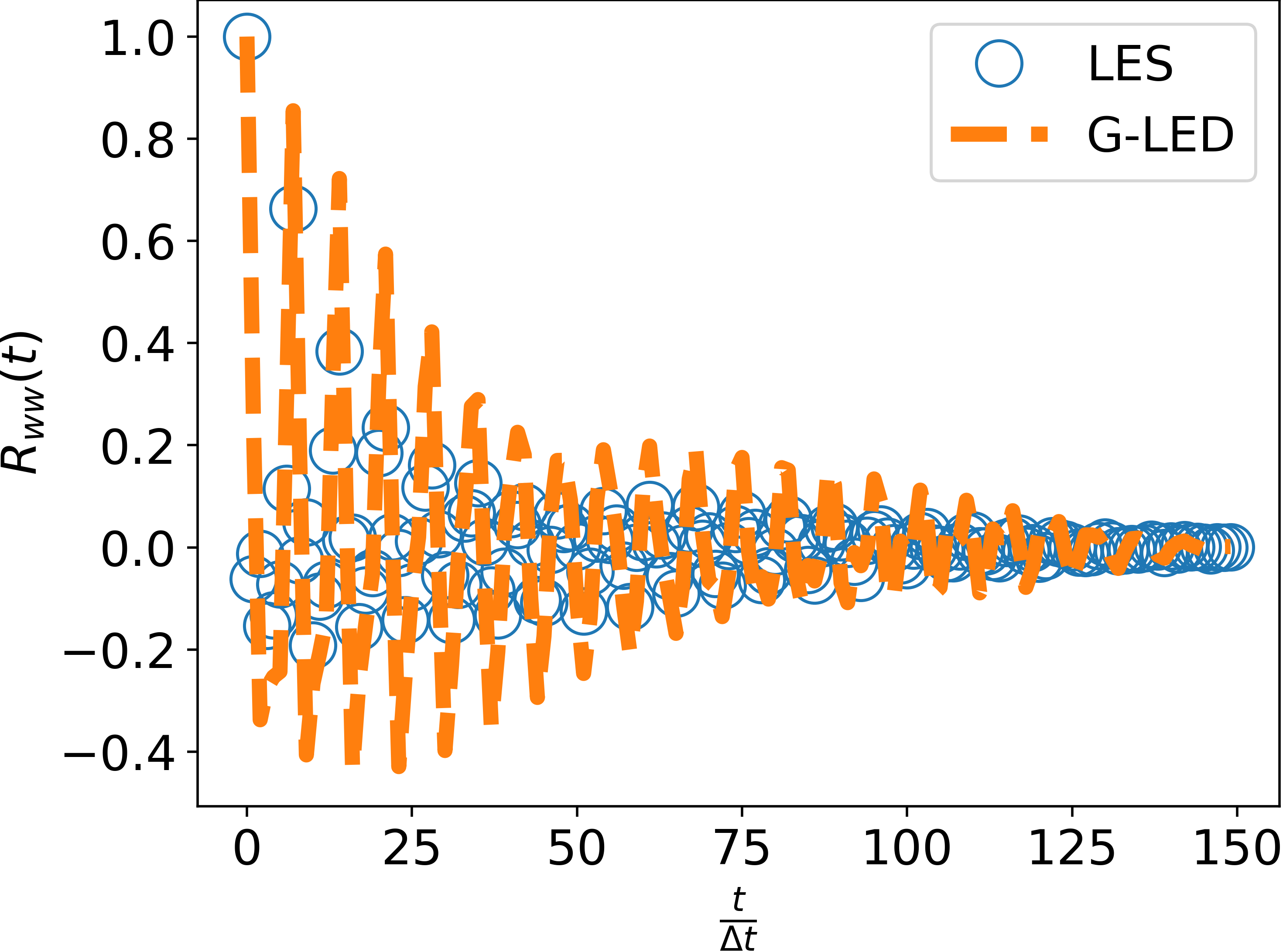}
	\caption{Temporal correlation at $y+=4.57, 171, 320$ (\textit{from left to right}) for streamwise, wallnormall and spanwise velocity (\textit{from top to bottom}). } 
 \label{fig:channel_temporal_cor}
\end{figure}

\bibliographystyle{plain}
\bibliography{biblio,references}

\begin{thebibliography}{10}

\bibitem{agathos2024accelerating}
Konstantinos Agathos, Konstantinos Vlachas, Anthony Garland, and Eleni Chatzi.
\newblock Accelerating structural dynamics simulations with localised phenomena through matrix compression and projection-based model order reduction.
\newblock {\em International Journal for Numerical Methods in Engineering}, page e7445, 2024.

\bibitem{akhare2023physics}
Deepak Akhare, Tengfei Luo, and Jian-Xun Wang.
\newblock Physics-integrated neural differentiable (pindiff) model for composites manufacturing.
\newblock {\em Computer Methods in Applied Mechanics and Engineering}, 406:115902, 2023.

\bibitem{baldi2021deep}
Pierre Baldi.
\newblock {\em Deep learning in science}.
\newblock Cambridge University Press, 2021.

\bibitem{bar2019learning}
Yohai Bar-Sinai, Stephan Hoyer, Jason Hickey, and Michael~P Brenner.
\newblock Learning data-driven discretizations for partial differential equations.
\newblock {\em Proceedings of the National Academy of Sciences}, 116(31):15344--15349, 2019.

\bibitem{bui2008model}
Tan Bui-Thanh, Karen Willcox, and Omar Ghattas.
\newblock Model reduction for large-scale systems with high-dimensional parametric input space.
\newblock {\em SIAM Journal on Scientific Computing}, 30(6):3270--3288, 2008.

\bibitem{cachay2023dyffusion}
Salva~R{\"u}hling Cachay, Bo~Zhao, Hailey James, and Rose Yu.
\newblock Dyffusion: A dynamics-informed diffusion model for spatiotemporal forecasting.
\newblock {\em arXiv preprint arXiv:2306.01984}, 2023.

\bibitem{child2019generating}
Rewon Child, Scott Gray, Alec Radford, and Ilya Sutskever.
\newblock Generating long sequences with sparse transformers.
\newblock {\em arXiv preprint arXiv:1904.10509}, 2019.

\bibitem{climatenas}
National~Research Council.
\newblock {\em A National Strategy for Advancing Climate Modeling}.
\newblock The National Academies Press, 2012.

\bibitem{cvitanovic2010state}
Predrag Cvitanovi{\'c}, Ruslan~L Davidchack, and Evangelos Siminos.
\newblock On the state space geometry of the kuramoto--sivashinsky flow in a periodic domain.
\newblock {\em SIAM Journal on Applied Dynamical Systems}, 9(1):1--33, 2010.

\bibitem{dresdner2022learning}
Gideon Dresdner, Dmitrii Kochkov, Peter~Christian Norgaard, Leonardo Zepeda-Nunez, Jamie Smith, Michael Brenner, and Stephan Hoyer.
\newblock Learning to correct spectral methods for simulating turbulent flows.
\newblock {\em Transactions on Machine Learning Research}, 2022.

\bibitem{dura2019netpyne}
Salvador Dura-Bernal, Benjamin~A Suter, Padraig Gleeson, Matteo Cantarelli, Adrian Quintana, Facundo Rodriguez, David~J Kedziora, George~L Chadderdon, Cliff~C Kerr, Samuel~A Neymotin, et~al.
\newblock Netpyne, a tool for data-driven multiscale modeling of brain circuits.
\newblock {\em Elife}, 8:e44494, 2019.

\bibitem{fan2024differentiable}
Xiantao Fan and Jian-Xun Wang.
\newblock Differentiable hybrid neural modeling for fluid-structure interaction.
\newblock {\em Journal of Computational Physics}, 496:112584, 2024.

\bibitem{fukami2019super}
Kai Fukami, Koji Fukagata, and Kunihiko Taira.
\newblock Super-resolution reconstruction of turbulent flows with machine learning.
\newblock {\em Journal of Fluid Mechanics}, 870:106--120, 2019.

\bibitem{gao2023bayesian}
Han Gao, Xu~Han, Xiantao Fan, Luning Sun, Li-Ping Liu, Lian Duan, and Jian-Xun Wang.
\newblock Bayesian conditional diffusion models for versatile spatiotemporal turbulence generation.
\newblock {\em arXiv preprint arXiv:2311.07896}, 2023.

\bibitem{geneva2020modeling}
Nicholas Geneva and Nicholas Zabaras.
\newblock Modeling the dynamics of pde systems with physics-constrained deep auto-regressive networks.
\newblock {\em Journal of Computational Physics}, 403:109056, 2020.

\bibitem{geneva2020multi}
Nicholas Geneva and Nicholas Zabaras.
\newblock Multi-fidelity generative deep learning turbulent flows.
\newblock {\em arXiv preprint arXiv:2006.04731}, 2020.

\bibitem{geneva2022transformers}
Nicholas Geneva and Nicholas Zabaras.
\newblock Transformers for modeling physical systems.
\newblock {\em Neural Networks}, 146:272--289, 2022.

\bibitem{gupta2023mori}
Priyam Gupta, Peter~J Schmid, Denis Sipp, Taraneh Sayadi, and Georgios Rigas.
\newblock Mori-zwanzig latent space koopman closure for nonlinear autoencoder.
\newblock {\em arXiv preprint arXiv:2310.10745}, 2023.

\bibitem{han2021predicting}
XU~HAN, Han Gao, Tobias Pfaff, Jian-Xun Wang, and Liping Liu.
\newblock Predicting physics in mesh-reduced space with temporal attention.
\newblock In {\em International Conference on Learning Representations}, 2021.

\bibitem{ho2020denoising}
Jonathan Ho, Ajay Jain, and Pieter Abbeel.
\newblock Denoising diffusion probabilistic models.
\newblock {\em Advances in neural information processing systems}, 33:6840--6851, 2020.

\bibitem{hochreiter1997long}
Sepp Hochreiter and J{\"u}rgen Schmidhuber.
\newblock Long short-term memory.
\newblock {\em Neural computation}, 9(8):1735--1780, 1997.

\bibitem{jacobsen2023cocogen}
Christian Jacobsen, Yilin Zhuang, and Karthik Duraisamy.
\newblock Cocogen: Physically-consistent and conditioned score-based generative models for forward and inverse problems.
\newblock {\em arXiv preprint arXiv:2312.10527}, 2023.

\bibitem{jadhav2023stressd}
Yayati Jadhav, Joseph Berthel, Chunshan Hu, Rahul Panat, Jack Beuth, and Amir~Barati Farimani.
\newblock Stressd: 2d stress estimation using denoising diffusion model.
\newblock {\em Computer Methods in Applied Mechanics and Engineering}, 416:116343, 2023.

\bibitem{jiang2023mistral}
Albert~Q Jiang, Alexandre Sablayrolles, Arthur Mensch, Chris Bamford, Devendra~Singh Chaplot, Diego de~las Casas, Florian Bressand, Gianna Lengyel, Guillaume Lample, Lucile Saulnier, et~al.
\newblock Mistral 7b.
\newblock {\em arXiv preprint arXiv:2310.06825}, 2023.

\bibitem{kaltenbach2020incorporating}
Sebastian Kaltenbach and Phaedon-Stelios Koutsourelakis.
\newblock Incorporating physical constraints in a deep probabilistic machine learning framework for coarse-graining dynamical systems.
\newblock {\em Journal of Computational Physics}, 419:109673, 2020.

\bibitem{kaltenbach2021physics}
Sebastian Kaltenbach and Phaedon-Stelios Koutsourelakis.
\newblock Physics-aware, probabilistic model order reduction with guaranteed stability.
\newblock {\em ICLR}, 2021.

\bibitem{kaltenbach2023interpretable}
Sebastian Kaltenbach, Phaedon-Stelios Koutsourelakis, and Petros Koumoutsakos.
\newblock Interpretable reduced-order modeling with time-scale separation.
\newblock {\em arXiv preprint arXiv:2303.02189}, 2023.

\bibitem{kaltenbach2023semi}
Sebastian Kaltenbach, Paris Perdikaris, and Phaedon-Stelios Koutsourelakis.
\newblock Semi-supervised invertible neural operators for bayesian inverse problems.
\newblock {\em Computational Mechanics}, pages 1--20, 2023.

\bibitem{kassam2005fourth}
Aly-Khan Kassam and Lloyd~N Trefethen.
\newblock Fourth-order time-stepping for stiff pdes.
\newblock {\em SIAM Journal on Scientific Computing}, 26(4):1214--1233, 2005.

\bibitem{katharopoulos2020transformers}
Angelos Katharopoulos, Apoorv Vyas, Nikolaos Pappas, and Fran{\c{c}}ois Fleuret.
\newblock Transformers are rnns: Fast autoregressive transformers with linear attention.
\newblock In {\em International conference on machine learning}, pages 5156--5165. PMLR, 2020.

\bibitem{kevrekidis2004equation}
Ioannis~G Kevrekidis, C~William Gear, and Gerhard Hummer.
\newblock Equation-free: The computer-aided analysis of complex multiscale systems.
\newblock {\em AIChE Journal}, 50(7):1346--1355, 2004.

\bibitem{kevrekidis2003equation}
Ioannis~G Kevrekidis, C~William Gear, James~M Hyman, Panagiotis~G Kevrekidid, Olof Runborg, Constantinos Theodoropoulos, et~al.
\newblock Equation-free, coarse-grained multiscale computation: Enabling mocroscopic simulators to perform system-level analysis.
\newblock {\em Communications in Mathematical Sciences}, 1(4):715--762, 2003.

\bibitem{kivcic2023adaptive}
Ivica Ki{\v{c}}i{\'c}, Pantelis~R Vlachas, Georgios Arampatzis, Michail Chatzimanolakis, Leonidas Guibas, and Petros Koumoutsakos.
\newblock Adaptive learning of effective dynamics for online modeling of complex systems.
\newblock {\em Computer Methods in Applied Mechanics and Engineering}, 415:116204, 2023.

\bibitem{kingma2021variational}
Diederik Kingma, Tim Salimans, Ben Poole, and Jonathan Ho.
\newblock Variational diffusion models.
\newblock {\em Advances in neural information processing systems}, 34:21696--21707, 2021.

\bibitem{kingma2013auto}
Diederik~P Kingma and Max Welling.
\newblock Auto-encoding variational bayes.
\newblock {\em arXiv preprint arXiv:1312.6114}, 2013.

\bibitem{kohl2020learning}
Georg Kohl, Kiwon Um, and Nils Thuerey.
\newblock Learning similarity metrics for numerical simulations.
\newblock In {\em International Conference on Machine Learning}, pages 5349--5360. PMLR, 2020.

\bibitem{kovachki2023neural}
Nikola Kovachki, Zongyi Li, Burigede Liu, Kamyar Azizzadenesheli, Kaushik Bhattacharya, Andrew Stuart, and Anima Anandkumar.
\newblock Neural operator: Learning maps between function spaces with applications to pdes.
\newblock {\em Journal of Machine Learning Research}, 24(89):1--97, 2023.

\bibitem{li2023seeds}
Lizao Li, Rob Carver, Ignacio Lopez-Gomez, Fei Sha, and John Anderson.
\newblock Seeds: Emulation of weather forecast ensembles with diffusion models.
\newblock {\em arXiv preprint arXiv:2306.14066}, 2023.

\bibitem{li2022fourier}
Zongyi Li, Daniel~Zhengyu Huang, Burigede Liu, and Anima Anandkumar.
\newblock Fourier neural operator with learned deformations for pdes on general geometries.
\newblock {\em arXiv preprint arXiv:2207.05209}, 2022.

\bibitem{lienen2023generative}
Marten Lienen, Jan Hansen-Palmus, David L{\"u}dke, and Stephan G{\"u}nnemann.
\newblock Generative diffusion for 3d turbulent flows.
\newblock {\em arXiv preprint arXiv:2306.01776}, 2023.

\bibitem{linot2023stabilized}
Alec~J Linot, Joshua~W Burby, Qi~Tang, Prasanna Balaprakash, Michael~D Graham, and Romit Maulik.
\newblock Stabilized neural ordinary differential equations for long-time forecasting of dynamical systems.
\newblock {\em Journal of Computational Physics}, 474:111838, 2023.

\bibitem{linot2020deep}
Alec~J Linot and Michael~D Graham.
\newblock Deep learning to discover and predict dynamics on an inertial manifold.
\newblock {\em Physical Review E}, 101(6):062209, 2020.

\bibitem{lu2021learning}
Lu~Lu, Pengzhan Jin, Guofei Pang, Zhongqiang Zhang, and George~Em Karniadakis.
\newblock Learning nonlinear operators via deeponet based on the universal approximation theorem of operators.
\newblock {\em Nature machine intelligence}, 3(3):218--229, 2021.

\bibitem{mahadevan2016impact}
Amala Mahadevan.
\newblock The impact of submesoscale physics on primary productivity of plankton.
\newblock {\em Annual review of marine science}, 8:161--184, 2016.

\bibitem{menier2023interpretable}
Emmanuel Menier, Sebastian Kaltenbach, Mouadh Yagoubi, Marc Schoenauer, and Petros Koumoutsakos.
\newblock Interpretable learning of effective dynamics for multiscale systems.
\newblock {\em arXiv preprint arXiv:2309.05812}, 2023.

\bibitem{mezic2013analysis}
Igor Mezi{\'c}.
\newblock Analysis of fluid flows via spectral properties of the koopman operator.
\newblock {\em Annual review of fluid mechanics}, 45:357--378, 2013.

\bibitem{mezic2021koopman}
Igor Mezi{\'c}.
\newblock Koopman operator, geometry, and learning of dynamical systems.
\newblock {\em Not. Am. Math. Soc.}, 68(7):1087--1105, 2021.

\bibitem{moser2023numerical}
Robert~D Moser.
\newblock Numerical challenges in turbulence simulation.
\newblock In {\em Numerical Methods in Turbulence Simulation}, pages 1--43. Elsevier, 2023.

\bibitem{it:2015-014}
Timofey Mukha and Mattias Liefvendahl.
\newblock Large-eddy simulation of turbulent channel flow.
\newblock Technical Report 2015-014, Department of Information Technology, Uppsala University, May 2015.

\bibitem{nicoud1999subgrid}
Franck Nicoud and Fr{\'e}d{\'e}ric Ducros.
\newblock Subgrid-scale stress modelling based on the square of the velocity gradient tensor.
\newblock {\em Flow, turbulence and Combustion}, 62(3):183--200, 1999.

\bibitem{ogoke2023inexpensive}
Francis Ogoke, Quanliang Liu, Olabode Ajenifujah, Alexander Myers, Guadalupe Quirarte, Jack Beuth, Jonathan Malen, and Amir~Barati Farimani.
\newblock Inexpensive high fidelity melt pool models in additive manufacturing using generative deep diffusion.
\newblock {\em arXiv preprint arXiv:2311.16168}, 2023.

\bibitem{Palmer2015}
Tim Palmer.
\newblock Modelling: Build imprecise supercomputers.
\newblock {\em Nature}, 526(7571):32--33, 2015.

\bibitem{pan2023neural}
Shaowu Pan, Steven~L Brunton, and J~Nathan Kutz.
\newblock Neural implicit flow: a mesh-agnostic dimensionality reduction paradigm of spatio-temporal data.
\newblock {\em Journal of Machine Learning Research}, 24(41):1--60, 2023.

\bibitem{peng2021multiscale}
Grace~CY Peng, Mark Alber, Adrian Buganza~Tepole, William~R Cannon, Suvranu De, Savador Dura-Bernal, Krishna Garikipati, George Karniadakis, William~W Lytton, Paris Perdikaris, et~al.
\newblock Multiscale modeling meets machine learning: What can we learn?
\newblock {\em Archives of Computational Methods in Engineering}, 28:1017--1037, 2021.

\bibitem{pfaff2020learning}
Tobias Pfaff, Meire Fortunato, Alvaro Sanchez-Gonzalez, and Peter~W Battaglia.
\newblock Learning mesh-based simulation with graph networks.
\newblock {\em arXiv preprint arXiv:2010.03409}, 2020.

\bibitem{pope2001turbulent}
Stephen~B Pope.
\newblock Turbulent flows.
\newblock {\em Measurement Science and Technology}, 12(11):2020--2021, 2001.

\bibitem{prabhudesai2023diffusion}
Mihir Prabhudesai, Tsung-Wei Ke, Alexander~Cong Li, Deepak Pathak, and Katerina Fragkiadaki.
\newblock Diffusion-tta: Test-time adaptation of discriminative models via generative feedback.
\newblock In {\em Thirty-seventh Conference on Neural Information Processing Systems}, 2023.

\bibitem{price2023gencast}
Ilan Price, Alvaro Sanchez-Gonzalez, Ferran Alet, Timo Ewalds, Andrew El-Kadi, Jacklynn Stott, Shakir Mohamed, Peter Battaglia, Remi Lam, and Matthew Willson.
\newblock Gencast: Diffusion-based ensemble forecasting for medium-range weather.
\newblock {\em arXiv preprint arXiv:2312.15796}, 2023.

\bibitem{racca2022modelling}
Alberto Racca, Nguyen Anh~Khoa Doan, and Luca Magri.
\newblock Modelling spatiotemporal turbulent dynamics with the convolutional autoencoder echo state network.
\newblock {\em arXiv preprint arXiv:2211.11379}, 2022.

\bibitem{radford2019language}
Alec Radford, Jeffrey Wu, Rewon Child, David Luan, Dario Amodei, Ilya Sutskever, et~al.
\newblock Language models are unsupervised multitask learners.
\newblock {\em OpenAI blog}, 1(8):9, 2019.

\bibitem{ren2023physr}
Pu~Ren, Chengping Rao, Yang Liu, Zihan Ma, Qi~Wang, Jian-Xun Wang, and Hao Sun.
\newblock Physr: Physics-informed deep super-resolution for spatiotemporal data.
\newblock {\em Journal of Computational Physics}, 492:112438, 2023.

\bibitem{ren2022phycrnet}
Pu~Ren, Chengping Rao, Yang Liu, Jian-Xun Wang, and Hao Sun.
\newblock Phycrnet: Physics-informed convolutional-recurrent network for solving spatiotemporal pdes.
\newblock {\em Computer Methods in Applied Mechanics and Engineering}, 389:114399, 2022.

\bibitem{rossinelli2013a}
Diego Rossinelli, Babak Hejazialhosseini, Panagiotis Hadjidoukas, Costas Bekas, Alessandro Curioni, Adam Bertsch, Scott Futral, Steffen~J. Schmidt, Nikolaus~A. Adams, and Petros Koumoutsakos.
\newblock 11 {PFLOP/s} simulations of cloud cavitation collapse.
\newblock In {\em Proceedings of the International Conference on High Performance Computing, Networking, Storage and Analysis}, SC '13, pages 3:1--3:13, New York, NY, USA, 2013. ACM.

\bibitem{saharia2022photorealistic}
Chitwan Saharia, William Chan, Saurabh Saxena, Lala Li, Jay Whang, Emily~L Denton, Kamyar Ghasemipour, Raphael Gontijo~Lopes, Burcu Karagol~Ayan, Tim Salimans, et~al.
\newblock Photorealistic text-to-image diffusion models with deep language understanding.
\newblock {\em Advances in Neural Information Processing Systems}, 35:36479--36494, 2022.

\bibitem{sanchez2020learning}
Alvaro Sanchez-Gonzalez, Jonathan Godwin, Tobias Pfaff, Rex Ying, Jure Leskovec, and Peter Battaglia.
\newblock Learning to simulate complex physics with graph networks.
\newblock In {\em International conference on machine learning}, pages 8459--8468. PMLR, 2020.

\bibitem{sharma2024exawind}
Ashesh Sharma, Michael~J Brazell, Ganesh Vijayakumar, Shreyas Ananthan, Lawrence Cheung, Nathaniel deVelder, Marc~T Henry~de Frahan, Neil Matula, Paul Mullowney, Jon Rood, et~al.
\newblock Exawind: Open-source cfd for hybrid-rans/les geometry-resolved wind turbine simulations in atmospheric flows.
\newblock {\em Wind Energy}, 2024.

\bibitem{shu2023physics}
Dule Shu, Zijie Li, and Amir~Barati Farimani.
\newblock A physics-informed diffusion model for high-fidelity flow field reconstruction.
\newblock {\em Journal of Computational Physics}, 478:111972, 2023.

\bibitem{song2020denoising}
Jiaming Song, Chenlin Meng, and Stefano Ermon.
\newblock Denoising diffusion implicit models.
\newblock {\em arXiv preprint arXiv:2010.02502}, 2020.

\bibitem{song2019generative}
Yang Song and Stefano Ermon.
\newblock Generative modeling by estimating gradients of the data distribution.
\newblock {\em Advances in neural information processing systems}, 32, 2019.

\bibitem{song2020improved}
Yang Song and Stefano Ermon.
\newblock Improved techniques for training score-based generative models.
\newblock {\em Advances in neural information processing systems}, 33:12438--12448, 2020.

\bibitem{song2020score}
Yang Song, Jascha Sohl-Dickstein, Diederik~P Kingma, Abhishek Kumar, Stefano Ermon, and Ben Poole.
\newblock Score-based generative modeling through stochastic differential equations.
\newblock {\em arXiv preprint arXiv:2011.13456}, 2020.

\bibitem{sun2023unifying}
Luning Sun, Xu~Han, Han Gao, Jian-Xun Wang, and Liping Liu.
\newblock Unifying predictions of deterministic and stochastic physics in mesh-reduced space with sequential flow generative model.
\newblock In {\em Thirty-seventh Conference on Neural Information Processing Systems}, 2023.

\bibitem{tao2010nonintrusive}
Molei Tao, Houman Owhadi, and Jerrold~E Marsden.
\newblock Nonintrusive and structure preserving multiscale integration of stiff odes, sdes, and hamiltonian systems with hidden slow dynamics via flow averaging.
\newblock {\em Multiscale Modeling and Simulation}, 8(4):1269--1324, 2010.

\bibitem{vaswani2017attention}
Ashish Vaswani, Noam Shazeer, Niki Parmar, Jakob Uszkoreit, Llion Jones, Aidan~N Gomez, {\L}ukasz Kaiser, and Illia Polosukhin.
\newblock Attention is all you need.
\newblock {\em Advances in neural information processing systems}, 30, 2017.

\bibitem{vlachas2022multiscale}
Pantelis~R Vlachas, Georgios Arampatzis, Caroline Uhler, and Petros Koumoutsakos.
\newblock Multiscale simulations of complex systems by learning their effective dynamics.
\newblock {\em Nature Machine Intelligence}, 4(4):359--366, 2022.

\bibitem{vlachas2018data}
Pantelis~R Vlachas, Wonmin Byeon, Zhong~Y Wan, Themistoklis~P Sapsis, and Petros Koumoutsakos.
\newblock Data-driven forecasting of high-dimensional chaotic systems with long short-term memory networks.
\newblock {\em Proceedings of the Royal Society A: Mathematical, Physical and Engineering Sciences}, 474(2213):20170844, 2018.

\bibitem{vlachas2018}
Pantelis~R Vlachas, Wonmin Byeon, Zhong~Y Wan, Themistoklis~P Sapsis, and Petros Koumoutsakos.
\newblock Data-driven forecasting of high-dimensional chaotic systems with long short-term memory networks.
\newblock {\em Proc. R. Soc. A}, 474(2213):20170844, 2018.

\bibitem{weinan2007heterogeneous}
E~Weinan, Bjorn Engquist, Xiantao Li, Weiqing Ren, and Eric Vanden-Eijnden.
\newblock Heterogeneous multiscale methods: a review.
\newblock {\em Communications in computational physics}, 2(3):367--450, 2007.

\bibitem{wilcox1988multiscale}
David~C Wilcox.
\newblock Multiscale model for turbulent flows.
\newblock {\em AIAA journal}, 26(11):1311--1320, 1988.

\bibitem{zhang2019deep}
Ruiyang Zhang, Zhao Chen, Su~Chen, Jingwei Zheng, Oral B{\"u}y{\"u}k{\"o}zt{\"u}rk, and Hao Sun.
\newblock Deep long short-term memory networks for nonlinear structural seismic response prediction.
\newblock {\em Computers \& Structures}, 220:55--68, 2019.

\bibitem{zhang2020physics}
Ruiyang Zhang, Yang Liu, and Hao Sun.
\newblock Physics-informed multi-lstm networks for metamodeling of nonlinear structures.
\newblock {\em Computer Methods in Applied Mechanics and Engineering}, 369:113226, 2020.

\end{thebibliography}

\end{document}